\documentclass[jair,11pt,twoside,theapa,letter]{article}

\usepackage{theapa}
\usepackage{helvet}
\usepackage{rotating}
\usepackage{jair}
\usepackage{graphicx}

\jairheading{26}{2006}{35-99}{8/05}{5/06} \ShortHeadings{Planning
Graph Heuristics for Belief Space Search}{Bryce, Kambhampati, \&
Smith}

\def\citep#1{\cite{#1}}
\def\citet#1{\citeA{#1}}
\input epsf
\usepackage{times}
\usepackage{courier}
\usepackage{latexsym}
\usepackage{amsmath, amssymb}
\usepackage{epic}
\usepackage{eepic}
\usepackage{graphics}

\def\caltalt{{\cal C}{\sl AltAlt}}

\def\bold#1{{#1}}
\def\und#1{\medskip{\noindent\bf #1:}}

\begin{document}

\title{Planning Graph Heuristics for Belief Space Search}

%%%%%%%%%%%%%%%PS/PDF%%%%%%%%%%%%%%%%%%%
\author{\name Daniel Bryce \email dan.bryce@asu.edu\\
\name Subbarao Kambhampati, \email rao@asu.edu\\
\addr Department of Computer Science and Engineering\\
 Ira A. Fulton School of Engineering\\
 Arizona State University,  Brickyard Suite 501\\
 699 South Mill Avenue, Tempe, AZ 85281
 \AND \name David E. Smith \email de2smith@email.arc.nasa.gov \\
\addr NASA Ames Research Center\\
Intelligent Systems Division, MS 269-2\\
Moffett Field, CA 94035-1000 }
%%%%%%%%%%%%%%%%%%%%%%%%%%%%%%%%%%%%%%%%

%%%%%%%%%%%%HTML%%%%%%%%%%%%%%%%%%%%%%%%
%\author{
%Daniel Bryce \& Subbarao Kambhampati\\
%\{dan.bryce, rao\}\@ asu.edu\\
%Department of Computer Science and Engineering\\
%Ira A. Fulton School of Engineering\\
%Arizona State University,  Brickyard Suite 501\\
%699 South Mill Avenue, Tempe, AZ 85281\\
%\and
%David E. Smith\\
%de2smith@email.arc.nasa.gov \\
%NASA Ames Research Center\\
%Intelligent Systems Division, MS 269-2\\
%Moffett Field, CA 94035-1000
%}
%%%%%%%%%%%%%%%%%%%%%%%%%%%%%%%%%%%%%%%%

\maketitle

\begin{abstract}
Some recent works in conditional planning have proposed reachability
heuristics to improve planner scalability, but many lack a formal
description of the properties of their distance estimates. To place
previous work in context and extend work on heuristics for
conditional planning, we provide a formal basis for distance
estimates between belief states.  We give a definition for the
distance between belief states that relies on aggregating underlying
state distance measures.  We give several techniques to aggregate
state distances and their associated properties.  Many existing
heuristics exhibit a subset of the properties, but in order to
provide a standardized comparison we present several generalizations
of planning graph heuristics that are used in a single planner.  We
compliment our belief state distance estimate framework by also
investigating efficient planning graph data structures that
incorporate BDDs to compute the most effective heuristics.

We developed two planners to serve as test-beds for our
investigation.  The first, \caltalt, is a conformant regression
planner that uses A* search.  The second, $POND$, is a conditional
progression planner that uses AO* search. We show the relative
effectiveness of our heuristic techniques within these planners.
We also compare the performance of these planners with several
state of the art approaches in conditional planning.
\end{abstract}

\section{Introduction}

Ever since CGP \citep{AAAI98_IAAI98*889} and SGP \citep{SGP} a
series of planners have been developed for tackling conformant and
conditional planning problems -- including GPT
\citep{bonet00planning}, C-Plan \citep{castellini01satconformant},
PKSPlan \citep{petrick02kbincomplete}, Frag-Plan
\citep{kurien02fragplan}, MBP \citep{bertoli01planning}, KACMBP
\citep{bertoli02kacmbp}, CFF \citep{cff}, and YKA
\citep{Rintanen03}. Several of these planners are extensions of
heuristic state space planners that search in the space of ``belief
states'' (where a belief state is a set of possible states). Without
full-observability, agents need belief states to capture state
uncertainty arising from starting in an uncertain state or by
executing actions with uncertain effects in a known state.  We focus
on the first type of uncertainty, where an agent starts in an
uncertain state but has deterministic actions. We seek strong plans,
where the agent will reach the goal with certainty despite its
partially known state.  Many of the aforementioned planners find
strong plans, and heuristic search planners are currently among the
best.  Yet a foundation for what constitutes a good distance-based
heuristic for belief space has not been adequately investigated.

\und{Belief Space Heuristics} Intuitively, it can be argued that the
heuristic merit of a belief state depends on at least two
factors--the size of the belief state (i.e., the uncertainty in the
current state), and the distance of the individual states in the
belief state from a destination belief state.  The question of
course is how to compute these measures and which are most
effective.  Many approaches estimate belief state distances in terms
of individual state to state distances between states in two belief
states, but either lack effective state to state distances or ways
to aggregate the state distances. For instance the MBP planner
\citep{bertoli01planning} counts the number of states in the current
belief state.  This amounts to assuming each state distance has unit
cost, and planning for each state can be done independently. The GPT
planner \citep{bonet00planning} measures the state to state
distances exactly and takes the maximum distance, assuming the
states of the belief state positively interact.

\und{Heuristic Computation Substrates} We characterize several
approaches to estimating belief state distance by describing them in
terms of underlying state to state distances.  The basis of our
investigation is in adapting classical planning reachability
heuristics to measure state distances and developing state distance
aggregation techniques to measure interaction between plans for
states in a belief state.  We take three fundamental approaches to
measure the distance between two belief states. The first approach
does not involve aggregating state distance measures, rather we use
a classical planning graph to compute a representative state
distance.  The second retains distinctions between individual states
in the belief state by using multiple planning graphs, akin to CGP
\citep{AAAI98_IAAI98*889}, to compute many state distance measures
which are then aggregated. The third employs a new planning graph
generalization, called the Labelled Uncertainty Graph ($LUG$), that
blends the first two to measure a single distance between two belief
states. With each of these techniques we will discuss the types of
heuristics that we can compute with special emphasis on relaxed
plans. We present several relaxed plan heuristics that differ in
terms of how they employ state distance aggregation to make stronger
assumptions about how states in a belief state can co-achieve the
goal through action sequences that are independent, positively
interact, or negatively interact.

Our motivation for the first of the three planning graph techniques
for measuring belief state distances is to try a minimal extension
to classical planning heuristics to see if they will work for us.
Noticing that our use of classical planning heuristics ignores
distinctions between states in a belief state and may provide
uninformed heuristics, we move to the second approach where we
possibly build exponentially many planning graphs to get a better
heuristic. With the multiple planning graphs we extract a heuristic
from each graph and aggregate them to get the belief state distance
measure. If we assume the states of a belief state are independent,
we can aggregate the measures with a summation.  Or, if we assume
they positively interact we can use a maximization.  However, as we
will show, relaxed plans give us a unique opportunity to measure
both positive interaction and independence among the states by
essentially taking the union of several relaxed plans. Moreover,
mutexes play a role in measuring negative interactions between
states. Despite the utility of having robust ways to aggregate state
distances, we are still faced with the exponential blow up in the
number of planning graphs needed. Thus, our third approach seeks to
retain the ability to measure the interaction of state distances but
avoid computing multiple graphs and extracting heuristics from each.
The idea is to condense and symbolically represent multiple planning
graphs in a single planning graph, called a Labelled Uncertainty
Graph ($LUG$). Loosely speaking, this single graph unions the causal
support information present in the multiple graphs and pushes the
disjunction, describing sets of possible worlds (i.e., initial
literal layers), into ``labels". The planning graph vertices are the
same as those present in multiple graphs, but redundant
representation is avoided. For instance an action that was present
in all of the multiple planning graphs would be present only once in
the $LUG$ and labelled to indicate that it is applicable in a
planning graph projection from each possible world. We will describe
how to extract heuristics from the $LUG$ that make implicit
assumptions about state interaction without explicitly aggregating
several state distances.

Ideally, each of the planning graph techniques considers every state
in a belief state to compute heuristics, but as belief states grow
in size this could become uninformed or costly. For example, the
single classical planning graph ignores distinctions between
possible states where the heuristic based on multiple graphs leads
to the construction of a planning graph for each state.  One way to
keep costs down is to base the heuristics on only a subset of the
states in our belief state.  We evaluate the effect of such a
sampling on the cost of our heuristics.  With a single graph we
sample a single state and with multiple graphs and the $LUG$ we
sample some percent of the states. We evaluate state sampling to
show when it is appropriate, and find that it is dependent on how we
compute heuristics with the states.

\und{Standardized Evaluation of Heuristics} An issue in evaluating
the effectiveness of heuristic techniques is the many architectural
differences between planners that use the heuristics. It is quite
hard to pinpoint the global effect of the assumptions underlying
their heuristics on performance. For example, GPT is outperformed by
MBP--but it is questionable as to whether the credit for this
efficiency is attributable to the differences in heuristics, or
differences in search engines (MBP uses a BDD-based search).  Our
interest in this paper is to systematically evaluate a spectrum of
approaches for computing heuristics for belief space planning. Thus
we have implemented heuristics similar to GPT and MBP and use them
to compare against our new heuristics developed around the notion of
overlap (multiple world positive interaction and independence). We
implemented the heuristics within two planners, the Conformant-{\sl
$AltAlt$} planner ($\caltalt$) and the Partially-Observable
Non-Deterministic planner ($POND$).  $POND$ does handle search with
non-deterministic actions, but for the bulk of the paper we discuss
deterministic actions.  This more general action formulation, as
pointed out by \citet{AAAI98_IAAI98*889}, can be translated into
initial state uncertainty. Alternatively, in Section 8.2 we discuss
a more direct approach to reason with non-deterministic actions in
the heuristics.

\und{External Evaluation} Although our main interest in this paper
is to evaluate the relative advantages of a spectrum of belief space
planning heuristics in a normalized setting, we also compare the
performance of the best heuristics from this work to current state
of the art conformant and conditional planners. Our empirical
studies show that planning graph based heuristics provide effective
guidance compared to cardinality heuristics as well as the
reachability heuristic used by GPT and CFF, and our planners are
competitive with BDD-based planners such as MBP and YKA, and
GraphPlan-based ones such as CGP and SGP.  We also notice that our
planners gain scalability with our heuristics and retain reasonable
quality solutions, unlike several of the planners we compare
against.

\medskip

The rest of this paper is organized as follows. We first present the
$\caltalt$ and $POND$ planners by describing their state and action
representations as well as their search algorithms.  To understand
search guidance in the planners, we then discuss appropriate
properties of heuristic measures for belief space planning.  We
follow with a description of the three planning graph substrates
used to compute heuristics.  We carry out an empirical evaluation in
the next three sections, by describing our test setup, presenting a
standardized internal comparison, and finally comparing with several
other state of the art planners. We end with related research,
discussion, prospects for future work, and various concluding
remarks.

\section{Belief Space Planners}

Our planning formulation uses regression search to find strong
conformant plans and progression search to find strong conformant
and conditional plans.  A strong plan guarantees that after a finite
number of actions executed from any of the many possible initial
states, all resulting states are goal states. Conformant plans are a
special case where the plan has no conditional plan branches, as in
classical planning. Conditional plans are a more general case where
plans are structured as a graph because they include conditional
actions (i.e. the actions have causative and observational effects).
In this presentation, we restrict conditional plans to DAGs, but
there is no conceptual reason why they cannot be general graphs. Our
plan quality metric is the maximum plan path length.

We formulate search in the space of belief states, a technique
described by \citet{bonet00planning}. The planning problem $P$ is
defined as the tuple $\langle D, BS_I, BS_G\rangle$, where $D$ is a
domain description, $BS_I$ is the initial belief state, and $BS_G$
is the goal belief state (consisting of all states satisfying the
goal).  The domain $D$ is a tuple $\langle F, A\rangle$, where $F$
is a set of fluents and $A$ is a set of actions.

\und{Logical Formula Representation} We make extensive use of
logical formulas over $F$ to represent belief states, actions, and
$LUG$ labels, so we first explain a few conventions.  We refer to
every fluent in $F$ as either a positive literal or a negative
literal, either of which is denoted by $l$.  When discussing the
literal $l$, the opposite polarity literal is denoted $\neg l$. Thus
if $l$ = $\neg$at(location1), then $\neg l$ = at(location1).  We
reserve the symbols $\perp$ and $\top$ to denote logical false and
true, respectively.  Throughout the paper we define the conjunction
of an empty set equivalent to $\top$, and the disjunction of an
empty set as $\perp$.

Logical formulas are propositional sentences comprised of literals,
disjunction, conjunction, and negation.  We refer to the set of
models of a formula $f$ as ${\cal M}(f)$. We consider the
disjunctive normal form of a logical formula $f$, $\hat{\xi}(f)$,
and the conjunctive normal form of $f$, $\kappa(f)$. The DNF is seen
as a disjunction of ``constituents'' $\hat{S}$ each of which is a
conjunction of literals. Alternatively the CNF is seen as a
conjunction of ``clauses'' $C$ each of which is a disjunction of
literals.\footnote{It is easy to see that ${\cal M}(f)$ and
$\hat{\xi}(f)$ are readily related. Specifically each constituent
contains $k$ of the $|F|$ literals, corresponding to $2^{|F|-k}$
models.} We find it useful to think of DNF and CNF represented as
sets -- a disjunctive set of constituents or a conjunctive set of
clauses.  We also refer to the complete representation $\xi(f)$ of a
formula $f$ as a DNF where every constituent -- or in this case
state $S$ -- is a model of $f$.

\und{Belief State Representation} A world state, $S$, is represented
as a complete interpretation over fluents. We also refer to states
as possible worlds. A belief state $BS$ is a set of states and is
symbolically represented as a propositional formula over $F$. A
state $S$ is in the set of states represented by a belief state $BS$
if $S \in {\cal M}(BS)$, or equivalently $S \models BS$.

For pedagogical purposes, we use the bomb and toilet with clogging
and sensing problem, BTCS, as a running example for this paper.\footnote{We are aware
of the negative publicity associated with the B\&T problems and we
do in fact handle more interesting problems with difficult
reachability and uncertainty (e.g. Logistics and Rovers), but to
simplify our discussion we choose this small problem.} BTCS is a
problem that includes two packages, one of which contains a bomb,
and there is also a toilet in which we can dunk packages to defuse
potential bombs. The goal is to disarm the bomb and the only
allowable actions are dunking a package in the toilet (DunkP1,
DunkP2), flushing the toilet after it becomes clogged from dunking
(Flush), and using a metal-detector to sense if a package contains
the bomb (DetectMetal). The fluents encoding the problem denote that
the bomb is armed (arm) or not, the bomb is in a package (inP1,
inP2) or not, and that the toilet is clogged (clog) or not. We also
consider a conformant variation on BTCS, called BTC, where there is
no DetectMetal action.

The belief state representation of the BTCS initial condition, in
clausal representation is: \smallskip

\noindent$\kappa(BS_I)$ = arm $\wedge \neg$clog $\wedge$ (inP1
$\vee$ inP2) $\wedge$ ($\neg$inP1 $\vee \neg$inP2),
\smallskip

\noindent and in constituent representation is:
\smallskip

\noindent $\hat{\xi}(BS_I)$ = (arm $\wedge \neg$ clog $\wedge$ inP1
$\wedge \neg$inP2) $\vee$ (arm $\wedge \neg$ clog $\wedge \neg$inP1
$\wedge$ inP2).

\smallskip

\noindent The goal of BTCS has the clausal and constituent
representation:
\smallskip

\noindent$\kappa(BS_G)$ = $\hat{\xi}(BS_G)$ = $\neg$arm.
\smallskip

\noindent However, the goal has the complete representation:

\smallskip
\noindent $\xi(BS_G)$ = ($\neg$arm $\wedge$ clog $\wedge$ inP1
$\wedge \neg$inP2) $\vee$ ($\neg$arm $\wedge$ clog $\wedge \neg$inP1
$\wedge$ inP2) $\vee$\\
\hspace*{1.8cm}($\neg$arm $\wedge \neg$clog $\wedge$ inP1 $\wedge
\neg$inP2) $\vee$ ($\neg$arm $\wedge \neg$clog $\wedge
\neg$inP1 $\wedge$ inP2) $\vee$\\
\hspace*{1.8cm}($\neg$arm $\wedge$ clog $\wedge \neg$inP1 $\wedge
\neg$inP2) $\vee$
($\neg$arm $\wedge$ clog $\wedge$ inP1 $\wedge$ inP2) $\vee$\\
\hspace*{1.8cm}($\neg$arm $\wedge \neg$clog $\wedge \neg$inP1
$\wedge \neg$inP2) $\vee$ ($\neg$arm $\wedge \neg$clog $\wedge$ inP1
$\wedge$ inP2).

\smallskip

\noindent The last four states (disjuncts) in the complete
representation are unreachable, but consistent with the goal
description.

\und{Action Representation} We represent actions as having both
causative and observational effects.  All actions $a$ are described
by a tuple $\langle \rho^e(a), \Phi(a), \Theta(a) \rangle$ where
$\rho^e(a)$ is an execution precondition, $\Phi(a)$ is a set of
causative effects, and $\Theta(a)$ is a set of observations. The
execution precondition, $\rho^e(a)$, is a conjunction of literals
that must hold for the action to be executable.  If an action is
executable, we apply the set of causative effects to find successor
states and then apply the observations to partition the successor
states into observational classes.

Each causative effect $\varphi^j(a) \in \Phi(a)$ is a conditional
effect of the form $\rho^j(a)$$\implies$$\varepsilon^j(a)$, where
the antecedent $\rho^j(a)$ and consequent $\varepsilon^j(a)$ are
both a conjunction of literals.  We handle disjunction in
$\rho^e(a)$ or a $\rho^j(a)$ by replicating the respective action or
effect with different conditions, so with out loss of generality we
assume conjunctive preconditions.  However, we cannot split
disjunction in the effects. Disjunction in an effect amounts to
representing a set of non-deterministic outcomes.  Hence we do not
allow disjunction in effects thereby restricting to deterministic
effects. By convention $\varphi^0(a)$ is an unconditional effect,
which is equivalent to a conditional effect where $\rho^0(a) =
\top$.

The only way to obtain observations is to execute an action with
observations.  Each observation formula $o^j(a) \in \Theta(a)$ is a
possible sensor reading.  For example, an action $a$ that observes
the truth values of two fluents $p$ and $q$ defines $\Theta(a) = \{p
\wedge q, \neg p \wedge q, p \wedge \neg q, \neg p \wedge \neg q\}$.
This differs slightly from the conventional description of
observations in the conditional planning literature. Some works
(e.g., \citeR{Rintanen03}) describe an observation as a list of
observable formulas, then define possible sensor readings as all
boolean combinations of the formulas.  We directly define the
possible sensor readings, as illustrated by our example.  We note
that our convention is helpful in problems where some boolean
combinations of observable formulas will never be sensor readings.

\medskip

The causative and sensory actions for the example  BTCS problem are:

\smallskip

\noindent DunkP1: $\langle \rho^e$ = $\neg$clog, $\Phi = \{\varphi^0
=$ clog, $\varphi^1 =$ inP1 $\implies \neg$arm$ \}, \Theta =
\{\}\rangle$,

\noindent DunkP2: $\langle \rho^e = \neg$clog, $\Phi = \{\varphi^0
=$ clog, $\varphi^1 =$ inP2 $\implies \neg$arm$\}, \Theta =
\{\}\rangle$,

\noindent Flush: $\langle \rho^e = \top$, $\Phi = \{\varphi^0 =
\neg$clog$\}, \Theta = \{\}\rangle$, and

\noindent DetectMetal: $\langle  \rho^e = \top, \Phi = \emptyset,
\Theta = \{o^0 =$ inP1, $o^1 = \neg$inP1$\}\rangle$.

\subsection{Regression}

We perform regression in the \caltalt\ planner to find conformant
plans by starting with the goal belief state and regressing it
non-deterministically over all relevant actions. An action (without
observations) is relevant for regressing a belief state if (i) its
unconditional effect is consistent with every state in the belief
state and (ii) at least one effect consequent contains a literal
that is present in a constituent of the belief state. The first part
of relevance requires that every state in the successor belief state
is actually reachable from the predecessor belief state and the
second ensures that the action helps support the successor.

Following \citet{Pednault87Synthesizing}, regressing a belief state
$BS$ over an action $a$, with conditional effects, involves finding
the execution, causation, and preservation formulas. We define
regression in terms of clausal representation, but it can be
generalized for arbitrary formulas. The regression of a belief state
is a conjunction of the regression of clauses in $\kappa(BS)$.
Formally, the result $BS'$ of regressing the belief state $BS$ over
the action $a$ is defined as:\footnote{Note that $BS'$ may not be in
clausal form after regression (especially when an action has
multiple conditional effects).}
\begin{equation}\label{regress}\notag
 BS' = {\tt Regress}(BS, a) = \Pi(a) \wedge
\left(\bigwedge_{C\in{\kappa(BS})}\; \bigvee_{l \in C}
\left(\Sigma(a, l) \wedge IP(a,{l})\right)\right)
\end{equation}

\noindent {\bf Execution formula} ($\Pi({a})$) is the execution
precondition $\rho^e(a)$. This is what must hold in $BS'$ for $a$ to
have been applicable.

\noindent {\bf Causation formula} ($\Sigma(a,{l})$) for a literal
$l$ w.r.t all effects $\varphi^i(a)$ of an action $a$ is defined as
the weakest formula that must hold in the state before $a$ such that
$l$ holds in $BS$.  The intuitive meaning is that $l$ already held
in $BS'$, or the antecedent $\rho^i(a)$ must have held in $BS'$ to
make $l$ hold in $BS$.  Formally $\Sigma(a,l)$ is defined as:
\begin{equation}\label{cause}\notag
\Sigma(a,{l}) = l \vee \bigvee_{\substack{i : l \in
\varepsilon^{i}(a)}} \rho^i(a)
\end{equation}

\noindent {\bf Preservation formula} ($IP(a,{l})$) of a literal $l$
w.r.t. all effects $\varphi^i(a)$ of action $a$ is defined as the
formula that must be true before $a$ such that $l$ is not violated
by any effect $\varepsilon^i(a)$.  The intuitive meaning is that the
antecedent of every effect that is inconsistent with $l$ could not
have held in $BS'$.  Formally $IP(a,{l})$ is defined as:
\begin{equation}\label{pres}\notag
IP(a,{l}) = \bigwedge_{\substack{i : \neg l \in \varepsilon^{i}(a)
}} \neg\rho^i(a)
\end{equation}

Regression has also been formalized in the MBP planner
\citep{cimatti00conformant} as a symbolic pre-image computation of
BDDs \citep{bryant-ieeetc86}. While our formulation is syntactically
different, both approaches compute the same result.

\subsection{$\caltalt$}

The \caltalt\ planner uses the regression operator to generate
children in an A* search.  Regression terminates when search node
expansion generates a belief state $BS$ which is logically entailed
by the initial belief state $BS_I$.  The plan is the sequence of
actions regressed from $BS_G$ to obtain the belief state entailed by
$BS_I$.

\begin{figure}
\center\scalebox{.4}{ \includegraphics{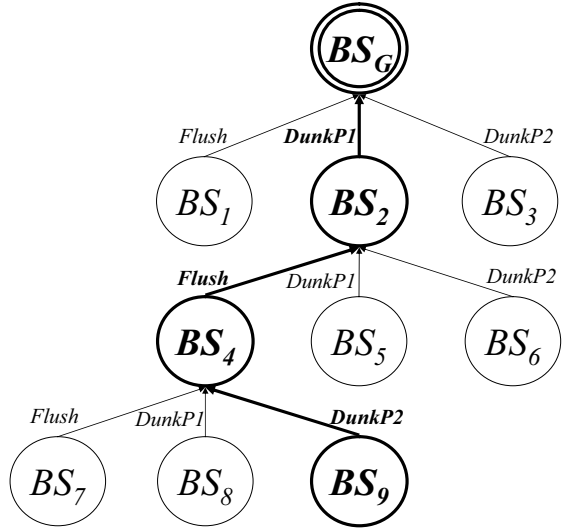}}
\caption{\label{btcregex} Illustration of the regression search
path for a conformant plan in the $BTC$ problem. }
\end{figure}

For example, in the BTC problem, Figure \ref{btcregex}, we have:

\smallskip

\noindent$BS_2 = $Regress$(BS_G,$ DunkP1) = $\neg$clog $\wedge$
($\neg$arm $\vee$ inP1).

\smallskip

\noindent The first clause is the execution formula and the second
clause is the causation formula for the conditional effect of DunkP1
and $\neg$arm.

Regressing $BS_2$ with Flush gives:

\smallskip

\noindent$BS_4 = $ Regress$(BS_2,$ Flush$) = $ ($\neg$arm $\vee$
inP1).

\smallskip

\noindent For $BS_4$, the execution precondition of Flush is $\top$,
the causation formula is $\top \vee \neg$clog $ = \top$, and
($\neg$arm $\vee$ inP1) comes by persistence of the causation
formula.

Finally, regressing $BS_4$ with DunkP2 gives:

\smallskip

\noindent $BS_9 = $ Regress$(BS_4,$ DunkP2) = $\neg$clog $\wedge$
($\neg$arm $\vee$ inP1 $\vee$ inP2).

\smallskip

\noindent We terminate at $BS_9$ because $BS_I \models BS_9$.  The
plan is DunkP2, Flush, DunkP1.

\subsection{Progression}

In progression we can handle both causative effects and
observations, so in general, progressing the action $a$ over the
belief state $BS$ generates the set of successor belief states $B$.
The set of belief states $B$ is empty when the action is not
applicable to $BS$ ($BS \not\models \rho^e(a)$).

Progression of a belief state $BS$ over an action $a$ is best
understood as the union of the result of applying $a$ to each model
of $BS$ but we in fact implement it as BDD images, as in the MBP
planner \citep{bertoli01planning}.  Since we compute progression in
two steps, first finding a causative successor, and second
partitioning the successor into observational classes, we explain
the steps separately.  The causative successor $BS'$ is found by
progressing the belief state $BS$ over the causative effects of the
action $a$. If the action is applicable, the causative successor is
the disjunction of causative progression (Progress$_c$) for each
state in $BS$ over $a$:
\begin{equation}\label{progress}\notag
    BS' = {\tt Progress}_c(BS, a) = \left\{
    \begin{array}
    {c @{\quad:\quad}l}
                                \perp & BS \not\models \rho^e(a)\\
                                \bigvee\limits_{S \in {\cal
    M}(BS)} {\tt Progress}_c(S, a) & {\tt otherwise}\\
                                \end{array}
     \right.
\end{equation}

The progression of an action $a$ over a state $S$ is the conjunction
of every literal that persists (no applicable effect consequent
contains the negation of the literal) and every literal that is
given as an effect (an applicable effect consequent contains the
literal).
\begin{equation}\label{prog}\notag
     S' = {\tt Progress}_c(S, a) = \bigwedge\limits_{\substack{l:l \in S \;{\tt and}\\
     \neg \exists_j\; S \models \rho^j(a) \;{\tt and}\\ \neg l \in \varepsilon^j(a) }}
    l \wedge \bigwedge\limits_{\substack{l: \exists_j\; S \models \rho^j(a)  \;{\tt and}
    \\  l \in \varepsilon^j(a)}} l
\end{equation}

Applying the observations of an action results in the set of
successors $B$.  The set is found (in Progress$_s$) by individually
taking the conjunction of each sensor reading $o^j(a)$ with the
causative successor $BS'$.  Applying the observations $\Theta(a)$ to
a belief state $BS'$ results in a set $B$ of belief states, defined
as:
\begin{equation}\label{progressions}\notag
 B = {\tt Progress}_s(BS', a) = \left\{
    \begin{array}{l @{\quad:\quad}l}
                                \perp & BS' = \perp \\
                                \{BS'\} & \Theta(a) = \emptyset\\
                                \{BS'' | BS'' = o^j(a) \wedge BS'\}& {\tt otherwise}\\
                                \end{array}
     \right.\\
\end{equation}

\noindent The full progression is computed as:

\medskip

$B$ = Progress($BS, a$) = Progress$_s($Progress$_c(BS, a),a)$.

\subsection{$POND$}

We use top down AO* search \citep{Nilson}, in the $POND$ planner to
generate conformant and conditional plans.  In the search graph, the
nodes are belief states and the hyper-edges are actions. We need AO*
because applying an action with observations to a belief state
divides the belief state into observational classes.  We use
hyper-edges for actions because actions with observations have
several possible successor belief states, all of which must be
included in a solution.

The AO* search consists of two repeated steps: expand the current
partial solution, and then revise the current partial solution.
Search ends when every leaf node of the current solution is a belief
state that satisfies the goal and no better solution exists (given
our heuristic function).  Expansion involves following the current
solution to an unexpanded leaf node and generating its children.
Revision is a dynamic programming update at each node in the current
solution that selects a best hyper-edge (action). The update assigns
the action with minimum cost to start the best solution rooted at
the given node. The cost of a node is the cost of its best action
plus the average cost of its children (the nodes connected through
the best action). When expanding a leaf node, the children of all
applied actions are given a heuristic value to indicate their
estimated cost.

The main differences between our formulation of AO* and that of
\citet{Nilson} are that we do not allow cycles in the search graph,
we update the costs of nodes with an average rather than a
summation, and use a weighted estimate of future cost. The first
difference is to ensure that plans are strong (there are a finite
number of steps to the goal), the second is to guide search toward
plans with lower average path cost, and the third is to bias our
search to trust the heuristic function. We define our plan quality
metric (maximum plan path length) differently than the metric our
search minimizes for two reasons. First, it is easier to compare to
other competing planners because they measure the same plan quality
metric. Second, search tends to be more efficient using the average
instead of the maximum cost of an action's children.  By using
average instead of maximum, the measured cost of a plan is lower --
this means that we are likely to search a shallower search graph to
prove a solution is not the best solution.

\medskip

\begin{figure}[t]
\center{\scalebox{.4}{
\includegraphics{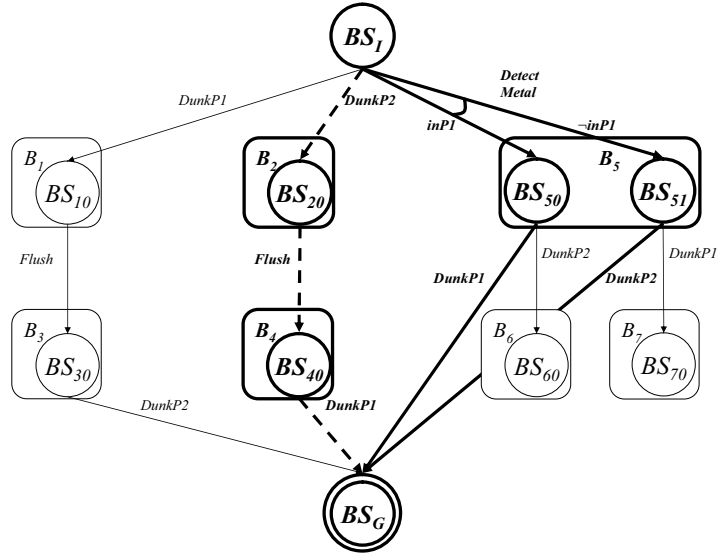}}}
\caption{\label{btcprogex} Illustration of progression search for a
conformant plan (bold dashed edges) and a conditional plan (bold
solid edges) in the BTCS problem. }
\end{figure}

Conformant planning, using actions without observations, is a
special case for AO* search, which is similar to A* search.  The
hyper-edges that represent actions are singletons, leading to a
single successor belief state. Consider the BTC problem (BTCS
without the DetectMetal action) with the future cost (heuristic
value) set to zero for every search node.  We show the search graph
in Figure \ref{btcprogex} for this conformant example as well as a
conditional example, described shortly. We can expand the initial
belief state by progressing it over all applicable actions. We get:

\smallskip

\noindent $B_1 = \{BS_{10}\} =$ Progress($BS_I,$ DunkP1)
\\\hspace*{.6cm}$=$ \{(inP1 $\wedge \neg$inP2 $\wedge$ clog $\wedge
\neg$arm) $\vee$ ($\neg$inP1 $\wedge$ inP2 $\wedge$ clog $\wedge$
arm)\}

\smallskip

\noindent and

\smallskip

\noindent $B_3 = \{BS_{20}\} =$ Progress($BS_I,$ DunkP2)
\\\hspace*{.6cm}$=$ \{(inP1 $\wedge \neg$inP2 $\wedge$ clog $\wedge$
arm) $\vee$ ($\neg$inP1 $\wedge$ inP2 $\wedge$ clog $\wedge
\neg$arm)\}.

\smallskip

\noindent  Since $\neg$clog already holds in every state of the
initial belief state, applying Flush to $BS_I$ leads to $BS_I$
creating a cycle. Hence, a hyper-edge for Flush is not added to the
search graph for $BS_I$. We assign a cost of zero to $BS_{10}$ and
$BS_{20}$, update the internal nodes of our best solution, and add
DunkP1 to the best solution rooted at $BS_I$ (whose cost is now
one).

We expand the leaf nodes of our best solution, a single node
$BS_{10}$, with all applicable actions.  The only applicable action
is Flush, so we get:

\smallskip

\noindent  $B_3 = \{BS_{30}\} =$ Progress($BS_{10}$, Flush)
\\\hspace*{.6cm}$=$ \{(inP1 $\wedge \neg$inP2 $\wedge\neg$clog
$\wedge \neg$arm) $\vee$ ($\neg$inP1 $\wedge$ inP2 $\wedge\neg$clog
$\wedge$ arm)\}.

\smallskip

\noindent  We assign a cost of zero to $BS_{30}$ and update our best
solution. We choose Flush as the best action for $BS_{10}$ (whose
cost is now one), and choose DunkP2 as the best action for $BS_I$
(whose cost is now one). DunkP2 is chosen for $BS_I$ because its
successor $BS_{20}$ has a cost of zero, as opposed to $BS_{10}$
which now has a cost of one.

Expanding the leaf node $BS_{20}$ with the only applicable action,
Flush, we get:

\smallskip

\noindent  $B_4 = \{BS_{40}\} =$ Progress($BS_{20}$, Flush)
\\\hspace*{.6cm}$=$ \{($\neg$inP1 $\wedge$ inP2 $\wedge\neg$clog
$\wedge$arm) $\vee$ (inP1 $\wedge \neg$inP2 $\wedge\neg$clog
$\wedge\neg$ arm)\}.

\smallskip

\noindent  We update $BS_{40}$ (to have cost zero) and $BS_{20}$ (to
have a cost of one), and choose Flush as the best action for
$BS_{20}$. The root node $BS_I$ has two children, each with cost
one, so we arbitrarily choose DunkP1 as the best action.

We expand $BS_{30}$ with the relevant actions to get $BS_G$ with the
DunkP2 action.  DunkP1 creates a cycle back to $BS_{10}$ so it is
not added to the search graph.  We now have a solution where all
leaf nodes are terminal. While it is only required that a terminal
belief state contains a subset of the states in $BS_G$, in this case
the terminal belief state contains exactly the states in $BS_G$. The
cost of the solution is three because, through revision, $BS_{30}$
has a cost of one, which sets $BS_{10}$ to a cost of two. However,
this means now that $BS_I$ has cost of three if its best action is
DunkP1.  Instead, revision sets the best action for $BS_I$ to DunkP2
because its cost is currently two.

We then expand $BS_{40}$ with DunkP1 to find that its successor is
$BS_G$.  DunkP2 creates a cycle back to $BS_{20}$ so it is not added
to the search graph.  We now have our second valid solution because
it contains no unexpanded leaf nodes.  Revision sets the cost of
$BS_{40}$ to one, $BS_{20}$ to two, and $BS_I$ to three.  Since all
solutions starting at $BS_I$ have equal cost (meaning there are now
cheaper solutions), we can terminate with the plan DunkP2, Flush,
DunkP1, shown in bold with dashed lines in Figure \ref{btcprogex}.

\medskip

As an example of search for a conditional plan in $POND$, consider
the BTCS example whose search graph is also shown in Figure
\ref{btcprogex}. Expanding the initial belief state, we get:

\smallskip

\noindent  $B_1 = \{BS_{10}\} =$ Progress($BS_I,$ DunkP1),

\smallskip

\noindent $B_2 = \{BS_{20}\} =$ Progress($BS_I,$ DunkP2),

\smallskip

\noindent  and

\smallskip

\noindent  $B_5 = \{BS_{50}, BS_{51}\}$ = Progress$(BS_I,
$DetectMetal)
\\\hspace*{.6cm}$=$ $\{$inP1 $\wedge \neg$inP2 $\wedge \neg$clog
$\wedge$ arm, $\neg$inP1 $\wedge$ inP2 $\wedge \neg$clog $\wedge$
arm$\}$.

\smallskip

\noindent  Each of the leaf nodes is assigned a cost of zero, and
DunkP1 is chosen arbitrarily for the best solution rooted at $BS_I$
because the cost of each solution is identical. The cost of
including each hyper-edge is the average cost of its children plus
its cost, so the cost of using DetectMetal is (0+0)/2 + 1 = 1. Thus,
our root $BS_I$ has a cost of one.

As in the conformant problem we expand $BS_{10}$, giving its child a
cost of zero and $BS_{10}$ a cost of one. This changes our best
solution at $BS_I$ to use DunkP2, and we expand $BS_{20}$, giving
its child a cost of zero and it a cost of one. Then we choose
DetectMetal to start the best solution at $BS_I$ because it gives
$BS_I$ a cost of one, where using either Dunk action would give
$BS_I$ a cost of two.

We expand the first child of DetectMetal, $BS_{50}$, with DunkP1 to
get:

\smallskip

\noindent  \{inP1 $\wedge \neg$inP2 $\wedge$ clog $\wedge
\neg$arm\},

\smallskip

\noindent  which is a goal state, and DunkP2 to get:

\smallskip

\noindent  $B_6 = \{BS_{60}\}$ = Progress$(BS_{50}, $DunkP2) =
\{inP1 $\wedge \neg$inP2 $\wedge $clog $\wedge$ arm\}.

\smallskip

\noindent  We then expand the second child, $BS_{51}$, with DunkP2
to get:

\smallskip

\noindent  \{$\neg$inP1 $\wedge$ inP2 $\wedge$ clog $\wedge
\neg$arm\},

\smallskip

\noindent  which is also a goal state and DunkP1 to get:

\smallskip

\noindent  $B_7 = \{BS_{70}\}$ = Progress$(BS_{51}, $DunkP1) =
\{$\neg$inP1 $\wedge$ inP2 $\wedge $clog $\wedge$ arm\}.

\smallskip

\noindent  While none of these new belief states are not equivalent
to $BS_G$, two of them entail $BS_G$, so we can treat them as
terminal by connecting the hyper-edges for these actions to $BS_G$.
We choose DunkP1 and DunkP2 as best actions for $BS_{50}$ and
$BS_{51}$ respectively and set the cost of each node to one. This in
turn sets the cost of using DetectMetal for $BS_I$ to (1+1)/2 + 1 =
2. We terminate here because this plan has cost equal to the other
possible plans starting at $BS_I$ and all leaf nodes satisfy the
goal. The plan is shown in bold with solid lines in Figure
\ref{btcprogex}.

\section{Belief State Distance}

In both the \caltalt\ and $POND$ planners we need to guide search
node expansion with heuristics that estimate the plan distance
$dist(BS, BS')$ between two belief states $BS$ and $BS'$.  By
convention, we assume $BS$ precedes $BS'$ (i.e., in progression $BS$
is a search node and $BS'$ is the goal belief state, or in
regression $BS$ is the initial belief state and $BS'$ is a search
node).  For simplicity, we limit our discussion to progression
planning.  Since a strong plan (executed in $BS$) ensures that every
state $S \in {\cal M}(BS)$ will transition to some state $S' \in
{\cal M}(BS')$, we define the plan distance between $BS$ and $BS'$
as the number of actions needed to transition every state $S \in
{\cal M}(BS)$ to a state $S' \in {\cal M}(BS')$. Naturally, in a
strong plan, the actions used to transition a state $S_1 \in {\cal
M}(BS)$ may affect how we transition another state $S_2 \in {\cal
M}(BS)$. There is usually some degree of positive or negative
interaction between $S_1$ and $S_2$ that can be ignored or captured
in estimating plan distance.\footnote{Interaction between states
captures the notion that actions performed to transition one state
to the goal may interfere (negatively interact) or aid with
(positively interact) transitioning other states to goals states.}
In the following we explore how to perform such estimates by using
several intuitions from classical planning state distance
heuristics.

\begin{figure*}[t]
\center{\scalebox{.5}{
\includegraphics{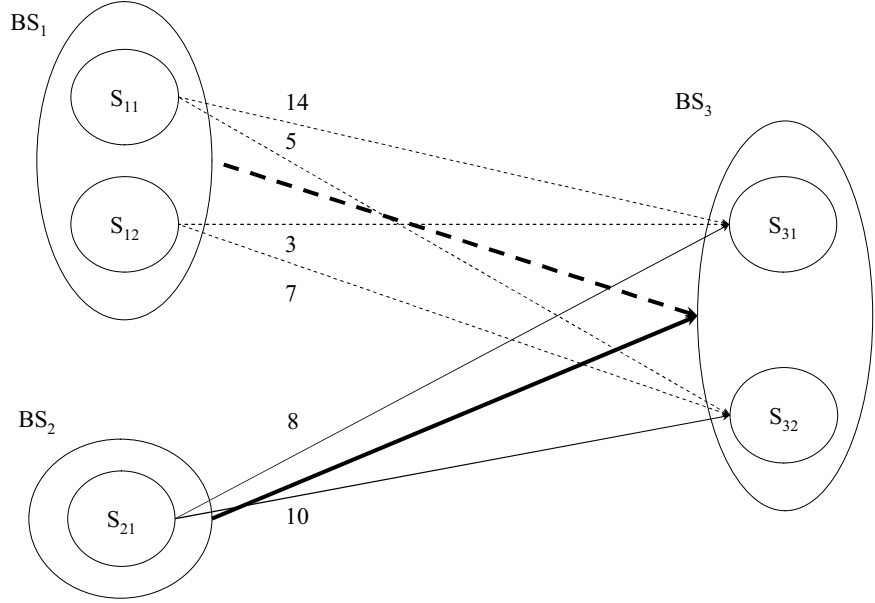}}} \caption[Heuristic Characteristics
Example]{\label{heur_charac} Conformant Plan Distance Estimation in
Belief Space}
\end{figure*}

We start with an example search scenario in Figure
\ref{heur_charac}.  There are three belief states $BS_1$ (containing
states $S_{11}$ and $S_{12}$), $BS_2$ (containing state $S_{21}$),
and $BS_3$ (containing states $S_{31}$ and $S_{32}$). The goal
belief state is $BS_3$, and the two progression search nodes are
$BS_1$ and $BS_2$. We want to expand the search node with the
smallest distance to $BS_3$ by estimating $dist(BS_1, BS_3)$ --
denoted by the bold, dashed line -- and $dist(BS_2, BS_3)$ --
denoted by the bold, solid line. We will assume for now that we have
estimates of state distance measures $dist(S, S')$ -- denoted by the
light dashed and solid lines with numbers.  The state distances can
be represented as numbers or action sequences.  In our example, we
will use the following action sequences for illustration:

\smallskip

\noindent $dist(S_{11}, S_{32}): (\{a_1, a_2\}, \{a_5\}, \{a_6,
a_7\})$,
\smallskip

\noindent $dist(S_{12}, S_{31}): (\{a_1, a_7\}, \{a_3\})$,
\smallskip

\noindent $dist(S_{21}, S_{31}): (\{a_3, a_6\}, \{a_9, a_2, a_1\},
\{a_0, a_8\}, \{a_5\})$.

\smallskip

\noindent In each sequence there may be several actions in each
step.  For instance, $dist(S_{21}, S_{31})$ has $a_3$ and $a_6$ in
its first step, and there are a total of eight actions in the
sequence -- meaning the distance is eight.  Notice that our example
includes several state distance estimates, which can be found with
classical planning techniques.  There are many ways that we can use
similar ideas to estimate belief state distance once we have
addressed the issue of belief states containing several states.

\und{Selecting States for Distance Estimation} There exists a
considerable body of literature on estimating the plan distance
between states in classical planning
\citep{bonet99planning,nguyen02planning,hoffmann:nebel:jair-01}, and
we would like to apply it to estimate the plan distance between two
belief states, say $BS_1$ and $BS_3$.  We identify four possible
options for using state distance estimates to compute the distance
between belief states $BS_1$ and $BS_3$:
\begin{itemize}
  \item Sample a State Pair: We can sample a single state from $BS_1$ and a single state
  from $BS_3$, whose plan distance is used for the belief state
  distance.  For example, we might sample $S_{12}$ from $BS_1$ and
  $S_{31}$ from $BS_3$, then define $dist(BS_1, BS_3) = dist(S_{12}, S_{31})$.

  \item Aggregate States: We can form aggregate states for $BS_1$
  and $BS_3$ and measure their plan distance. An aggregate state is the union of the literals
needed to express a belief state formula, which we define as:
\begin{equation}\notag
\tilde{S}(BS) = \bigcup\limits_{\substack{l: l \in \hat{S}, \hat{S}
\in \hat{\xi}(BS)}} l
\end{equation}

\noindent  Since it is possible to express a belief state formula
with every literal (e.g., using $(q \vee \neg q) \wedge p$ to
express the belief state where $p$ is true), we assume a reasonably
succinct representation, such as a ROBDD \citep{bryant-ieeetc86}. It
is quite possible the aggregate states are inconsistent, but many
classical planning techniques (such as planning graphs) do not
require consistent states.  For example, with aggregate states we
would compute the belief state distance $dist(BS_1, BS_3) =
dist(\tilde{S}(BS_1), \tilde{S}(BS_3))$.

  \item Choose a Subset of States: We can choose a set of states
  (e.g., by random sampling)
  from $BS_1$ and a set of states from $BS_3$, and then compute state
  distances for all pairs of states from the sets.  Upon computing all state distances, we can
  aggregate the state distances (as we will describe shortly).  For example, we might sample
  both $S_{11}$ and $S_{12}$ from $BS_1$ and $S_{31}$ from $BS_3$,
  compute $dist(S_{11}, S_{31})$ and $dist(S_{12}, S_{31})$, and
  then aggregate the state distances to define $dist(BS_1, BS_3)$.

  \item Use All States:  We can use all states in $BS_1$ and $BS_3$,
  and, similar to sampling a subset of states (above), we can compute all distances for state pairs and aggregate the distances.
\end{itemize}

The former two options for computing belief state distance are
reasonably straightforward, given the existing work in classical
planning.  In the latter two options we compute multiple state
distances.  With multiple state distances there are two details
which require consideration in order to obtain a belief state
distance measure.  In the following we treat belief states as if
they contain all states because they can be appropriately replaced
with the subset of chosen states.

The first issue is that some of the state distances may not be
needed.  Since each  state in $BS_1$ needs to reach a  state in
$BS_3$, we should consider the distance for each state in $BS_1$ to
``a'' state in $BS_3$. However, we don't necessarily need the
distance for every state in $BS_1$ to ``every'' state in $BS_3$. We
will explore assumptions about which state distances need to be
computed in Section 3.1.

The second issue, which arises after computing the state distances,
is that we need to aggregate the state distances into a belief state
distance.  We notice that the popular state distance estimates used
in classical planning typically measure aggregate costs of state
features (literals).  Since we are planning in belief space, we wish
to estimate belief state distance with the aggregate cost of belief
state features (states).  In Section 3.2, we will examine several
choices for aggregating state distances and discuss how each
captures different types of state interaction.  In Section 3.3, we
conclude with a summary of the choices we make in order to compute
belief state distances.

\subsection{State Distance Assumptions}

When we choose to compute multiple state distances between two
belief states $BS$ and $BS'$, whether by considering all states or
sampling subsets, not all of the state distances are important. For
a given state in $BS$ we do not need to know the distance to every
state in $BS'$ because each state in $BS$ need only transition to
one state in $BS'$.  There are two assumptions that we can make
about the states reached in $BS'$ which help us define two different
belief state distance measures in terms of aggregate state
distances:

\begin{itemize}
  \item We can optimistically assume that each of the
earlier states $S \in {\cal M}(BS)$ can reach the closest of the
later states $S' \in {\cal M}(BS')$. With this assumption we compute
distance as:

\smallskip

$dist(BS, BS') = \bigtriangledown_{S\in {\cal M}(BS)}
\min\limits_{S' \in {\cal M}(BS')} dist(S, S')$.

  \item We can assume that all of the earlier states $S \in {\cal M}(BS)$ reach the
  same later state $S' \in {\cal M}(BS')$, where the aggregate distance is
minimum. With this assumption we compute distance as:

\smallskip

$dist(BS, BS') = \min\limits_{S' \in {\cal M}(BS')}
\bigtriangledown_{S\in {\cal M}(BS)} dist(S, S')$,

\end{itemize}

\noindent where $\bigtriangledown$ represents an aggregation
technique (several of which we will discuss shortly).

Throughout the rest of the paper we use the first definition for
belief state distance because it is relatively robust and easy to
compute.  Its only drawback is that it treats the earlier states in
a more independent fashion, but is flexible in allowing earlier
states to transition to different later states.  The second
definition measures more dependencies of the earlier states, but
restricts them to reach the same later state.  While the second may
sometimes be more accurate, it is misinformed in cases where all
earlier states cannot reach the same later state (i.e., the measure
would be infinite). We do not pursue the second method because it
may return distance measures that are infinite when they are in fact
finite.

As we will see in Section 4, when we discuss computing these
measures with planning graphs, we can implicitly find for each state
in $BS$ the closest state in $BS'$, so that we do not enumerate the
states $S'$ in the minimization term of the first belief state
distance (above). Part of the reason we can do this is that we
compute distance in terms of constituents $\hat{S}' \in
\hat{\xi}(BS')$ rather than actual states. Also, because we only
consider constituents of $BS'$, when we discuss sampling belief
states to include in distance computation we only sample from $BS$.
We can also avoid the explicit aggregation $\bigtriangledown$ by
using the $LUG$, but describe several choices for $\bigtriangledown$
to understand implicit assumptions made by the heuristics computed
on the $LUG$.

\subsection{State Distance Aggregation}

The aggregation function $\bigtriangledown$ plays an important role
in how we measure the distance between belief states.  When we
compute more than one state distance measure, either exhaustively or
by sampling a subset (as previously mentioned), we must combine the
measures by some means, denoted $\bigtriangledown$.  There is a
range of options for taking the state distances and aggregating them
into a belief state distance.  We discuss several assumptions
associated with potential measures:

\begin{itemize}
    \item Positive Interaction of States: Positive interaction
    assumes that the most difficult state in $BS$ requires actions
    that will help transition all other states in $BS$ to some state
    in $BS'$.  In our example, this means that we assume the actions used
    to transition $S_{11}$ to $S_{32}$ will help us transition $S_{12}$ to
    $S_{31}$ (assuming each state in $BS_1$ transitions to the closest state in $BS_3$).
    Inspecting the action sequences, we see they positively interact because both need actions
    $a_1$ and $a_7$.  We do not need to know the action sequences to assume
    positive interaction because we define the aggregation $\bigtriangledown$ as a maximization
    of numerical state distances:

\smallskip

    $dist(BS, BS') = \max\limits_{S\in {\cal M}(BS)} \min\limits_{S' \in
{\cal M}(BS')} dist(S, S')$.

\smallskip

    The belief state distances are $dist(BS_1, BS_3) = \max(\min(14, 5),$ $\min(3, 7)) =
    5$ and $dist(BS_2, BS_3) = \max(\min(8, 10)) = 8$.  In this case we
    prefer $BS_1$ to $BS_2$.  If each
    state distance is admissible and we do not sample from belief states, then assuming positive interaction
    is also admissible.

    \item Independence of States:  Independence assumes that each
    state in $BS$ requires actions that are different from all other states in $BS$
    in order to reach a state in $BS'$.  Previously, we found there was positive interaction in the action sequences
    to transition $S_{11}$ to $S_{32}$ and $S_{12}$ to
    $S_{31}$ because they shared actions $a_1$ and $a_7$.
    There is also some independence in these sequences
    because the first contains
    $a_2, a_5$, and $a_6$, where the second contains $a_3$. Again,
    we do not need to know the action sequences to assume
    independence because we define the aggregation $\bigtriangledown$
    as a summation of numerical state distances:

\smallskip

    $dist(BS, BS') = \sum\limits_{S\in {\cal M}(BS)} \min\limits_{S' \in
{\cal M}(BS')} dist(S, S')$.

\smallskip

    In our example, $dist(BS_1, BS_3) = \min(14, 5) + \min(3, 7) = 8$, and
    $dist(BS_2, BS_3) = \min(8, 10) = 8$.  In this case we have no
    preference over $BS_1$ and $BS_2$.

    We notice that using the cardinality of a belief state $|{\cal M}(BS)|$ to measure $dist(BS, BS')$ is
    a special case of assuming state independence, where
    $\forall S, S' dist(S, S') = 1$.
    If we use cardinality to measure distance in our example, then we have $dist(BS_1, BS_3) = |{\cal M}(BS_1)| = 2$,
    and $dist(BS_2, BS_3) = |{\cal M}(BS_2)| = 1$.  With cardinality we prefer
    $BS_2$ over $BS_1$ because we have better knowledge in $BS_2$.

    \item Overlap of States:  Overlap assumes that there is both
    positive interaction and independence between the actions used by states in $BS$
    to reach a state in $BS'$.  The intuition is that some actions can
    often be used for multiple states in $BS$ simultaneously and we
    should count these actions only once.  For example, when we computed
     $dist(BS_1, BS_3)$ by assuming
    positive interaction, we noticed that the action sequences for $dist(S_{11},
    S_{32})$ and $dist(S_{12}, S_{31})$ both used $a_1$ and $a_7$.  When we aggregate these sequences we would
    like to count $a_1$ and $a_7$ each only once because they potentially overlap.
    However, truly combining the action sequences for maximal overlap is a plan merging
    problem \citep{merging}, which can be as difficult as planning.
    Since our ultimate intent is to compute heuristics, we take a
    very simple approach to merging action sequences.  We introduce
    a plan merging operator $\Cup$ for $\bigtriangledown$ that picks a
    step at which we align the sequences and then unions the aligned
    steps.  We use the size of the resulting action sequence to
    measure belief state distance:

\smallskip

$dist(BS, BS') = \Cup_{S\in {\cal M}(BS)} \min\limits_{S' \in {\cal
M}(BS')} dist(S, S')$.

\smallskip

    Depending on the type of search, we define $\Cup$ differently.  We assume that sequences used in progression
    search start at the same time and those used in regression end at the same
    time. Thus, in progression all sequences are aligned at the
    first step before we union steps, and in regression all
    sequences are aligned at the last step before the union.

    For example, in progression $dist(S_{11}, S_{32}) \Cup dist(S_{12},
    S_{31})$ $= (\{a_1, a_2\}, \{a_5\}, \{a_6, a_7\}) \Cup (\{a_1, a_7\}, \{a_3\})$ $= (\{a_1, a_2, a_7\}, \{a_5, a_3\}, \{a_6,$ $a_7\})$ because we align the
    sequences at their first steps, then union each step.  Notice that this resulting sequence
    has seven actions, giving $dist(BS_1, BS_3) = 7$,
    whereas defining $\bigtriangledown$ as maximum gave a distance
    of five and as summation gave a distance of eight.  Compared with
    overlap, positive interaction tends to under estimate distance,
    and independence tends to over estimate distance.  As we
    will see during our empirical evaluation (in Section 6.5),
    accounting for overlap provides more accurate distance measures
    for many conformant planning domains.

    \item Negative Interaction of States: Negative interaction
    between states can appear in our example if transitioning state $S_{11}$ to
    state $S_{32}$ makes it more difficult (or even
    impossible) to transition state $S_{12}$ to state
    $S_{31}$.  This could happen if
    performing action $a_5$ for $S_{11}$ conflicts with action
    $a_3$ for $S_{12}$.  We can say that $BS_1$ cannot reach $BS_3$ if all possible action sequences
    that start in $S_{11}$ and $S_{12}$, respectively, and end in any $S \in {\cal M}(BS_3)$ negatively interact.

    There are two ways negative interactions play a role in belief state
    distances.  Negative interactions can allow us to prove it is
    impossible for a belief state $BS$ to reach a belief state
    $BS'$, meaning $dist(BS, BS') = \infty$, or they can potentially
    increase the distance by a finite amount.  We use only the
    first, more extreme, notion of negative interaction by computing ``cross-world'' mutexes \citep{AAAI98_IAAI98*889}
    to prune belief states from the search.  If we cannot prune a belief state, then
    we use one of the aforementioned techniques to aggregate state
    distances.  As such, we do not provide a concrete definition for $\bigtriangledown$
    to measure negative interaction.

    While we do not explore ways to adjust the distance measure for
    negative interactions, we mention some possibilities.  Like work
    in classical planning \citep{nguyen02planning}, we can penalize the distance measure $dist(BS_1, BS_3)$ to
    reflect additional cost associated with serializing conflicting actions.
    Additionally in conditional planning, conflicting actions can be conditioned on
    observations so that they do not execute in the same plan
    branch. A distance measure that uses observations would reflect the added cost of
    obtaining observations, as well as the change in cost associated with introducing
    plan branches (e.g., measuring average branch cost).

\end{itemize}

The above techniques for belief state distance estimation in terms
of state distances provide the basis for our use of multiple
planning graphs.  We will show in the empirical evaluation that
these measures affect planner performance very differently across
standard conformant and conditional planning domains.  While it can
be quite costly to compute several state distance measures,
understanding how to aggregate state distances sets the foundation
for techniques we develop in the $LUG$. As we have already
mentioned, the $LUG$ conveniently allows us to implicitly aggregate
state distances to directly measure belief state distance.

\subsection{Summary of Methods for Distance Estimation}

Since we explore several methods for computing belief state
distances on planning graphs, we provide a summary of the choices we
must consider, listed in Table \ref{features}.  Each column is
headed with a choice, containing possible options below. The order
of the columns reflects the order in which we consider the options.

\begin{table}
\center
\begin{tabular}{c | c | c | c | c | c}
State & State Distance & Planning & Mutex & Mutex  & Heuristic \\
Selection & Aggregation & Graph & Type & Worlds &  \\
\hline Single&+ Interaction&  $SG$ & None& Same& Max\\
Aggregate & Independence & $MG$ &Static &Intersect & Sum \\
Subset & Overlap & $LUG$ & Dynamic&Cross & Level \\
All& - Interaction& &Induced & & Relaxed Plan\\
\end{tabular}
\caption{\label{features} Features for a belief state distance
estimation.}
\end{table}

In this section we have covered the first two columns which relate
to selecting states from belief states for distance computation, as
well as aggregating multiple state distances into a belief state
distance.  We test options for both of these choices in the
empirical evaluation.

In the next section we will also expand upon how to aggregate
distance measures as well as discuss the remaining columns of Table
\ref{features}.  We will present each type of planning graph: the
single planning graph ($SG$), multiple planning graphs ($MG$), and
the labelled uncertainty graph ($LUG$).  Within each planning graph
we will describe several types of mutex, including static, dynamic,
and induced mutexes. Additionally, each type of mutex can be
computed with respect to different possible worlds -- which means
the mutex involves planning graph elements (e.g., actions) when they
exist in the same world (i.e., mutexes are only computed within the
planning graph for a single state), or across worlds (i.e., mutexes
are computed between planning graphs for different states) by two
methods (denoted Intersect and Cross). Finally, we can compute many
different heuristics on the planning graphs to measure state
distances -- max, sum, level, and relaxed plan. We focus our
discussion on the planning graphs, same-world mutexes, and relaxed
plan heuristics in the next section. Cross-world mutexes and the
other heuristics are described in appendices.

\section{Heuristics}

This section discusses how we can use planning graph heuristics to
measure belief state distances.  We cover several types of planning
graphs and the extent to which they can be used to compute various
heuristics.  We begin with a brief background on planning graphs.

\und{Planning Graphs} Planning graphs serve as the basis for our
belief state distance estimation.  Planning graphs were initially
introduced in GraphPlan \citep{blum95fast} for representing an
optimistic, compressed version of the state space progression tree.
The compression lies in unioning the literals from every state at
subsequent steps from the initial state.  The optimism relates to
underestimating the number of steps it takes to support sets of
literals (by tracking only a subset of the infeasible tuples of
literals). GraphPlan searches the compressed progression (or
planning graph) once it achieves the goal literals in a level with
no two goal literals marked infeasible. The search tries to find
actions to support the top level goal literals, then find actions to
support the chosen actions and so on until reaching the first graph
level.  The basic idea behind using planning graphs for search
heuristics is that we can find the first level of a planning graph
where a literal in a state appears; the index of this level is a
lower bound on the number of actions that are needed to achieve a
state with the literal. There are also techniques for estimating the
number of actions required to achieve sets of literals.  The
planning graphs serve as a way to estimate the reachability of state
literals and discriminate between the ``goodness'' of different
search states. This work generalizes such literal estimations to
belief space search by considering both GraphPlan and CGP style
planning graphs plus a new generalization of planning graphs, called
the $LUG$.

Planners such as CGP \citep{AAAI98_IAAI98*889} and SGP \citep{SGP}
adapt the GraphPlan idea of compressing the search space with a
planning graph by using multiple planning graphs, one for each
possible world in the initial belief state.  CGP and SGP search on
these planning graphs, similar to GraphPlan, to find conformant and
conditional plans. The work in this paper seeks to apply the idea of
extracting search heuristics from planning graphs, previously used
in state space search
\citep{nguyen02planning,hoffmann:nebel:jair-01,bonet99planning} to
belief space search.

\und{Planning Graphs for Belief Space} This section proceeds by
describing four classes of heuristics to estimate belief state
distance $NG, SG, MG,$ and $LUG$.  $NG$ heuristics are techniques
existing in the literature that are not based on planning graphs,
$SG$ heuristics are techniques based on a single classical planning
graph, $MG$ heuristics are techniques based on multiple planning
graphs (similar to those used in CGP) and $LUG$ heuristics use a new
labelled planning graph.  The $LUG$ combines the advantages of $SG$
and $MG$ to reduce the representation size and maintain
informedness.  Note that we do not include observations in any of
the planning graph structures as SGP \citep{SGP} would, however we
do include this feature for future work. The conditional planning
formulation directly uses the planning graph heuristics by ignoring
observations, and our results show that this still gives good
performance.

\begin{figure}[t] \center\scalebox{.5}{
\includegraphics{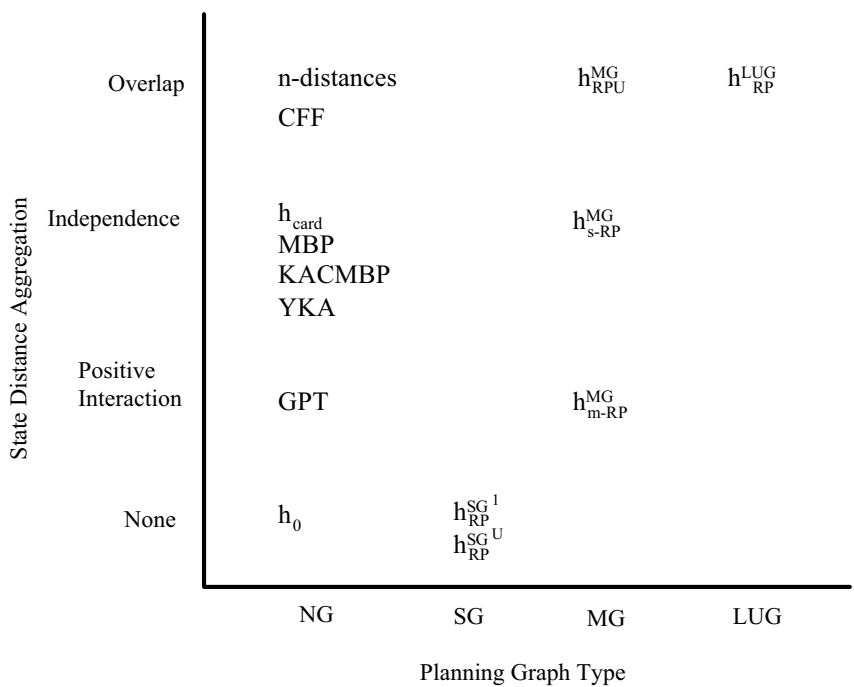}} \caption{\label{hclasses}
Taxonomy of heuristics with respect to planning graph type and
state distance aggregation.  Blank entries indicate that the
combination is meaningless or not possible.}
\end{figure}

In Figure \ref{hclasses} we present a taxonomy of distance measures
for belief space.  The taxonomy also includes related planners,
whose distance measures will be characterized in this section.  All
of the related planners are listed in the $NG$ group, despite the
fact that some actually use planning graphs, because they do not
clearly fall into one of our planning graph categories. The figure
shows how different substrates (horizontal axis) can be used to
compute belief state distance by aggregating state to state
distances under various assumptions (vertical axis).  Some of the
combinations are not considered because they do not make sense or
are impossible.  The reasons for these omissions will be discussed
in subsequent sections.  While there are a wealth of different
heuristics one can compute using planning graphs, we concentrate on
relaxed plans because they have proven to be the most effective in
classical planning and in our previous studies \citep{ICAPS04-BrKa}.
We provide additional descriptions of other heuristics like max,
sum, and level in Appendix A.

\und{Example} To illustrate the computation of each heuristic, we
use an example derived from BTC called Courteous BTC (CBTC) where a
courteous package dunker has to disarm the bomb and leave the toilet
unclogged, but some discourteous person has left the toilet clogged.
The initial belief state of CBTC in clausal representation is:

\smallskip

\noindent $\kappa(BS_I) = $ arm $\wedge$ clog $\wedge$ (inP1 $\vee$
inP2) $\wedge$ ($\neg$inP1 $\vee \neg$inP2),

\smallskip

\noindent and the goal is:

\smallskip

\noindent $\kappa(BS_G) = $ $\neg$clog $\wedge \neg$arm.

\smallskip

\noindent The optimal action sequences to reach $BS_G$ from $BS_I$
are:

\smallskip

\noindent Flush, DunkP1, Flush, DunkP2, Flush,

\smallskip

\noindent and

\smallskip

\noindent Flush, DunkP2, Flush, DunkP1, Flush.

\smallskip

\noindent Thus the optimal heuristic estimate for the distance
between $BS_I$ and $BS_G$, in regression, is $h^{*}(BS_G)$ = 5
because in either plan there are five actions.

\medskip

We use planning graphs for both progression and regression search.
In regression search the heuristic estimates the cost of the current
belief state w.r.t. the initial belief state and in progression
search the heuristic estimates the cost of the goal belief state
w.r.t. the current belief state.  Thus, in regression search the
planning graph(s) are built (projected) once from the possible
worlds of the initial belief state, but in progression search they
need to be built at each search node.  We introduce a notation
$BS_i$ to denote the belief state for which we find a heuristic
measure, and $BS_P$ to denote the belief state that is used to
construct the initial layer of the planning graph(s).  In the
following subsections we describe computing heuristics for
regression, but they are generalized for progression by changing
$BS_i$ and $BS_P$ appropriately.

In the previous section we discussed two important issues involved
in heuristic computation: sampling states to include in the
computation and using mutexes to capture negative interactions in
the heuristics.  We will not directly address these issues in this
section, deferring them to discussion in the respective empirical
evaluation sections, 6.4 and 6.2.  The heuristics below are computed
once we have decided on a set of states to use, whether by sampling
or not.  Also, as previously mentioned, we only consider sampling
states from the belief state $BS_P$ because we can implicitly find
closest states from $BS_i$ without sampling.  We only explore
computing mutexes on the planning graphs in regression search.  We
use mutexes to determine the first level of the planning graph where
the goal belief state is reachable (via the level heuristic
described in Appendix A) and then extract a relaxed plan starting at
that level. If the level heuristic is $\infty$ because there is no
level where a belief state is reachable, then we can prune the
regressed belief state.

\medskip

We proceed by describing the various substrates used for computing
belief space distance estimates.  Within each we describe the
prospects for various types of world aggregation. In addition to
our heuristics, we mention related work in the relevant areas.

\subsection{Non Planning Graph-based Heuristics ($NG$)}

We group many heuristics and planners into the $NG$ group because
they are not using $SG$, $MG$, or $LUG$ planning graphs.  Just
because we mention them in this group does not mean they are not
using planning graphs in some other form.

\und{No Aggregation} Breadth first search uses a simple heuristic,
$h_0$ where the heuristic value is set to zero.  We mention this
heuristic so that we can gauge the effectiveness of our search
substrates relative to improvements gained through using heuristics.

\und{Positive Interaction Aggregation} The GPT planner
\citep{bonet00planning} measures belief state distance as the
maximum of the minimum state to state distance of states in the
source and destination belief states, assuming optimistic
reachability as mentioned in Section 3.  GPT measures state
distances exactly, in terms of the minimum number of transitions in
the state space.  Taking the maximum state to state distance is akin
to assuming positive interaction of states in the current belief
state.

\und{Independence Aggregation} The MBP planner
\citep{bertoli01planning}, KACMBP planner \citep{bertoli02kacmbp},
YKA planner \citep{Rintanen03}, and our comparable {\bf $h_{card}$}
heuristic measure belief state distance by assuming every state to
state distance is one, and taking the summation of the state
distances (i.e. counting the number of states in a belief state).
This measure can be useful in regression because goal belief states
are partially specified and contain many states consistent with a
goal formula and many of the states consistent with the goal formula
are not reachable from the initial belief state.  Throughout
regression, many of the unreachable states are removed from
predecessor belief states because they are inconsistent with the
preconditions of a regressed action. Thus, belief states can reduce
in size during regression and their cardinality may indicate they
are closer to the initial belief state.  Cardinality is also useful
in progression because as belief states become smaller, the agent
has more knowledge and it can be easier to reach a goal state.

In CBTC, $h_{card}(BS_G) = 4$ because $BS_G$ has four states
consistent with its complete representation:

\smallskip

\noindent $\xi(BS_G) = $ ($\neg$inP1
$\wedge\neg$inP2$\wedge\neg$clog $\wedge \neg$arm) $\vee$
($\neg$inP1 $\wedge$ inP2 $\wedge \neg$clog $\wedge \neg$arm)
$\vee$\\ \hspace*{2cm}(inP1 $\wedge\neg$inP2 $\wedge \neg$clog
$\wedge \neg$arm) $\vee$ (inP1 $\wedge$ inP2 $\wedge \neg$clog
$\wedge \neg$arm).

\smallskip

\noindent Notice, this may be uninformed for $BS_G$ because two of
the states in $\xi(BS_G)$ are not reachable, like: (inP1 $\wedge$
inP2 $\wedge \neg$clog $\wedge \neg$arm).  If there are $n$
packages, then there would be $2^{n-1}$ unreachable states
represented by $\xi(BS_G)$.  Counting unreachable states may
overestimate the distance estimate because we do not need to plan
for them.  In general, in addition to the problem of counting
unreachable states, cardinality does not accurately reflect distance
measures.  For instance, MBP reverts to breadth first search in
classical planning problems because state distance may be large or
small but it still assigns a value of one.

\und{Overlap Aggregation} \citet{rintanen04} describes n-Distances
which generalize the belief state distance measure in GPT to
consider the maximum n-tuple state distance.  The measure involves,
for each n-sized tuple of states in a belief state, finding the
length of the actual plan to transition the n-tuple to the
destination belief state.  Then the maximum n-tuple distance is
taken as the distance  measure.

For example, consider a belief state with four states.  With an n
equal to two, we would define six belief states, one for each size
two subset of the four states.  For each of these belief states we
find a real plan, then take the maximum cost over these plans to
measure the distance for the original four state belief state.  When
n is one, we are computing the same measure as GPT, and when n is
equal to the size of the belief state we are directly solving the
planning problem.  While it is costly to compute this measure for
large values of n, it is very informed as it accounts for overlap
and negative interactions.

\smallskip

The CFF planner \citep{cff} uses a version of a relaxed planning
graph to extract relaxed plans.  The relaxed plans measure the cost
of supporting a set of goal literals from all states in a belief
state. In addition to the traditional notion of a relaxed planning
graph that ignores mutexes, CFF also ignores all but one antecedent
literal in conditional effects to keep their relaxed plan reasoning
tractable.  The CFF relaxed plan does capture overlap but ignores
some subgoals and all mutexes.  The way CFF ensures the goal is
supported in the relaxed problem is to encode the relaxed planning
graph as a satisfiability problem.  If the encoding is satisfiable,
the chosen number of action assignments is the distance measure.

\begin{figure}[t]
\center{\scalebox{.5}{\includegraphics{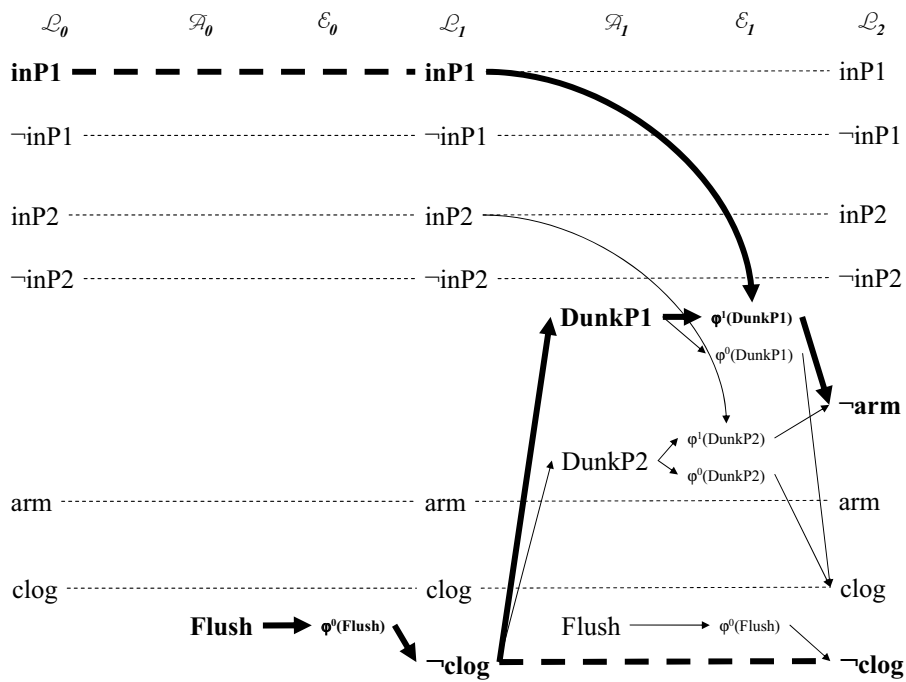}}
\caption[Single Graph $CBTC$]{\label{sgbtc} Single planning graph
for CBTC, with relaxed plan components in bold.  Mutexes omitted.}}
\end{figure}

\subsection{Single Graph Heuristics ($SG$)}

The simplest approach for using planning graphs for belief space
planning heuristics is to use a ``classical'' planning graph. To
form the initial literal layer from the projected belief state, we
could either sample a single state (denoted $SG^1$) or use an
aggregate state (denoted $SG^U$). For example, in CBTC (see Figure
\ref{sgbtc}) assuming regression search with $BS_P = BS_I$, the
initial level ${\cal L}_0$ of the planning graph for $SG^1$ might
be:

\smallskip

\noindent ${\cal L}_0$ = \{arm, clog, inP1, $\neg$inP2\}

\smallskip

\noindent and for $SG^U$ it is defined by the aggregate state
$\tilde{S}(BS_P)$:

\smallskip

\noindent ${\cal L}_0$ = \{arm, clog, inP1, inP2, $\neg$inP1,
$\neg$inP2\}.

\smallskip

\noindent Since these two versions of the single planning graph have
identical semantics, aside from the initial literal layer, we
proceed by describing the $SG^U$ graph and point out differences
with $SG^1$ where they arise.

Graph construction is identical to classical planning graphs
(including mutex propagation) and stops when two subsequent literal
layers are identical (level off).  We use the planning graph
formalism used in IPP \citep{ipp} to allow for explicit
representation of conditional effects, meaning there is a literal
layer ${\cal L}_k$, an action layer ${\cal A}_k$, and an effect
layer ${\cal E}_k$ in each level $k$.  Persistence for a literal
$l$, denoted by $l_p$, is represented as an action where
$\rho^e(l_p) = \varepsilon^{0}(l_p) = l$.  A literal is in ${\cal
L}_k$ if an effect from the previous effect layer ${\cal E}_{k-1}$
contains the literal in its consequent.  An action is in the action
layer ${\cal A}_k$ if every one of its execution precondition
literals is in ${\cal L}_k$.  An effect is in the effect layer
${\cal E}_k$ if its associated action is in the action layer ${\cal
A}_k$ and every one of its antecedent literals is in ${\cal L}_k$.
Using conditional effects in the planning graph avoids factoring an
action with conditional effects into a possibly exponential number
of non-conditional actions, but adds an extra planning graph layer
per level.  Once our graph is built, we can extract heuristics.

\und{No Aggregation}  Relaxed plans within a single planning graph
are able to measure, under the most optimistic assumptions, the
distance between two belief states.  The relaxed plan represents a
distance between a subset of the initial layer literals and the
literals in a constituent of our belief state.  In the $SG^U$, the
literals from the initial layer that are used for support may not
hold in a single state of the projected belief state, unlike the
$SG^1$. The classical relaxed plan heuristic $h^{SG}_{RP}$ finds a
set of (possibly interfering) actions to support the goal
constituent. The relaxed plan $RP$ is a subgraph of the planning
graph, of the form \{${\cal A}^{RP}_{0}$, ${\cal E}^{RP}_{0}$,
${\cal L}^{RP}_{1}$, ..., ${\cal A}^{RP}_{b-1}$, ${\cal
E}^{RP}_{b-1}$, ${\cal L}^{RP}_{b}$\}. Each of the layers contains a
subset of the vertices in the corresponding layer of the planning
graph.

More formally, we find the relaxed plan to support the constituent
$\hat{S} \in \hat{\xi}(BS_i)$ that is reached earliest in the graph
(as found by the $h^{SG}_{level}(BS_i)$ heuristic in Appendix A).
Briefly, $h^{SG}_{level}(BS_i)$ returns the first level $b$ where a
constituent of $BS_i$ has all its literals in ${\cal L}_b$ and none
are marked pair-wise mutex.  Notice that this is how we incorporate
negative interactions into our heuristics.  We start extraction at
the level $b$, by defining ${\cal L}^{RP}_{b}$ as the literals in
the constituent used in the level heuristic. For each literal $l \in
{\cal L}^{RP}_{b}$, we select a supporting effect (ignoring mutexes)
from ${\cal E}_{b-1}$ to form the subset ${\cal E}^{RP}_{b-1}$. We
prefer persistence of literals to effects in supporting literals.
Once a supporting set of effects is found, we create ${\cal
A}^{RP}_{b-1}$ as all actions with an effect in ${\cal
E}^{RP}_{b-1}$. Then the needed preconditions for the actions and
antecedents for chosen effects in ${\cal A}^{RP}_{b-1}$ and ${\cal
E}^{RP}_{b-1}$ are added to the list of literals to support from
${\cal L}^{RP}_{b-2}$.  The algorithm repeats until we find the
needed actions from ${\cal A}_{0}$. A relaxed plan's value is the
summation of the number of actions in each action layer. A literal
persistence, denoted by a subscript ``p'', is treated as an action
in the planning graph, but in a relaxed plan we do not include it in
the final computation of $\mid{{\cal A}^{RP}_{j}}\mid$. The single
graph relaxed plan heuristic is computed as
\begin{equation}\label{sgrp}\notag
h^{SG}_{RP}(BS_i) = \sum\limits_{j = 0}^{b-1} \mid{{\cal
A}^{RP}_{j}}\mid
\end{equation}

For the CBTC problem we find a relaxed plan from the $SG^U$, as
shown in Figure \ref{sgbtc} as the bold edges and nodes.  Since
$\neg$arm and $\neg$clog are non mutex at level two, we can use
persistence to support $\neg$clog and DunkP1 to support $\neg$arm in
${\cal L}^{RP}_{2}$. In ${\cal L}^{RP}_{1}$ we can use persistence
for inP1, and Flush for $\neg$clog. Thus, $h^{SG}_{RP}(BS_G)$ = 2
because the relaxed plan is:
\smallskip

${\cal A}^{RP}_{0} = \{$inP1$_p$, Flush$\}$,

\smallskip
${\cal E}^{RP}_{0} = \{\varphi^0($inP1$_p)$, $\varphi^0($Flush$)$\},

\smallskip
${\cal L}^{RP}_{1} = \{$inP1$, \neg$clog$\}$,

\smallskip
${\cal A}^{RP}_{1} = \{\neg$clog$_p$, DunkP1$\}$,

\smallskip
${\cal E}^{RP}_{1} = \{\varphi^0(\neg$clog$_p)$,
$\varphi^1($DunkP1$)$\},

\smallskip
${\cal L}^{RP}_{2} = \{\neg$arm$, \neg$clog$\}$.

\smallskip

The relaxed plan does not use both DunkP2 and DunkP1 to support
$\neg$arm.  As a result $\neg$arm is not supported in all worlds
(i.e. it is not supported when the state where inP2 holds is our
initial state). Our initial literal layer threw away knowledge of
inP1 and inP2 holding in different worlds, and the relaxed plan
extraction ignored the fact that $\neg$arm needs to be supported in
all worlds. Even with an $SG^1$ graph, we see similar behavior
because we are reasoning with only a single world.  A single,
unmodified classical planning graph cannot capture support from all
possible worlds -- hence there is no explicit aggregation over
distance measures for states.  As a result, we do not mention
aggregating states to measure positive interaction, independence, or
overlap.

\subsection{Multiple Graph Heuristics ($MG$)}

Single graph heuristics are usually uninformed because the
projected belief state $BS_P$ often corresponds to multiple
possible states. The lack of accuracy is because single graphs are
not able to capture propagation of multiple world support
information. Consider the CBTC problem where the projected belief
state is $BS_I$ and we are using a single graph $SG^U$.  If DunkP1
were the only action we would say that $\neg$arm and $\neg$clog
can be reached at a cost of two, but in fact the cost is infinite
(since there is no DunkP2 to support $\neg$arm from all possible
worlds), and there is no strong plan.

To account for lack of support in all possible worlds and sharpen
the heuristic estimate, a set of multiple planning graphs $\Gamma$
is considered. Each $\gamma \in \Gamma$ is a single graph, as
previously discussed.  These multiple graphs are similar to the
graphs used by CGP \citep{AAAI98_IAAI98*889}, but lack the more
general cross-world mutexes.  Mutexes are only computed within
each graph, i.e. only same-world mutexes are computed. We
construct the initial layer ${\cal L}^{\gamma}_{0}$ of each graph
$\gamma$ with a different state $S \in {\cal M}(BS_P)$. With
multiple graphs, the heuristic value of a belief state is computed
in terms of all the graphs.  Unlike single graphs, we can compute
different world aggregation measures with the multiple planning
graphs.

While we get a more informed heuristic by considering more of the
states in ${\cal M}(BS_P)$, in certain cases it can be costly to
compute the full set of planning graphs and extract relaxed plans.
We will describe computing the full set of planning graphs, but will
later evaluate (in Section 6.4) the effect of computing a smaller
proportion of these.  The single graph $SG^1$ is the extreme case of
computing fewer graphs.

\begin{figure}[t]
\center{\scalebox{.5}{\includegraphics{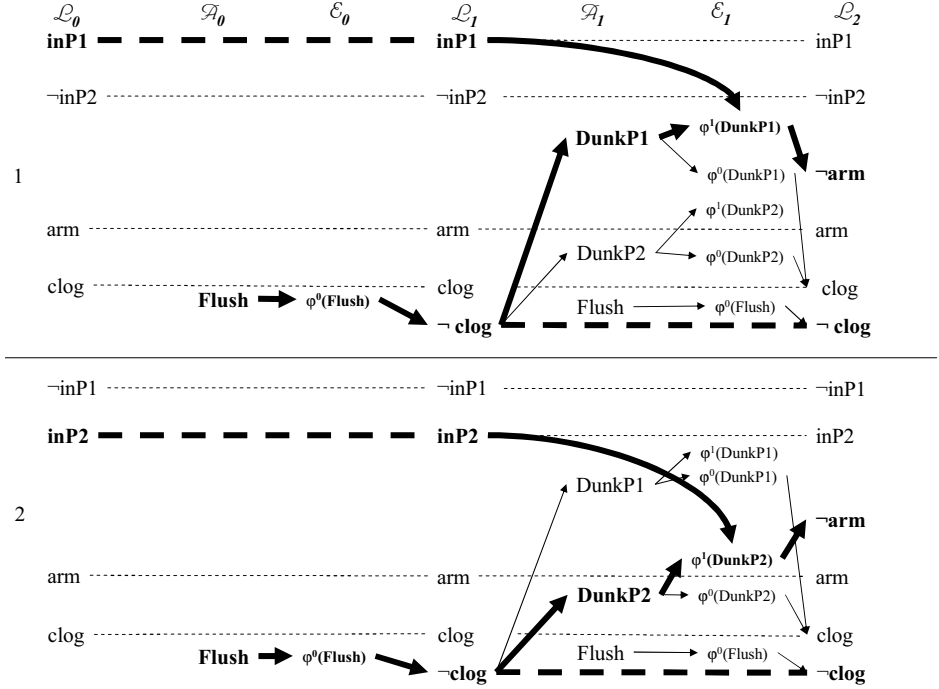}}
\caption[Multiple Graph $CBTC$]{\label{mgbtc} Multiple planning
graphs for CBTC, with relaxed plan components bolded.  Mutexes
omitted.} }
\end{figure}

To illustrate the use of multiple planning graphs, consider our
example CBTC. We build two graphs (Figure \ref{mgbtc}) for the
projected $BS_P$. They have the respective initial literal layers:

\smallskip

\noindent${\cal L}^{1}_{0} = \{$arm, clog, inP1, $\neg$inP2$\}$ and

\smallskip

\noindent${\cal L}^{2}_{0} = \{$arm, clog, $\neg$inP2, inP2$\}$.

\smallskip

\noindent In the graph for the first possible world, $\neg$arm comes
in only through DunkP1 at level 2. In the graph for the second
world, $\neg$arm comes in only through DunkP2 at level 2. Thus, the
multiple graphs show which actions in the different worlds
contribute to support the same literal.

A single planning graph is sufficient if we do not aggregate state
measures, so in the following we consider how to compute the
achievement cost of a belief state with multiple graphs by
aggregating state distances.

\und{Positive Interaction Aggregation} Similar to GPT
\citep{bonet00planning}, we can use the worst-case world to
represent the cost of the belief state $BS_i$ by using the
$h^{MG}_{m-RP}$ heuristic.  The difference with GPT is that we
compute a heuristic on planning graphs, where they compute plans in
state space.  With this heuristic we account for the number of
actions used in a given world, but assume positive interaction
across all possible worlds.

The $h^{MG}_{m-RP}$ heuristic is computed by finding a relaxed plan
$RP_{\gamma}$ on each planning graph $\gamma \in \Gamma$, exactly as
done on the single graph with $h^{SG}_{RP}$.  The difference is that
unlike the single graph relaxed plan $SG^U$, but like $SG^1$, the
initial levels of the planning graphs are states, so each relaxed
plan will reflect all the support needed in the world corresponding
to $\gamma$.  Formally:
\begin{equation}\notag
h^{MG}_{m-RP}(BS_i) =
\max\limits_{\gamma\in{\Gamma}}\left(\sum\limits_{j =
0}^{b_{\gamma}-1} \mid{{\cal A}^{RP_\gamma}_{j}}\mid \right)
\end{equation}
where $b_{\gamma}$ is the level of $\gamma$ where a constituent of
$BS_G$ was first reachable.

Notice that we are not computing all state distances between states
in $BS_P$ and $BS_i$.  Each planning graph $\gamma$ corresponds to a
state in $BS_P$, and from each $\gamma$ we extract a single relaxed
plan. We do not need to enumerate all states in $BS_i$ and find a
relaxed plan for each.  We instead support a set of literals from
one constituent of $BS_i$.  This constituent is estimated to be the
minimum distance state in $BS_i$ because it is the first constituent
reached in $\gamma$.

For CBTC, computing $h^{MG}_{m-RP}(BS_G)$ (Figure \ref{mgbtc})
finds:
\smallskip

\noindent${RP}_1 =$

\smallskip

${\cal A}^{RP_1}_{0} = \{$inP1$_p$, Flush$\}$,

\smallskip

${\cal E}^{RP_1}_{0} = \{\varphi^0($inP1$_p)$,
$\varphi^0($Flush$)$\},

\smallskip

${\cal L}^{RP_1}_{1} = \{$inP1$, \neg$clog$\}$,

\smallskip

${\cal A}^{RP_1}_{1} = \{\neg$clog$_p$, DunkP1$\}$,

\smallskip

${\cal E}^{RP_1}_{1} = \{\varphi^0(\neg$clog$_p)$,
$\varphi^1($DunkP1$)$\},

\smallskip

${\cal L}^{RP_1}_{2} = \{\neg$arm$, \neg$clog$\}$

\smallskip

\noindent and ${RP}_2 =$

\smallskip

${\cal A}^{RP_2}_{0} = \{$inP2$_p$, Flush$\}$,

\smallskip

${\cal E}^{RP_2}_{0} = \{\varphi^0($inP2$_p)$,
$\varphi^0($Flush$)$\},

\smallskip

${\cal L}^{RP_2}_{1} = \{$inP2$, \neg$clog$\}$,

\smallskip

${\cal A}^{RP_2}_{1} = \{\neg$clog$_p$, DunkP2$\}$,

\smallskip

${\cal E}^{RP_2}_{1} = \{\varphi^0(\neg$clog$_p)$,
$\varphi^1($DunkP2$)$\},

\smallskip

${\cal L}^{RP_2}_{2} = \{\neg$arm$, \neg$clog$\}$.

\smallskip

\noindent Each relaxed plan contains two actions and taking the
maximum of the two relaxed plan values gives $h^{MG}_{m-RP}(BS_G) =
2$.  This aggregation ignores the fact that we must use different
Dunk actions each possible world.

\und{Independence Aggregation} We can use the $h^{MG}_{s-RP}$
heuristic to assume independence among the worlds in our belief
state. We extract relaxed plans exactly as described in the previous
heuristic and simply use a summation rather than maximization of the
relaxed plan costs.  Formally:
\begin{equation}\notag
h^{MG}_{s-RP}(BS_i) =
\sum\limits_{\gamma\in{\Gamma}}\left(\sum\limits_{j =
0}^{b_{\gamma}-1} \mid{{\cal A}^{RP_\gamma}_{j}}\mid \right)
\end{equation}
where $b_{\gamma}$ is the level of $\gamma$ where a constituent of
$BS_G$ was first reachable.

For CBTC, if computing $h^{MG}_{s-RP}(BS_G)$, we find the same
relaxed plans as in the $h^{MG}_{m-RP}(BS_G)$ heuristic, but sum
their values to get 2 + 2 = 4 as our heuristic.  This aggregation
ignores the fact that we can use the same Flush action for both
possible worlds.

\und{State Overlap Aggregation} We notice that in the two previous
heuristics we are either taking a maximization and not accounting
for some actions, or taking a summation and possibly accounting for
extra actions. We present the $h^{MG}_{RPU}$ heuristic to balance
the measure between positive interaction and independence of worlds.
Examining the relaxed plans computed by the two previous heuristics
for the CBTC example, we see that the relaxed plans extracted from
each graph have some overlap. Notice, that both ${\cal
A}^{RP_1}_{0}$ and ${\cal A}^{RP_2}_{0}$ contain a Flush action
irrespective of which package the bomb is in -- showing some
positive interaction. Also, ${\cal A}^{RP_1}_{1}$ contains DunkP1,
and ${\cal A}^{RP_2}_{1}$ contains DunkP2 -- showing some
independence. If we take the layer-wise union of the two relaxed
plans, we would get a unioned relaxed plan:
\smallskip

\noindent $RP_{U} = $

\smallskip

${\cal A}^{RP_U}_{0} = \{$inP1$_p$, Flush$\}$,

\smallskip

${\cal E}^{RP_U}_{0} = \{\varphi^0($inP1$_p)$,
$\varphi^0($inP2$_p)$, $\varphi^0($Flush$)$\},

\smallskip

${\cal L}^{RP_U}_{1} = \{$inP1, inP2$, \neg$clog$\}$,

\smallskip

${\cal A}^{RP_U}_{1} = \{\neg$clog$_p$, DunkP1, DunkP2$\}$,

\smallskip

${\cal E}^{RP_U}_{1} = \{\varphi^0(\neg$clog$_p)$,
$\varphi^1($DunkP1$)$, $\varphi^1($DunkP2$)$\},

\smallskip

${\cal L}^{RP_U}_{2} = \{\neg$arm$, \neg$clog$\}$.
\smallskip

This relaxed plans accounts for the actions that are the same
between possible worlds and the actions that differ.  Notice that
Flush appears only once in layer zero and the Dunk actions both
appear in layer one.

In order to get the union of relaxed plans, we extract relaxed plans
from each $\gamma\in{\Gamma}$, as in the two previous heuristics.
Then if we are computing heuristics for regression search, we start
at the last level (and repeat for each level) by taking the union of
the sets of actions for each relaxed plan at each level into another
relaxed plan. The relaxed plans are {\em end-aligned}, hence the
unioning of levels proceeds from the last layer of each relaxed plan
to create the last layer of the $RP_{U}$ relaxed plan, then the
second to last layer for each relaxed plan is unioned and so on. In
progression search, the relaxed plans are {\em start-aligned} to
reflect that they all start at the same time, whereas in regression
we assume they all end at the same time.  The summation of the
number of actions of each action level in the unioned relaxed plan
is used as the heuristic value. Formally:
\begin{equation}\notag
h^{MG}_{RPU}(BS_i) = \sum\limits_{j = 0}^{b-1}  \mid{{\cal
A}^{RP_U}_{j}}\mid
\end{equation}
where $b$ is the greatest level $b_{\gamma}$ where a constituent
of $BS_G$ was first reachable.

For CBTC, we just found $RP_U$, so counting the number of actions
gives us a heuristic value of $h^{MG}_{RPU}(BS_G) = 3$.

\subsection{Labelled Uncertainty Graph Heuristics ($LUG$)}

The multiple graph technique has the advantage of heuristics that
can aggregate the costs of multiple worlds, but the disadvantage of
computing some redundant information in different graphs (c.f.
Figure \ref{mgbtc}) and using every graph to compute heuristics (c.f
$h_{RPU}^{MG}$). Our next approach addresses these limitations by
condensing the multiple planning graphs to a single planning graph,
called a labelled uncertainty graph ($LUG$).  The idea is to
implicitly represent multiple planning graphs by collapsing the
graph connectivity into one planning graph, but use annotations,
called labels ($\ell$), to retain information about multiple worlds.
While we could construct the $LUG$ by generating each of the
multiple graphs and taking their union, instead we define a direct
construction procedure.  We start in a manner similar to the unioned
single planning graph ($SG^U$) by constructing an initial layer of
all literals in our source belief state.  The difference with the
$LUG$ is that we can prevent loss of information about multiple
worlds by keeping a label for each literal the records which of the
worlds is relevant. As we will discuss, we use a few simple
techniques to propagate the labels through actions and effects and
label subsequent literal layers.  Label propagation relies on
expressing labels as propositional formulas and using standard
propositional logic operations.  The end product is a single
planning graph with labels on all graph elements; labels indicate
which of the explicit multiple graphs (if we were to build them)
contain each graph element.

We are trading planning graph structure space for label storage
space.  Our choice of BDDs to represent labels helps lower the
storage requirements on labels.  The worst-case complexity of the
$LUG$ is equivalent to the $MG$ representation. The $LUG$'s
complexity savings is not realized when the projected possible
worlds and the relevant actions for each are completely disjoint;
however, this does not often appear in practice.  The space savings
comes in two ways: (1) redundant representation of actions and
literals is avoided, and (2) labels that facilitate non-redundant
representation are stored as BDDs.  A nice feature of the BDD
package \citep{CUDD} we use is that it efficiently represents many
individual BDDs in a shared BDD that leverages common substructure.
Hence, in practice the $LUG$ contains the same information as $MG$
with much lower construction and usage costs.

In this section we present construction of the $LUG$ without
mutexes, then describe how to introduce mutexes, and finally discuss
how to extract relaxed plans.

\subsubsection{Label Propagation}

Like the single graph and multiple graphs, the $LUG$ is based on the
$IPP$ \citep{ipp} planning graph.  We extend the single graph to
capture multiple world causal support, as present in multiple
graphs, by adding labels to the elements of the action ${\cal A}$,
effect ${\cal E}$, and literal ${\cal L}$ layers.  We denote the
label of a literal $l$ in level $k$ as $\ell_k(l)$.  We can build
the $LUG$ for any belief state $BS_P$, and illustrate $BS_P = BS_I$
for the CBTC example.  A label is a formula describing a set of
states (in $BS_P$) from which a graph element is (optimistically)
{\em reachable}. We say a literal $l$ is reachable from a set of
states, described by $BS$, after $k$ levels, if $BS \models
\ell_k(l)$. For instance, we can say that $\neg$arm is reachable
after two levels if  ${\cal L}_2$ contains $\neg$arm and $BS_I
\models \ell_2(\neg$arm), meaning that the models of worlds where
$\neg$arm holds after two levels are a superset of the worlds in our
current belief state.

\begin{figure}[btcLUG]
\center{\scalebox{.6}{\includegraphics{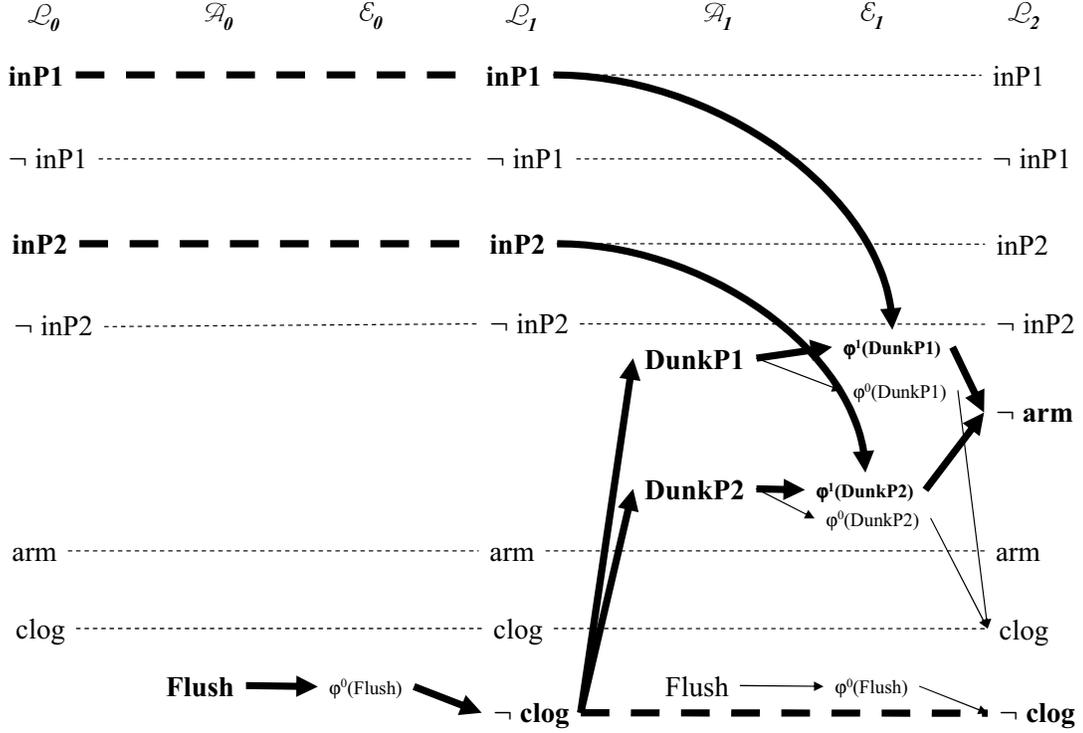}}
\vspace{-1cm} \caption[btcLUG]{\label{btcLUG} The $LUG$ skeleton
for CBTC, with no mutexes. The relaxed plan for $h^{LUG}_{RP}$ is
shown in bold.}}
\end{figure}

The intuitive definition of the $LUG$ is a planning graph skeleton,
that represents causal relations, over which we propagate labels to
indicate specific possible world support.  We show the skeleton for
CBTC in Figure \ref{btcLUG}.  Constructing the graph skeleton
largely follows traditional planning graph semantics, and label
propagation relies on a few simple rules. Each initial layer literal
is labelled, to indicate the worlds of $BS_P$ in which it holds, as
the conjunction of the literal with $BS_P$.  An action is labelled,
to indicate all worlds where its execution preconditions can be
co-achieved, as the conjunction of the labels of its execution
preconditions.  An effect is labelled, to indicate all worlds where
its antecedent literals and its action's execution preconditions can
be co-achieved, as the conjunction of the labels of its antecedent
literals and the label of its associated action. Finally, literals
are labelled, to indicate all worlds where they are given as an
effect, as the disjunction over all labels of effects in the
previous level that affect the literal. In the following we describe
label propagation in more detail and work through the CBTC example.

\und{Initial Literal Layer} The $LUG$ has an initial layer
consisting of every literal with a non false ($\perp$) label.  In
the initial layer the label $\ell_0(l)$ of each literal $l$ is
identical to $l \wedge BS_P$, representing the states of $BS_P$ in
which $l$ holds.  The labels for the initial layer literals are
propagated through actions and effects to label the next literal
layer, as we will describe shortly.  We continue propagation until
no label of any literal changes between layers, a condition referred
to as ``level off''.

The $LUG$ for CBTC, shown in Figure \ref{btcLUG} (without labels),
using $BS_P$=$BS_I$ has the initial literal layer:
\begin{equation}\label{initlit}
\begin{array}{l}\notag
{\cal L}_0 = \{{\tt inP1}, \neg {\tt inP2}, {\tt
inP2}, \neg {\tt inP1}, {\tt clog}, {\tt arm}\}\\
\ell_0({\tt inP1}) = \ell_0(\neg {\tt inP2}) = ({\tt arm} \wedge {\tt clog} \wedge {\tt inP1} \wedge \neg {\tt inP2}),\\
\ell_0({\tt inP2}) = \ell_0(\neg {\tt inP1}) = ({\tt arm} \wedge {\tt clog} \wedge \neg {\tt inP1} \wedge {\tt inP2}),\\
\ell_0({\tt clog}) = \ell_0({\tt arm}) = BS_P
\end{array}
\end{equation}

Notice that inP1 and inP2 have labels indicating the respective
initial states in which they hold, and clog and arm have $BS_P$ as
their label because they hold in all states in $BS_P$.

\und{Action Layer} Once the previous literal layer ${\cal L}_k$ is
computed, we construct and label the action layer ${\cal A}_k$.
${{\cal A}}_k$ contains causative actions from the action set $A$,
plus literal persistence.  An action is included in ${{\cal A}}_k$
if its label is not false (i.e. $\ell_k(a) \not= \perp$). The label
of an action at level $k$, is equivalent to the extended label of
its execution precondition:
\begin{equation}\label{acts}\notag
 \ell_k(a) = \ell^*_k(\rho^e(a))
\end{equation}

Above, we introduce the notation for extended labels $\ell_k^*(f)$
of a formula $f$ to denote the worlds of $BS_P$ that can reach $f$
at level $k$. We say that any propositional formula $f$ is reachable
from $BS$ after $k$ levels if $BS_i \models \ell^*_k(f)$.  Since we
only have labels for literals, we substitute the labels of literals
for the literals in a formula to get the extended label of the
formula. The extended label of a propositional formula $f$ at level
$k$, is defined:
\begin{equation}\label{formulalabel}\notag
\begin{array}{c}
 \ell^*_k(f \wedge f') = \ell^*_k(f) \wedge \ell^*_k(f'),\\
 \ell^*_k(f \vee f') = \ell^*_k(f) \vee \ell^*_k(f'),\\
 \ell^*_k(\neg (f \wedge f')) = \ell^*_k(\neg f \vee \neg f'),\\
 \ell^*_k(\neg (f \vee f')) = \ell^*_k(\neg f \wedge \neg f'),\\
 \ell^*_k(\top) = BS_P,\\
 \ell^*_k(\perp) = \perp,\\
  \ell^*_k(l) = \ell_k(l)\\
 \end{array}
\end{equation}
\normalsize

The zeroth action layer for CBTC is:
\begin{equation}
\begin{array}{l}\notag
{{\cal A}}_0 = \{{\tt Flush}, {\tt inP1}_p, \neg {\tt inP2}_p, {\tt
inP2}_p, \neg {\tt inP1}_p, {\tt clog}_p, {\tt arm}_p\}\\
\ell_0({\tt Flush}) = BS_P,\\
\ell_0({\tt inP1}_p) = \ell_0(\neg {\tt inP2}_p) = ({\tt arm} \wedge {\tt clog} \wedge {\tt inP1} \wedge \neg {\tt inP2}),\\
\ell_0({\tt inP2}_p) = \ell_0(\neg {\tt inP1}_p) = ({\tt arm} \wedge {\tt clog} \wedge \neg {\tt inP1} \wedge {\tt inP2}),\\
\ell_0({\tt clog}_p) = \ell_0({\tt arm}_p) = BS_P
\end{array}
\end{equation}

Each literal persistence has a label identical to the label of the
corresponding literal from the previous literal layer.  The Flush
action has $BS_P$ as its label because it is always applicable.

\medskip

\und{Effect Layer} The effect layer ${{\cal E}}_k$ depends both on
the literal layer ${\cal L}_k$ and action layer ${\cal A}_k$.
${{\cal E}}_k$ contains an effect $\varphi^j(a)$ if the effect has a
non false label (i.e. $\ell_k(\varphi^j(a)) \not= \perp$). Because
both the action and an effect must be applicable in the same world,
the label of the effect at level $k$ is the conjunction of the label
of the associated action with the extended label of the antecedent
\begin{equation}\label{effs}\notag
\ell_k(\varphi^j(a)) = \ell_k(a) \wedge \ell^*_k(\rho^j(a))
\end{equation}

The zeroth effect layer for CBTC is:
\begin{equation}
\begin{array}{l}\notag
{{\cal E}}_0 = \{\varphi^0({\tt Flush}), \varphi^0({\tt inP1}_p), \varphi^0(\neg {\tt inP2}_p), \varphi^0({\tt inP2}_p),\\
\qquad\;\;\,\varphi^0(\neg {\tt inP1}_p), \varphi^0({\tt clog}_p), \varphi^0({\tt arm}_p) \}\\
\ell_0(\varphi^0({\tt Flush})) =  BS_P\\
\ell_0(\varphi^0({\tt inP1}_p)) = \ell_0(\varphi^0(\neg {\tt inP2}_p)) = ({\tt arm} \wedge {\tt clog} \wedge {\tt inP1} \wedge \neg {\tt inP2}),\\
\ell_0(\varphi^0({\tt inP2}_p)) = \ell_0(\varphi^0(\neg {\tt inP1}_p)) = ({\tt arm} \wedge {\tt clog} \wedge \neg {\tt inP1} \wedge {\tt inP2}),\\
\ell_0(\varphi^0({\tt clog}_p)) = \ell_0(\varphi^0({\tt arm}_p)) =
BS_P
\end{array}
\end{equation}

Again, like the action layer, the unconditional effect of each
literal persistence has a label identical to the corresponding
literal in the previous literal layer.  The unconditional effect
of Flush has a label identical to the label of Flush.

\medskip

\und{Literal Layer} The literal layer ${\cal L}_k$ depends on the
previous effect layer ${\cal E}_{k-1}$, and contains only literals
with non false labels (i.e. $\ell_k(l) \not= \perp$). An effect
$\varphi^j(a) \in {\cal E}_{k-1}$ contributes to the label of a
literal $l$ when the effect consequent contains the literal $l$. The
label of a literal is the disjunction of the labels of each effect
from the previous effect layer that gives the literal:
\begin{equation}\label{literals}\notag
\ell_{k}(l) = \bigvee_{\substack{\varphi^j(a): l \in \varepsilon^{j}(a),\\
\varphi^j(a) \in {\cal E}_{k-1}}} \ell_{k-1}(\varphi^j(a))
\end{equation}

The first literal layer for CBTC is:
\begin{equation}
\begin{array}{l}\notag
{\cal L}_1 = \{ {\tt inP1}, \neg {\tt inP2}, {\tt inP2}, \neg {\tt
inP1}, \neg {\tt clog}, {\tt clog}, {\tt arm} \}\\
\ell_1({\tt inP1}) = \ell_1(\neg {\tt inP2}) = ({\tt arm} \wedge {\tt clog} \wedge {\tt inP1} \wedge \neg {\tt inP2}),\\
\ell_1({\tt inP2}) = \ell_1(\neg {\tt inP1}) = ({\tt arm} \wedge {\tt clog} \wedge \neg {\tt inP1} \wedge {\tt inP2}),\\
\ell_1(\neg {\tt clog}) = \ell_1({\tt clog}) = \ell_1({\tt arm}) =
BS_P
\end{array}
\end{equation}

This literal layer is identical to the initial literal layer, except
that $\neg$clog goes from having a false label (i.e. not existing in
the layer) to having the label $BS_P$.

\medskip

We continue to the level one action layer because ${\cal L}_1$ does
not indicate that $BS_G$ is reachable from $BS_P$ ($\neg$arm
$\not\in {\cal L}_1$). Action layer one is defined:
\begin{equation}
\begin{array}{l}\notag
{{\cal A}}_1 = \{ {\tt DunkP1}, {\tt DunkP2}, {\tt Flush}, {\tt
inP1}_p, \neg {\tt inP2}_p, {\tt inP2}_p, \neg {\tt inP1}_p, {\tt
clog}_p, {\tt arm}_p, \neg {\tt
clog}_p \}\\
\ell_1({\tt DunkP1}) = \ell_1({\tt DunkP2}) = \ell_1({\tt Flush}) = BS_P,\\
\ell_1({\tt inP1}_p) = \ell_1(\neg {\tt inP2}_p) = ({\tt arm} \wedge {\tt clog} \wedge {\tt inP1} \wedge \neg {\tt inP2}),\\
\ell_1({\tt inP2}_p) = \ell_1(\neg {\tt inP1}_p) = ({\tt arm} \wedge {\tt clog} \wedge \neg {\tt inP1} \wedge {\tt inP2}),\\
\ell_1({\tt clog}_p) = \ell_1({\tt arm}_p) = \ell_1(\neg {\tt
clog}_p) = BS_P
\end{array}
\end{equation}

This action layer is similar to the level zero action layer.  It
adds both Dunk actions because they are now executable.  We also
add the persistence for $\neg$clog.  Each Dunk action gets a label
identical to its execution precondition label.

The level one effect layer is:
\begin{equation}
\begin{array}{l}\notag
{{\cal E}}_1 = \{\varphi^0({\tt DunkP1}), \varphi^0({\tt DunkP2}), \varphi^1({\tt DunkP1}), \varphi^1({\tt DunkP2}), \varphi^0({\tt Flush}), \varphi^0({\tt inP1}_p),\\
\qquad\;\;\,\varphi^0(\neg {\tt inP2}_p), \varphi^0({\tt inP2}_p),\varphi^0(\neg {\tt inP1}_p), \varphi^0({\tt clog}_p), \varphi^0({\tt arm}_p), \varphi^0(\neg {\tt clog}_p)\}\\
\ell_1(\varphi^0({\tt DunkP1})) = \ell_1(\varphi^0({\tt DunkP2})) = \ell_1(\varphi^0({\tt Flush})) =  BS_P\\
\ell_1(\varphi^1({\tt DunkP1})) = ({\tt arm} \wedge {\tt clog} \wedge {\tt inP1} \wedge \neg {\tt inP2}),\\
\ell_1(\varphi^1({\tt DunkP2})) = ({\tt arm} \wedge {\tt clog} \wedge \neg {\tt inP1} \wedge {\tt inP2}),\\
\ell_1(\varphi^0(\neg {\tt inP2}_p)) = \ell_1(\varphi^0({\tt inP1}_p)) = ({\tt arm} \wedge {\tt clog} \wedge {\tt inP1} \wedge \neg {\tt inP2}),\\
\ell_1(\varphi^0(\neg {\tt inP1}_p)) = \ell_1(\varphi^0({\tt inP2}_p)) = ({\tt arm} \wedge {\tt clog} \wedge \neg {\tt inP1} \wedge {\tt inP2}),\\
\ell_1(\varphi^0({\tt clog}_p)) = \ell_1(\varphi^0({\tt arm}_p)) =
\ell_1(\varphi^0(\neg {\tt clog}_p)) = BS_P
\end{array}
\end{equation}

The conditional effects of the Dunk actions in CBTC (Figure
\ref{btcLUG}) have labels that indicate the possible worlds in which
they will give $\neg$arm because their antecedents do not hold in
all possible worlds.  For example, the conditional effect
$\varphi^1($DunkP1$)$ has the label found by taking the conjunction
of the action's label $BS_P$ with the antecedent label
$\ell^*_1$(inP1) to obtain $({\tt arm} \wedge {\tt clog} \wedge {\tt
inP1} \wedge \neg {\tt inP2})$.

Finally, the level two literal layer:
\begin{equation}
\begin{array}{l}\notag
{\cal L}_2 = \{ {\tt inP1}, \neg {\tt inP2}, {\tt inP2}, \neg {\tt
inP1}, \neg {\tt clog}, {\tt clog}, {\tt arm}, \neg {\tt arm} \}\\
\ell_2({\tt inP1}) = \ell_2(\neg {\tt inP2}) = ({\tt arm} \wedge {\tt clog} \wedge {\tt inP1} \wedge \neg {\tt inP2}),\\
\ell_2({\tt inP2}) = \ell_2(\neg {\tt inP1}) = ({\tt arm} \wedge {\tt clog} \wedge \neg {\tt inP1} \wedge {\tt inP2}),\\
\ell_2(\neg {\tt clog}) = \ell_2({\tt clog}) = \ell_2({\tt arm}) =
\ell_2(\neg {\tt arm}) = BS_P
\end{array}
\end{equation}

The labels of the literals for level 2 of CBTC indicate that
$\neg$arm is reachable from $BS_P$ because its label is entailed by
$BS_P$.  The label of $\neg$arm is found by taking the disjunction
of the labels of effects that give it, namely, $({\tt arm} \wedge
{\tt clog} \wedge {\tt inP1} \wedge \neg {\tt inP2})$ from the
conditional effect of DunkP1 and $({\tt arm} \wedge {\tt clog}
\wedge \neg {\tt inP1} \wedge {\tt inP2})$ from the conditional
effect of DunkP2, which reduces to $BS_P$. Construction could stop
here because $BS_P$ entails the label of the goal
$\ell^{*}_k(\neg$arm$\wedge \neg$clog)$ = \ell_k(\neg$arm$) \wedge
\ell_k(\neg$clog$) = BS_P \wedge BS_P = BS_P$. However, level off
occurs at the next level because there is no change in the labels of
the literals.

When level off occurs at level three in our example, we can say that
for any $BS$, where $BS \models BS_P$, that a formula $f$ is
reachable in $k$ steps if $BS \models \ell^*_k(f)$. If no such level
$k$ exists, then $f$ is not reachable from $BS$. If there is some
level $k$, where $f$ is reachable from $BS$, then the first such $k$
is a lower bound on the number of parallel plan steps needed to
reach $f$ from $BS$. This lower bound is similar to the classical
planning max heuristic \citep{nguyen02planning}.  We can provide a
more informed heuristic by extracting a relaxed plan to support $f$
with respect to $BS$, described shortly.

\subsubsection{Same-World Labelled Mutexes}

There are several types of mutexes that can be added to the $LUG$.
To start with, we only concentrate on those that can evolve in a
single possible world because same-world mutexes are more effective
as well as relatively easy to understand.  We extend the mutex
propagation that was used in the multiple graphs so that the mutexes
are on one planning graph.  The savings of computing mutexes on the
$LUG$ instead of multiple graphs is that we can reduce computation
when a mutex exits in several worlds.  In Appendix B we describe how
to handle cross-world mutexes, despite their lack of effectiveness
in the experiments we conducted. Cross-world mutexes extend the
$LUG$ to compute the same set of mutexes found by CGP
\citep{AAAI98_IAAI98*889}.

Same-world mutexes can be represented with a single label,
$\hat{\ell}_k(x_1, x_2)$, between two elements (actions, effect, or
literals). The mutex holds between elements $x_1$ and $x_2$ in all
worlds $S$ where $S \models \hat{\ell}_k(x_1, x_2)$.  If the
elements are not mutex in any world, we can assume the label of a
mutex between them is false $\perp$. We discuss how the labelled
mutexes are discovered and propagated for actions, effect relations,
and literals.

By using mutexes, we can refine what it means for a formula $f$ to
be reachable from a set of worlds $BS_P$.  We must ensure that for
every state in $BS_P$, there exists a state of $f$ that is
reachable.  A state $S'$ of $f$ is reachable from a state $S$ of
$BS_P$ when there are no two literals in $S'$ that are mutex in
world $S$ and $BS_P \models \ell_k^*(S)$.

In each of the action, effect, and literal layers there are
multiple ways for the same pair of elements to become mutex (e.g.
interference or competing needs).  Thus, the mutex label for a
pair is the disjunction of all labelled mutexes found for the pair
by some means.

\und{Action Mutexes}   The same-world action mutexes at a level $k$
are a set of labelled pairs of actions. Each pair is labelled with a
formula that indicates the set of possible worlds where the actions
are mutex.  The possible reasons for mutex actions are interference
and competing needs.

\begin{itemize}
\item {\bf Interference} Two actions $a, a'$ interfere if (1) the
unconditional effect consequent $\varepsilon^0(a)$ of one is
inconsistent with the execution precondition $\rho^e(a')$ of the
other, or (2) vice versa.  They additionally interfere if (3) both
unconditional effect consequents $\varepsilon^0(a)$ and
$\varepsilon^0(a')$ are inconsistent, or (4) both execution
preconditions $\rho^e(a)$ and $\rho^e(a')$ are inconsistent. The
mutex will exist in all possible world projections $\hat{\ell}_k(a,
a') = BS_P$. Formally, $a$ and $a'$ interfere if one of the
following holds:
\begin{equation}
\begin{array}{l}\notag
{\tt (1)}\;    \varepsilon^0(a) \wedge \rho^e(a') = \perp\\
{\tt (2)}\; \rho^e(a) \wedge \varepsilon^0(a') = \perp\\
{\tt (3)}\;   \varepsilon^0(a) \wedge \varepsilon^0(a') = \perp \\
{\tt (4)}\; \rho^e(a) \wedge \rho^e(a') = \perp
\end{array}
\end{equation}

\item {\bf Competing Needs} Two actions $a, a'$ have competing
needs in a world when a pair of literals from their execution
preconditions are mutex in the world.  The worlds where $a$ and $a'$
are mutex because of competing needs are described by:
\begin{equation}\notag
 \ell_k(a) \wedge \ell_k(a') \wedge \bigvee_{\substack{l \in \rho^j(a), l'\in \rho^j(a') }}
\hat{\ell}_k(l, l')
\end{equation}

In the above formula we find all worlds where a pair of execution
preconditions $l \in \rho^e(a), l'\in \rho^e(a')$ are mutex and both
actions are reachable.
\end{itemize}

\und{Effect Mutexes}   The effect mutexes are a set of labelled
pairs of effects.  Each pair is labelled with a formula that
indicates the set of possible worlds where the effects are mutex.
The possible reasons for mutex effects are associated action
mutexes, interference, competing needs, or induced effects.

\begin{itemize}
\item {\bf Mutex Actions}  Two effects $\varphi^i(a) \in \Phi(a),
\varphi^{j}(a') \in \Phi({a'})$ are mutex in all worlds where their
associated actions are mutex, $\hat{\ell}_k(a, a')$.

\item {\bf Interference} Like actions, two effects $\varphi^i(a), \varphi^{j}(a')$ interfere if (1) the
consequent $\varepsilon^i(a)$ of one is inconsistent with the
antecedent $\rho^j(a')$ of the other, or (2) vice versa. They
additionally interfere if (3) both effect consequents
$\varepsilon^i(a)$ and $\varepsilon^j(a')$ are inconsistent, or (4)
both antecedents $\rho^i(a)$ and $\rho^j(a')$ are inconsistent. The
mutex will exist in all possible world projections, so the label of
the mutex is $\hat{\ell}_k(\varphi^i(a), \varphi^{j}(a')) = BS_P$.
Formally, $\varphi^i(a)$ and $\varphi^{j}(a')$ interfere if one of
the following holds:
\begin{equation}
\begin{array}{l}\notag
{\tt (1)}\;    \varepsilon^i(a) \wedge \rho^j(a') = \perp\\
{\tt (2)}\; \rho^i(a) \wedge \varepsilon^j(a') = \perp\\
{\tt (3)}\;   \varepsilon^i(a) \wedge \varepsilon^j(a') = \perp \\
{\tt (4)}\; \rho^i(a) \wedge \rho^j(a') = \perp
\end{array}
\end{equation}

\item {\bf Competing Needs} Like actions, two effects have
competing needs in a world when a pair of literals from their
antecedents are mutex in a world. The worlds where $\varphi^i(a)$
and $\varphi^{j}(a')$ have a competing needs mutex are:

\begin{equation}\notag
\ell_k(\varphi^i(a)) \wedge \ell_k(\varphi^{j}(a')) \wedge
\bigvee_{\substack{l \in \rho^i(a), l'\in \rho^j(a') }}
\hat{\ell}_k(l, l')
\end{equation}

In the above formula we find all worlds where a pair of execution
preconditions $l \in \rho^i(a), l'\in \rho^j(a')$ are mutex and both
actions are reachable.

\item {\bf Induced} An induced effect $\varphi^j(a)$ of an effect $\varphi^i(a)$ is an effect of
the same action $a$ that may execute at the same time.  An effect is
induced by another in the possible worlds where they are both
reachable.  For example, the conditional effect of an action always
induces the unconditional effect of the action.

Induced mutexes, involving the inducing effect $\varphi^i(a)$, come
about when an induced effect $\varphi^j(a)$ is mutex with another
effect $\varphi^{h}(a')$ (see Figure \ref{induceex}). The induced
mutex is between (a) the effect $\varphi^{h}(a')$ that is mutex with
the induced effect $\varphi^j(a)$ and (b) the inducing effect
$\varphi^i(a)$. The label of the mutex is the conjunction of the
label of the mutex $\hat{\ell_k}(\varphi^j(a), \varphi^{h}(a'))$ and
the label of the induced effect $\varphi^j(a)$. For additional
discussion of the methodology behind induced mutexes we refer to
\citet{AAAI98_IAAI98*889}.

\begin{figure}[tbp]
\center{\scalebox{.45}{\includegraphics{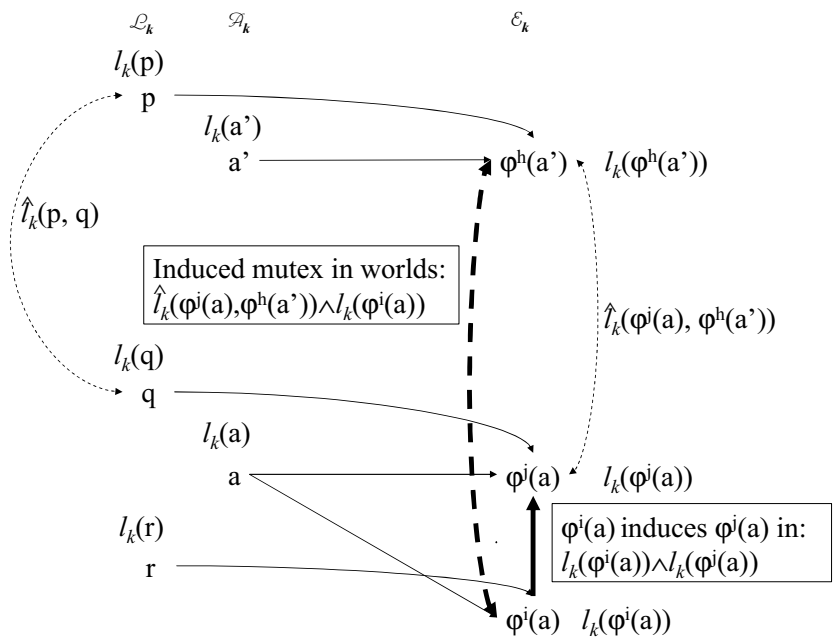}}
\caption[btcLUG]{\label{induceex} Effect $\varphi^i(a)$ induces
effect $\varphi^j(a)$.  $\varphi^j(a)$ is mutex with
$\varphi^{h}(a')$, so $\varphi^i(a)$ is induced mutex with
$\varphi^{h}(a')$.}}
\end{figure}
\end{itemize}

\und{Literal Mutexes}  The literal mutexes are a set of labelled
pairs of literals.  Each pair is labelled with a formula that
indicates the set of possible worlds where the literals are mutex.
The only reason for mutex literals is inconsistent support.

\begin{itemize}
\item {\bf Inconsistent Support} Two literals have inconsistent
support in a possible world at level $k$ when there are no two
non-mutex effects that support both literals in the world. The label
of the literal mutex at level $k$ is a disjunction of all worlds
where they have inconsistent support. The worlds for an inconsistent
support mutex between $l$ and $l'$ are:

\begin{equation}\label{litinconsmuxlab}\notag
\bigvee_{\substack{S: \forall \varphi^i(a), \varphi^{j}(a') \in
{\cal E}_{k-1},\\ {\tt where }\, l \in \varepsilon^i(a), l'\in
\varepsilon^{j}(a'),\\ S \models \hat{\ell}_{k-1}(\varphi^i(a),
\varphi^{j}(a'))}} S
\end{equation}

The meaning of the above formula is that the two literals are mutex
in all worlds $S$ where all pairs of effects that support the
literals in $S$ are mutex in $S$.
\end{itemize}

\subsubsection{$LUG$ Heuristics}

The heuristics computed on the $LUG$ can capture measures similar to
the $MG$ heuristics, but there exists a new opportunity to make use
of labels to improve heuristic computation efficiency. A single
planning graph is sufficient if there is no state aggregation being
measured, so we do not mention such measures for the $LUG$.

\und{Positive Interaction Aggregation} Unlike $MG$ heuristics, we do
not compute positive interaction based relaxed plans on the $LUG$.
The $MG$ approach to measure positive interaction across each state
in a belief state is to compute multiple relaxed plans and take
their maximum value.  To get the same measure on the $LUG$ we would
still need to extract multiple relaxed plans, the situation we are
trying to avoid by using the $LUG$.  While the graph construction
overhead may be lowered by using the $LUG$, the heuristic
computation could take too long.  Hence, we do not compute relaxed
plans on the $LUG$ to measure positive interaction alone, but we do
compute relaxed plans that measure overlap (which measures positive
interaction).

\und{Independence Aggregation} Like positive interaction
aggregation, we need a relaxed plan for every state in the projected
belief state to find the summation of the costs. Hence, we do not
compute relaxed plans that assume independence.

\und{State Overlap Aggregation} A relaxed plan extracted from the
$LUG$ to get the $h^{LUG}_{RP}$ heuristic resembles the unioned
relaxed plan in the $h^{MG}_{RPU}$ heuristic. Recall that the
$h^{MG}_{RPU}$ heuristic extracts a relaxed plan from each of the
multiple planning graphs (one for each possible world) and unions
the set of actions chosen at each level in each of the relaxed
plans. The $LUG$ relaxed plan heuristic is similar in that it counts
actions that have positive interaction in multiple worlds only once
and accounts for independent actions that are used in subsets of the
possible worlds. The advantage of $h^{LUG}_{RP}$ is that we find
these actions with a single pass on one planning graph.

We are trading the cost of computing multiple relaxed plans for
the cost of manipulating $LUG$ labels to determine what lines of
causal support are used in what worlds.  In the relaxed plan we
want to support the goal with every state in $BS_P$, but in doing
so we need to track which states in $BS_P$ use which paths in the
planning graph. A subgoal may have several different (and possibly
overlapping) paths from the worlds in $BS_P$.

A $LUG$ relaxed plan is a set of layers: $\{{\cal A}^{RP}_0, {\cal
E}^{RP}_0, {\cal L}^{RP}_{1}, ...,$ ${\cal A}^{RP}_{b-1}, {\cal
E}^{RP}_{b-1},$ ${\cal L}^{RP}_{b}\}$, where ${\cal A}^{RP}_r$ is
a set of actions, ${\cal E}^{RP}_r$ is a set of effects, and
${\cal L}^{RP}_{r+1}$ is a set of clauses.  The elements of the
layers are labelled to indicate the worlds of $BS_P$ where they
are chosen for support.  The relaxed plan is extracted from the
level  $b = h^{LUG}_{level}(BS_i)$ (i.e., the first level where
$BS_i$ is reachable, also described in Appendix A).

Please note that we are extracting the relaxed plan for $BS_i$ in
terms of clauses, and not literals, which is different than the
$SG$ and $MG$ versions of relaxed plans.  Previously we found the
constituent of $BS_i$ that was first reached on a planning graph
and now we do not commit to any one constituent.  Our rationale is
that we were possibly using different constituents in each of the
multiple graphs, and in this condensed version of the multiple
graphs we still want to be able to support different constituents
of the $BS_i$ in different worlds.  We could also use the
constituent representation of $BS_i$ in defining the layers of the
relaxed plan, but choose the clausal representation of $BS_i$
instead because we know that we have to support each clause.
However with constituents we know we only need to support one (but
we don't need to know which one).

The relaxed plan, shown in bold in Figure \ref{btcLUG}, for $BS_I$
to reach $BS_G$ in CBTC is listed as follows:
\begin{equation}\notag
\noindent\begin{array}{l}\notag
{\cal A}^{RP}_0 = \{{\tt inP1}_p, {\tt inP2}_p, {\tt Flush}\},\\
\quad\ell^{RP}_0({\tt inP1}_p) = ({\tt arm} \wedge \neg {\tt clog} \wedge {\tt inP1} \wedge \neg {\tt inP2}),\\
\quad\ell^{RP}_0({\tt inP2}_p) = ({\tt arm} \wedge \neg {\tt clog} \wedge \neg {\tt inP1} \wedge  {\tt inP2}),\\
\quad\ell^{RP}_0({\tt Flush}) = BS_P,\\
\\
{\cal E}^{RP}_0 = \{\varphi^0({\tt inP1}_p), \varphi^0({\tt inP2}_p), \varphi^0({\tt Flush})\},\\
\quad\ell^{RP}_0(\varphi^0({\tt inP1}_p)) = ({\tt arm} \wedge \neg {\tt clog} \wedge {\tt inP1} \wedge \neg {\tt inP2}),\\
\quad\ell^{RP}_0(\varphi^0({\tt inP2}_p)) = ({\tt arm} \wedge \neg {\tt clog} \wedge \neg {\tt inP1} \wedge  {\tt inP2}),\\
\quad\ell^{RP}_0(\varphi^0({\tt Flush})) = BS_P,\\
\\
{\cal L}^{RP}_1 = \{{\tt inP1}, {\tt inP2}, \neg {\tt clog}\},\\
\quad\ell^{RP}_1({\tt inP1}) = ({\tt arm} \wedge \neg {\tt clog} \wedge {\tt inP1} \wedge \neg {\tt inP2}),\\
\quad\ell^{RP}_1({\tt inP2}) = ({\tt arm} \wedge \neg {\tt clog} \wedge \neg {\tt inP1} \wedge  {\tt inP2}) ,\\
\quad\ell^{RP}_1(\neg {\tt clog}) = BS_P,\\
\\
{\cal A}^{RP}_1 = \{{\tt DunkP1}, {\tt DunkP2}, \neg {\tt clog}_p\},\\
\quad\ell^{RP}_1({\tt DunkP1}) = ({\tt arm} \wedge \neg {\tt clog} \wedge {\tt inP1} \wedge \neg {\tt inP2}),\\
\quad\ell^{RP}_1({\tt DunkP2}) = ({\tt arm} \wedge \neg {\tt clog} \wedge \neg {\tt inP1} \wedge  {\tt inP2}),\\
\quad\ell^{RP}_1(\neg {\tt clog}_p) = BS_P,\\
\\
{\cal E}^{RP}_1 = \{\varphi^1({\tt DunkP1}), \varphi^1({\tt DunkP2}), \varphi^0(\neg {\tt clog}_p)\},\\
\quad\ell^{RP}_1(\varphi^1({\tt DunkP1})) = ({\tt arm} \wedge \neg {\tt clog} \wedge {\tt inP1} \wedge \neg {\tt inP2}),\\
\quad\ell^{RP}_1(\varphi^1({\tt DunkP2})) = ({\tt arm} \wedge \neg {\tt clog} \wedge \neg {\tt inP1} \wedge  {\tt inP2}),\\
\quad\ell^{RP}_1(\varphi^0(\neg {\tt clog}_p)) = BS_P,\\
\\
{\cal L}^{RP}_2 = \{\neg {\tt arm}, \neg {\tt clog}\},\\
\quad\ell^{RP}_2(\neg {\tt arm}) = BS_P,\\
\quad\ell^{RP}_2(\neg {\tt clog}) = BS_P\\
\end{array}
\end{equation}

We start by forming ${\cal L}_2^{RP}$ with the clauses in
$\kappa(BS_G)$, namely $\neg$arm and $\neg$clog; we label the
clauses with $BS_P$ because they need to be supported by all states
in our belief state.  Next, we support each clause in ${\cal
L}_2^{RP}$ with the relevant effects from ${\cal E}_1$ to form
${\cal E}_1^{RP}$. For $\neg$clog we use persistence because it
supports $\neg$clog in all worlds described by $BS_P$ (this is an
example of positive interaction of worlds). For $\neg$arm the
relevant effects are the respective $\varphi^1$ from each Dunk
action.  We choose both effects to support $\neg$arm because we need
to support $\neg$arm in all worlds of $BS_P$, and each effect gives
support in only one world (this is an example of independence of
worlds).  We then insert the actions associated with each chosen
effect into ${\cal A}_1^{RP}$ with the appropriate label indicating
the worlds where it was {\em needed}, which in general is fewer
worlds than where it is {\em reachable} (i.e. it is always the case
that $\ell^{RP}_r(\cdot) \models \ell_r(\cdot)$). Next we form
${\cal L}_1^{RP}$ with the execution preconditions of actions in
${\cal A}_1^{RP}$ and antecedents of effects in ${\cal E}_1^{RP}$,
which are $\neg$clog, inP1, and inP2, labelled with all worlds where
an action or effect needed them.  In the same fashion as level two,
we support the literals at level one, using persistence for inP1 and
inP2, and Flush for $\neg$clog.  We stop here, because we have
supported all clauses at level one.

For the general case, extraction starts at the level $b$ where
$BS_i$ is first reachable from $BS_P$. The first relaxed plan
layers we construct are ${\cal A}^{RP}_{b-1}, {\cal E}^{RP}_{b-1},
{\cal L}^{RP}_{b}$, where ${\cal L}^{RP}_{b}$ contains all clauses
$C \in \kappa(BS_i)$, labelled as $\ell^{RP}_k(C) = BS_P$.

For each level $r$, $1 \leq r \leq b$, we support each clause in
${\cal L}^{RP}_{r}$ by choosing relevant effects from ${\cal
E}_{r-1}$ to form ${\cal E}^{RP}_{r-1}$.  An effect $\varphi^j(a)$
is relevant if it is reachable in some of the worlds where we need
to support $C$ (i.e. $\ell_{r-1}(\varphi^j(a)) \wedge
\ell^{RP}_{r}(C) \not= \perp$) and the consequent gives a literal $l
\in C$.  For each clause, we have to choose enough supporting
effects so that the chosen effect worlds are a superset of the
worlds we need to support the clause, formally:
\begin{equation}\label{confrplitk}\notag
\forall_{C \in {\cal L}^{RP}_{r}} \ell^{RP}_r(C) \models
\left(\bigvee_{\substack{\varphi^j(a):  l \in \varepsilon^j(a),\\
l \in C,\\ \varphi^j(a) \in {\cal E}_{r-1} }}
\ell_{r-1}^{RP}(\varphi^j(a))\right)
\end{equation}

We think of supporting a clause in a set of worlds as a set cover
problem where effects cover subsets of worlds. Our algorithm to
cover the worlds of a clause with worlds of effects is a variant of
the well known greedy algorithm for set cover \citep{GSetCover}. We
first choose all relevant persistence effects that can cover worlds,
then choose action effects that cover the most new worlds. Each
effect we choose for support is added to ${\cal E}^{RP}_{r-1}$ and
labelled with the new worlds it covered for $C$. Once all clauses in
${\cal L}^{RP}_{r}$ are covered, we form the action layer ${\cal
A}^{RP}_{r-1}$ as all actions that have an effect in ${\cal
E}^{RP}_{r-1}$.  The actions in ${\cal A}^{RP}_{r-1}$ are labelled
to indicate all worlds where any of their effects were labelled in
${\cal E}^{RP}_{r-1}$. \normalsize

We obtain the next subgoal layer, ${\cal L}^{RP}_{r-1}$, by adding
literals from the execution preconditions of actions in ${\cal
A}^{RP}_{r-1}$ and antecedents of effects in ${\cal E}^{RP}_{r-1}$.
Each literal $l \in {\cal L}^{RP}_{r-1}$ is labelled to indicate all
worlds where any action or effect requires $l$. We support the
literals in ${\cal L}^{RP}_{r-1}$ in the same fashion as ${\cal
L}^{RP}_{r}$. We continue to support literals with effects, insert
actions, and insert action and effect preconditions until we have
supported all literals in ${\cal L}^{RP}_{1}$.

Once we get a relaxed plan, the relaxed plan heuristic,
$h_{RP}^{LUG}(BS_i)$, is the summation of the number of actions in
each action layer, formally:

\begin{equation}\label{lugrp}\notag
h^{LUG}_{RP}(BS_i) = \sum\limits_{i = 0}^{b-1} \mid{{\cal
A}^{RP}_i}\mid
\end{equation}

Thus in our CBTC example we have $h^{LUG}_{RP}(BS_G) = 3$.  Notice
that if we construct the $LUG$ without mutexes for CBTC we reach the
goal after two layers.  If we had included mutexes the $LUG$, then
it would reach the goal after three layers.  The way we use mutexes
will not change our relaxed plan because we do not use mutexes to
influence relaxed plan extraction.  Mutexes only help to identify
when a the belief state $BS_i$ is not reachable from $BS_P$.

\section{Empirical Evaluation: Setup}

This section presents our implementation of the $\caltalt$ and
$POND$ planners and the domains we use in the experiments.  All
tests were run in Linux on an x86 machine with a 2.66GHz P4
processor and 1GB RAM with a timeout of 20 minutes. Both \caltalt\
and $POND$ used a heuristic weight of five in the, respective, A*
and AO* searches. We compare with the competing approaches (CGP,
SGP, GPT v1.40, MBP v0.91, KACMBP, YKA, and CFF) on several domains
and problems.  Our planners and all domain and problem files for all
of the compared planners can be found in the online appendix.

\begin{figure}[t]
\center{\scalebox{.5}{
\includegraphics{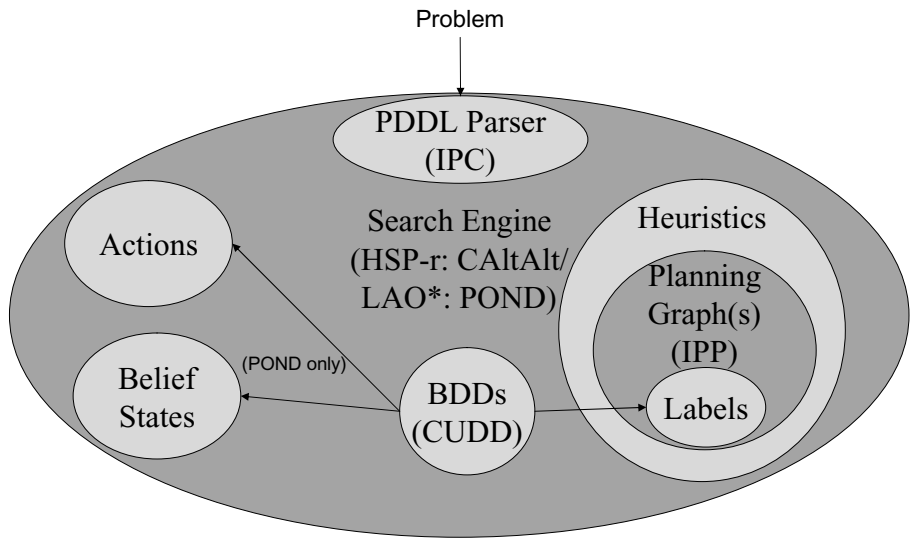}}} \vspace*{-1cm}\caption{The
implementations of \caltalt\ and $POND$ rely on many existing
technologies. The search engine is guided by heuristics extracted
from planning graphs.\label{Carch}}
\end{figure}

\subsection{Implementation}

The implementation of \caltalt\ uses several off-the-shelf planning
software packages.  Figure \ref{Carch} shows a diagram of the system
architecture for \caltalt\ and $POND$.  While \caltalt\ extends the
name of $AltAlt$, it relies on a limited subset of the
implementation. The components of \caltalt\ are the IPC parser for
PDDL 2.1 (slightly extended to allow uncertain initial conditions),
the HSP-r search engine \citep{bonet99planning}, the IPP planning
graph \citep{ipp}, and the CUDD BDD package \citep{CUDD} to
implement the $LUG$ labels. The custom parts of the implementation
include the action representation, belief state representation,
regression operator, and the heuristic calculation.

The implementation of $POND$ is very similar to \caltalt\ aside from
the search engine, and state and action representation.  $POND$ also
uses the IPP source code for planning graphs.  $POND$ uses modified
LAO* \citep{hansen-zilberstein:2001a} source code from Eric Hansen
to perform AO* search, and CUDD \citep{CUDD} to represent belief
states and actions.   Even with deterministic actions it is possible
to obtain cycles from actions with observations because we are
planning in belief space. $POND$ constructs the search graph as a
directed acyclic graph by employing a cycle-checking algorithm. If
adding a hyper-edge to the search graph creates a cycle, then the
hyper-edge cannot represent an action in a strong plan and is hence
not added to the graph.

\subsection{Domains}

Table \ref{domainspecs} shows some of the relative features of the
different problems we used to evaluate our approach.  The table
shows the number of initial states, goal literals, fluents, actions,
and optimal plan lengths.  This can be used as a guide to gauge the
difficulty of the problems, as well as our performance.

\begin{table}[t]
\center\scalebox{.95}{
\begin{tabular}{|c||c|c|c|c|c|c|c|} \hline
Problem & Initial & Goal  & Fluents & Causative & Observational
 & Optimal  & Optimal
\\
& States &  Literals & &  Actions & Actions &
 Parallel &  Serial
\\
\hline  Rovers1 & 1 & 1 & 66 & 88 & 0 \{12\} & 5 \{5\} & 5 \{5\}
\\ Rovers2 & 2 & 1& 66 & 88 & 0 \{12\} & 8 \{7\} & 8 \{7\}
\\ Rovers3 & 3 & 1 & 66 & 88 & 0 \{12\} & 10 \{?\} & 10 \{8\}
\\ Rovers4& 4 & 1 & 66 & 88 & 0  \{12\} & 13 \{?\} & 13 \{10\}
\\ Rovers5 & 16 & 3 & 71 & 97 & 0  \{12\} & ? \{?\} & 20 \{?\}
\\ Rovers6 & 12 & 3 & 119 & 217 & 0 \{18\}& ? \{?\}& ? \{?\} \\
\hline  Logistics1 & 2 & 1 & 29 & 70 & 0 \{10\}& 6 \{6\} & 9 \{7\}
\\ Logistics2 & 4 & 2 & 36 & 106 & 0 \{20\} & 6 \{?\}& 15 \{12\}
\\ Logistics3 & 2 & 1 & 58 & 282 & 0 \{21\}  & 8 \{?\}& 11 \{8\}
\\ Logistics4 & 4 & 2 & 68 & 396 & 0 \{42\} & 8 \{?\}& 18 \{?\}
\\ Logistics5 & 8 & 3 & 78 & 510 & 0 \{63\}& ? \{?\} & 28 \{?\} \\
\hline  BT($n$) & $n$ & 1 & $n$+1 & $n$ & 0 \{$n$\} & 1 \{1\} & $n$ \{$n$-1\}   \\
\hline BTC($n$) & $n$ & 1 & $n$+2 & $n$+1 & 0 \{$n$\} & 2$n$-1 \{2\} & 2$n$-1 \{$n$-1\}\\
\hline CubeCenter($n$) & $n$$^3$ & 3 & 3$n$ & 6 & 0 & (3$n$-3)/2 & (9$n$-3)/2 \\
\hline Ring($n$) & $n$3$^n$ & $n$ & 4$n$ & 4 & 0 & 3$n$-1 & 3$n$-1\\
\hline
\end{tabular}}
\caption{\label{domainspecs} Features of test domains and problems -
Number of initial states, Number of goal literals, Number of
fluents, Number of causative actions, Number of Observational
Actions, Optimal number of parallel plan steps, Optimal number of
serial plan steps.  Data for conditional versions of domains is in
braces; plan lengths for conditional plans are maximum conditional
branch length. }
\end{table}

\paragraph{Conformant Problems}

In addition to the standard domains used in conformant
planning--such as Bomb-in-the-Toilet, Ring, and Cube Center, we also
developed two new domains Logistics and Rovers.  We chose these new
domains because they have more difficult subgoals, and have many
plans of varying length.

The Ring domain involves a ring of $n$ rooms where each room is
connected to two adjacent rooms.  Each room has a window which can
be open, closed, or locked.  The goal is to have every window
locked. Initially, any state is possible -- we could be in any room
and each window could be in any configuration.  There are four
actions: move right, move left, close the window in the current
room, and lock the window in the current room.  Closing a window
only works if the window is open, and locking a window only works if
the window is closed.  A good conformant plan involves moving in one
direction closing and locking the window in each room.

The Cube Center domain involves a three-dimensional grid (cube)
where there are six actions -- it is possible to move in two
directions along each dimension.  Each dimension consists of $n$
possible locations.  Moving in a direction along which there are no
more grid points leaves one in the same position. Using this
phenomena, it is possible to localize in each dimension by
repeatedly moving in the same direction. Initially it is possible to
be at any location in the cube and the goal is to reach the center.
A good conformant plan involves localizing in a corner and then
moving to the center.

The Rovers domain is a conformant adaptation of the analogous domain
of the classical planning track of the International Planning
Competition \citep{IPC}. The added uncertainty to the initial state
uses conditions that determine whether an image objective is visible
from various vantage points due to weather, and the availability of
rock and soil samples. The goal is to upload an image of an
objective and some rock and soil sample data.  Thus a conformant
plan requires visiting all of the possible vantage points and taking
a picture, plus visiting all possible locations of soil and rock
samples to draw samples.

The first five Rovers problems have 4 waypoints.  Problems one
through four have one through four locations, respectively, at
which a desired imaging objective is possibly visible (at least
one will work, but we don't know which one).  Problem 5 adds some
rock and soil samples as part of the goal and several waypoints
where one of each can be obtained (again, we don't know which
waypoint will have the right sample). Problem 6 adds two more
waypoints, keeps the same goals as Problem 5 and changes the
possible locations of the rock and soil samples.  In all cases the
waypoints are connected in a tree structure, as opposed to
completely connected.

The Logistics domain is a conformant adaptation of the classical
Logistics domain where trucks and airplanes move packages.  The
uncertainty is the initial locations of packages.  Thus, any
actions relating to the movement of packages have a conditional
effect that is predicated on the package actually being at a
location.  In the conformant version, the drivers and pilots
cannot sense or communicate a package's actual whereabouts. The
problems scale by adding packages and cities.

The Logistics problems consist of one airplane, and cities with an
airport, a post office, and a truck. The airplane can travel
between airports and the trucks can travel within cities.  The
first problem has two cities and one package that could start at
either post office, and the goal is to get the package to the
second city's airport.  The second problem adds another package at
the same possible starting points and having the same destination.
The third problem has three cities with one package that could be
at any post office and has to reach the third airport.  The fourth
problem adds a second package to the third problem with the same
starting and ending locations.  The fifth problem has three cities
and three packages, each at one of two of the three post offices
and having to reach different airports.

\paragraph{Conditional Problems}

For conditional planning we consider domains from the literature:
Bomb-in-the-Toilet with sensing BTS, and Bomb-in-the-Toilet with
clogging and sensing BTCS. We also extend the conformant Logistics
and Rovers to include sensory actions.

The Rovers problem allows for the rover, when it is at a
particular waypoint, to sense the availability of image, soil, or
rock data at that location.  The locations of the collectable data
are expressed as one-of constraints, so the rover can deduce the
locations of collectable data by failing to sense the other
possibilities.

Logistics has observations to determine if a package at a location
exists, and the observation is assumed to be made by a driver or
pilot at the particular location.  Since there are several drivers
and a pilot, different agents make the observations. The
information gained by the agents is assumed to be automatically
communicated to the others, as the planner is the agent that has
all the knowledge.\footnote{This problem may be interesting to
investigate in a multi-agent planning scenario, assuming no global
communication (e.g. no radio dispatcher).}

\section{Empirical Evaluation: Inter-Heuristic Comparison}

We start by comparing the heuristic approaches within our
planners.  In the next section, we continue by describing how our
planners, using the best heuristics, compare against other state
of the art approaches.  In this section we intend to validate our claims that
belief space heuristics that measure overlap perform well across several domains.
We further justify using the $LUG$ over multiple planning graphs and applying mutexes to
improve heuristics in regression through pruning belief states.

We compare many techniques within
\caltalt\ and $POND$ on our conformant planning domains, and in
addition we test the heuristics in $POND$ on the conditional
domains.  Our performance metrics include the total planning time
and the number of search nodes expanded. Additionally, when
discussing mutexes we analyze planning graph construction time.
We proceed by showing how the heuristics perform in \caltalt\, and
then how various mutex computation schemes for the $LUG$ can
affect performance. Then we present how $POND$ performs with the
different heuristics in both conformant and conditional domains,
explore the effect of sampling a proportion of worlds to build
$SG^1$, $MG$, and $LUG$ graphs, and compare the heuristic
estimates in $POND$ to the optimal plan length to gauge heuristic
accuracy. We finish with a summary of important conclusions.

We only compute mutexes in the planning graphs for \caltalt\
because the planning graph(s) are only built once in a search
episode and mutexes help prune the inconsistent belief states
encountered in regression search.  We abstain from computing
mutexes in $POND$ because in progression we build new planning
graphs for each search node and we want to keep graph computation
time low.  With the exception of our discussion on sampling worlds
to construct the planning graphs, the planning graphs are
constructed deterministically.  This means that the single graph
is the unioned single graph $SG^U$, and the $MG$ and $LUG$ graphs
are built for all possible worlds.

\subsection{\caltalt} The results for \caltalt\ in the conformant
Rovers, Logistics, BT, and BTC domains, in terms of total time and
number of expanded search nodes, are presented in Table \ref{tab1}.
We show the number of expanded nodes because it gives an indication
of how well a heuristic guides the planner. The total time captures
the amount of time computing the heuristic and searching. A high
total time with a high number of search nodes indicates a poor
heuristic, and a high total time and low number of search nodes
indicates an expensive but informed heuristic.

We do not discuss the Ring and Cube Center domains for \caltalt\
because it cannot solve even the smallest instances.
Due to implementation details the planner performs very
poorly when domains have actions with several conditional effects
and hence does not scale.  The trouble stems from a weak
implementation for bringing general propositional formulas (obtained by
regression with several conditional effects) into CNF.

\begin{table}[t]
\center\scalebox{.8}{
\begin{tabular}{|c||c|c||c|c|c|c|} \hline
Problem & $h_0$ & $h_{card}$ &    $h^{SG}_{RP}$ &  $h^{MG}_{m-RP}$&    $h^{MG}_{RPU}$&  $h^{LUG(FX)}_{RP}$  \\
\hline  Rovers 1 & 2255/5 &   18687/14 & \bold{543/5} & 542/5&
\bold{185/5}& 15164/5
\\
2 & 49426/8   & TO  & 78419/8 & \bold{8327/8}& 29285/9 & 32969/8
\\
3 & TO   & - & 91672/10 & 20162/10 & \bold{2244/11} &16668/10
\\
4 & -   & - & TO & 61521/16 & \bold{3285/15} &31584/13
\\
5 & -   & -  & - & TO & TO & TO
\\
6 & -  & -  & - & - & - & -
\\
\hline Logistics 1 & 1108/9 &   4268/9 &  198/9 & \bold{183/9} &
1109/9 & 1340/9
\\
2 & TO &  TO  & 7722/15 & 15491/15 & 69818/19 & 18535/15
\\
3 & - &  - & 3324/14 & 70882/14 & TO & 16458/15
\\
4 & - &  - &\bold{141094/19}& TO & - & \bold{178068/19}
\\
5 & - &  - & TO & - & - & TO
\\
\hline BT 2 & 19/2  & 14/2 & 18/2 & 20/2 & 21/2 & 12/2
\\
10 & 4837/10  & \bold{56/10} & 5158/10 & 8988/10 &  342/10 &
\bold{71/10}
\\
20 & TO  & \bold{418/20} & TO & TO & 2299/20 & \bold{569/20}
\\
30 & -  & \bold{1698/30} & - & - &9116/30 &\bold{2517/30}
\\
40 & -   & \bold{5271/40} & - & - & 44741/40 & \bold{7734/40}
\\
50 & -   & \bold{12859/50} & - & - & TO &\bold{18389/50}
\\ 60 & -   & \bold{26131/60}& - & - & - & \bold{37820/60}
\\
70 & -  & \bold{48081/70}  & - & - & - & \bold{70538/70}
\\
80 & -   & \bold{82250/80}  & - & - & - & \bold{188603/80}
\\
\hline BTC 2 & 30/3  & 16/3  & 16/3 & 33/3 & 23/3 & 18/3
\\
10 & 15021/19   & \bold{161/19} & 15679/19 & 41805/19 &
\bold{614/19} & 1470/19
\\
20 & TO &  \bold{1052/39}  & TO & TO & \bold{2652/39} & 51969/39
\\
30 & - &   \bold{3823/59} & - & - & \bold{9352/59} & 484878/59
\\
40 & - &   \bold{11285/79}& - & - & \bold{51859/79} & TO
\\
50 & - &  \bold{26514/99}& - & - & TO & -
\\
60 & - &   \bold{55687/119}& - & - & - & -
\\
70 & - &   \bold{125594/140} & - & - & - & -
\\
\hline
\end{tabular}}
\caption{\label{tab1} Results for \caltalt\ for conformant Rovers,
Logistics, BT, and BTC. The data is Total Time / \# Expanded
Nodes, ``TO'' indicates a time out (20 minutes) and ``-''
indicates no attempt.}
\end{table}

We describe the results from left to right in Table \ref{tab1},
comparing the different planning graph structures and relaxed plans
computed on each planning graph.  We start with the non-planning
graph heuristics $h_0$ and $h_{card}$.  As expected, $h_0$,
breadth-first search, does not perform well in a large portion of
the problems, shown by the large number of search nodes and
inability to scale to solve larger problems. We notice that with the
$h_{card}$ heuristic performance is very good in the BT and BTC
problems (this confirms the results originally seen by
\citeR{cimatti01conformant}). However, $h_{card}$ does not perform
as well in the Rovers and Logistics problems because the size of a
belief state, during planning, does not necessarily indicate that
the belief state will be in a good plan.  Part of the reason
$h_{card}$ works so well in some domains is that it measures
knowledge, and plans for these domains are largely based on
increasing knowledge.  The reason $h_{card}$ performs poorly on
other domains is that finding causal support (which it does not
measure) is more important than knowledge for these domains.

Next, for a single planning graph ($SG^U$), \caltalt\ does
reasonably well with the $h_{RP}^{SG}$ heuristic in the Rovers and
Logistics domains, but fails to scale very well on the BT and BTC
domains. Rovers and Logistics have comparatively fewer initial
worlds than the BT and BTC problems. Moreover the deterministic
plans, assuming each initial state is the real state, are somewhat
similar for Rovers and Logistics, but mostly independent for BT and
BTC. Therefore, approximating a fully observable plan with the
single graph relaxed plan is reasonable when plans for achieving the
goal from each world have high positive interaction. However,
without high positive interaction the heuristic degrades quickly
when the number of initial worlds increases.

With multiple planning graphs, \caltalt\ is able to perform better
in the Rovers domain, but takes quite a bit of time in the
Logistics, BT, and BTC domains.  In Rovers, capturing distance
estimates for individual worlds and aggregating them by some means
tends to be better than aggregating worlds and computing a single
distance estimate (as in a single graph). In Logistics, part of
the reason computing multiple graphs is so costly is that we are
computing mutexes on each of the planning graphs.  In BT and BTC,
the total time increases quickly because the number of planning
graphs, and number of relaxed plans for every search node increase
so much as problems get larger.

Comparing the two multiple graph heuristics\footnote{We show
$h_{s-RP}^{MG}$ with $POND$.} in \caltalt\, namely $h_{m-RP}^{MG}$
and $h_{RPU}^{MG}$, we can see the effect of our choices for state
distance aggregation.  The $h_{m-RP}^{MG}$ relaxed plan heuristic
aggregates state distances, as found on each planning graph, by
taking the maximum distance.  The $h_{RPU}^{MG}$ unions the
relaxed plans from each graph, and counts the number of actions in
the unioned relaxed plan.  As with the single graph relaxed plan,
the $h_{m-RP}^{MG}$ relaxed plan essentially measures one state to
state distance; thus, performance suffers on the BT and BTC
domains.  However, using the unioned relaxed plan heuristic, we
capture the independence among the multiple worlds so that we
scale up better in BT and BTC. Despite the usefulness of the
unioned relaxed plan, it is costly to compute and scalability is
limited, so we turn to the $LUG$ version of this same measure.

With the $LUG$, we use the $h_{RP}^{LUG(FX)}$ heuristic in \caltalt.
This heuristic uses a $LUG$ with full cross-world mutexes (denoted
by $FX$).  As in the similar $h_{RPU}^{MG}$ heuristic, measuring
overlap is important, and improving the speed of computing the
heuristic tends to improve the scalability of \caltalt. While
\caltalt\ is slower in the Rovers and BTC domains when using the
$LUG$, we note that it is because of the added cost of computing
cross-world mutexes -- we are able to improve the speed by relaxing
the mutexes, as we will describe shortly.

\begin{table}[t]
\center \scalebox{.73 }{ \rotatebox{90}{
\begin{tabular}{|c||c|c|c|c|c|c|c|c|} \hline
Problem &  $h^{LUG (NX)}_{RP}$& $h^{LUG (StX)}_{RP}$& $h^{LUG (DyX)}_{RP}$& $h^{LUG (FX)}_{RP}$& $h^{LUG (DyX-SX)}_{RP}$& $h^{LUG (DyX-IX)}_{RP}$& $h^{LUG (FX-SX)}_{RP}$ & $h^{LUG (FX-IX)}_{RP}$ \\
\hline Rovers 1 & \bold{13}/1112/51 & 19/1119/51 & 15453/89/6 &
15077/87/6 & 15983/87/6 & 15457/87/6 & 15098/86/6 &
15094/\bold{85}/\bold{6}
\\
2 & 20/904/41 & \bold{16}/903/41 & 13431/\bold{138}/\bold{8} &
32822/147/8 & 10318/139/8 & 10625/134/8 & 10523/138/8 & 14550/138/8
\\
3 & \bold{13}/8704/384 & 17/8972/384 & 17545/185/10 & 16481/187/10 &
10643/185/10 & 11098/209/10 & 10700/191/10 &
11023/\bold{184}/\bold{10}
\\
4 & TO & TO & 32645/441/14 & 31293/291/14 & 14988/291/14 &
16772/291/14 & \bold{14726}/\bold{290}/\bold{14} & 16907/290/14
\\
5 & - & - & 698575/3569/45 & TO & 61373/3497/45 & 379230/3457/45 &
\bold{60985}/\bold{3388}/\bold{45} & 378869/3427/45
\\
6 & - & - & TO & - & \bold{217507}/3544/37 & 565013/3504/37 &
225213/\bold{3408}/\bold{37} & 588336/3512/37
\\
\hline Logistics 1 & \bold{5}/868/81 & 10/868/81 & 1250/117/9 &
1242/\bold{98}/\bold{9} & 791/116/9 & 797/117/9 & 796/115/9 &
808/115/9
\\
2 & \bold{10}/63699/1433 & 88/78448/1433 & 16394/622/15 &
18114/421/15 & 2506/356/15 & 7087/428/15 & 2499/\bold{352}/\bold{15}
& 6968/401/15
\\
3 & TO & TO & 17196/1075/15 & 16085/\bold{373}/\bold{15} &
10407/403/15 & 10399/408/15 & \bold{10214}/387/15 & 10441/418/15
\\
4 & - & - & \bold{136702}/1035/19 & 176995/1073/19 & 24214/648/19 &
71964/871/19 & 23792/\bold{642}/\bold{19} & 71099/858/19
\\
5 & - & - & TO & TO & \bold{52036}/2690/41 & 328114/4668/52 &
52109/\bold{2672}/\bold{41} & 324508/4194/52
\\
\hline  BT 2 & 1/34/2 & 0/13/2 & 0/13/2 & \bold{0/12/2} & 0/16/2 &
0/15/2 & 0/25/2 & 0/13/2
\\
10 & 4/72/10 & 4/\bold{56}/\bold{10} & \bold{13}/57/10 & 13/58/10 &
12/59/10 & 14/59/10 & 13/59/10 & 14/56/10
\\
20 & \bold{19}/452/20 & 22/448/20 & 120/453/20 & 120/449/20 &
102/450/20 & 139/454/20 & 105/\bold{444}/\bold{20} & 137/454/20
\\
30 & 62/1999/30 & \bold{59}/\bold{1981}/\bold{30} & 514/1999/30 &
509/2008/30 & 421/1994/30 & 600/2007/30 & 413/1986/30 & 596/2002/30
\\
40 & \bold{130}/6130/40 & 132/6170/40 & 1534/6432/40 & 1517/6217/40
& 1217/6326/40 & 1822/6163/40 & 1196/\bold{6113}/\bold{40} &
1797/6127/40
\\
50 & \bold{248}/\bold{14641}/\bold{50} & 255/14760/50 &
3730/14711/50 & 3626/14763/50 & 2866/14707/50 & 4480/14676/50 &
2905/14867/50 & 4392/14683/50
\\
60 & \bold{430}/30140/60 & 440/\bold{29891/60} & 7645/30127/60 &
7656/30164/60 & 5966/30017/60 & 9552/30337/60 & 5933/30116/60 &
9234/29986/60
\\
70 & \bold{680}/\bold{55202/70} & 693/55372/70 & 15019/55417/70 &
14636/55902/70 & 11967/55723/70 & 18475/55572/70 & 11558/55280/70 &
18081/55403/70
\\
80 & \bold{1143}/135760/80 & 1253/140716/80 & 26478/132603/80 &
26368/162235/80 & 21506/136149/80 & 32221/\bold{105654/80} &
21053/139079/80 & 32693/109508/80
\\
\hline BTC 2 & 0/62/3 & 1/16/3 & \bold{0}/15/3 & 4/\bold{14/3} &
0/16/3 & 1/14/3 & 1/13/3 & 2/14/3
\\
10 & 4/93/19 & \bold{4}/77/19 & 14/78/19 & 1388/82/19 & 13/76/19 &
16/\bold{75/19} & 14/75/19 & 440/81/19
\\
20 & \bold{21}/546/39 & 32/\bold{545/39} & 139/553/39 & 51412/557/39
& 105/546/39 & 140/549/39 & 110/555/39 & 19447/568/39
\\
30 & \bold{58}/2311/59 & 61/2293/59 & 543/2288/59 & 482578/2300/59 &
427/2294/59 & 606/2300/59 & 444/\bold{2287/59} & 199601/2401/59
\\
40 & \bold{133}/6889/79 & 149/6879/79 & 1564/6829/79 & TO &
1211/\bold{6798}/79 & 1824/6816/79 & 1253/6830/79 & 1068019/6940/79
\\
50 & \bold{260/15942/99} & 261/16452/99 & TO & - & 2890/16184/99 &
4412/16414/99 & 2926/16028/99 & TO
\\
60 & \bold{435/32201/119} & 443/32923/119 & - & - & 6045/32348/119 &
9492/32350/119 & 6150/32876/119 & -
\\
70 & \bold{742}/62192/139 & 745/\bold{61827/139} & - & - & TO & TO &
TO & -
\\
\hline
\end{tabular}}} \caption{\label{tab2} Results for \caltalt\ using $h^{LUG}_{RP}$
 with mutex schemes. The data is Graph Construction Time (ms)/All
Other Time (ms)/\# Expanded Nodes, ``TO'' indicates a time out (20
minutes) and ``-'' indicates no attempt.} \vspace{-1.6cm}
\end{table}

\subsection{Mutexes} Mutexes are used to help determine when a belief state is unreachable.
Mutexes improve the pruning power of heuristics by accounting for negative interactions.
The mutexes are only used to improve our heuristics,
so it is reasonable to compute only a subset of the mutexes.
We would like to know which mutexes are the most cost effective because
the number of possible mutexes we can find is quite
large.

We can use several schemes to compute a subset of the mutexes. The
schemes combine different types of mutexes with types of cross-world
checking. The mutex types are: computing no mutexes (NX), computing
only static interference mutexes (StX), computing (StX) plus
inconsistent support and competing needs mutexes -- dynamic mutexes
(DyX), and computing (DyX) plus induced mutexes -- full mutexes
(FX). The cross-world checking (see appendix B) reduction schemes
are: computing mutexes across same-worlds (SX) and computing mutexes
across pairs of worlds in the intersection (conjunction) of element
labels (IX).

Table \ref{tab2} shows that within \caltalt, using the relaxed plan
heuristic and changing the way we compute mutexes on the $LUG$ can
drastically alter performance. Often, the cross-world mutexes are so
numerous that building the $LUG$ takes too much time.  To see if we
could reduce graph construction overhead without hindering
performance, we evaluated $h_{RP}^{LUG}$ when the LUG is built (a)
considering all cross-world relations, for the schemes (NX), (StX),
(DyX), and (FX); and (b) same-world relations for the schemes
(DyX-SX) and (FX-SX), and (c) cross-world relations for all possible
worlds pairs in the intersection of element's labels (DyX-IX) and
(FX-IX).

The results show that simpler problems like BT and BTC do not
benefit as much from advanced computation of mutexes beyond static
interference. However, for the Rovers and Logistics problems,
advanced mutexes play a larger role.  Mainly, interference,
competing needs, and inconsistent support mutexes are important. The
competing needs and inconsistent support mutexes seem to have a
large impact on the informedness of the guidance given by the $LUG$,
as scalability improves most here. Induced mutexes don't improve
search time much, and only add to graph computation time. A possible
reason induced mutexes don't help as much in these domains is that
all the actions have at most two effects, an unconditional and
conditional effect.   Reducing cross-world mutex checking also helps
quite a bit. It seems that only checking same-world mutexes is
sufficient to solve large problems. Interestingly, the $MG$ graphs
compute same-world interference, competing needs, and inconsistent
support mutexes within each graph, equating to the same scenario as
(DyX-SX), however, the LUG provides a much faster construction time,
evidenced by the $LUG$'s ability to out-scale $MG$.

\subsection{$POND$}  We show the total time and the number of
expanded nodes for $POND$ solving the conformant problems (including
Ring and Cube Center) in Table \ref{tab3}, and for $POND$ solving
the conditional problems in Table \ref{tab5}. As with \caltalt\ we
show the total time and number of expanded nodes for each test. We
also add the $h_{s-RP}^{MG}$ heuristic, not implemented in \caltalt,
that takes the summation of the values of relaxed plans extracted
from multiple planning graphs.  We do not compute mutexes on any of
the planning graphs used for heuristics in $POND$ mainly because we
build planning graphs for each search node.  We proceed by first
commenting on the performance of $POND$, with the different
heuristics, in the conformant domains, then discuss the conditional
domains.

\begin{table}[t]
\center \scalebox{.75 }{
\begin{tabular}{|c||c|c||c||c|c|c||c|} \hline
Problem & $h_0$ & $h_{card}$ & $h^{SG}_{RP}$ & $h^{MG}_{m-RP}$ &$h^{MG}_{s-RP}$ &$h^{MG}_{RPU}$ & $h^{LUG}_{RP}$  \\
\hline Rovers 1 & 540/36 & 520/21 & 590/6 & 580/6 & 580/6 & 580/6
& 590/6
\\
2 & 940/249 & 790/157 & 700/15 & 1250/32 & 750/10 & 830/13 &
680/11
\\
3 & 3340/1150 & 2340/755 & 3150/230 & 3430/77 & 1450/24 & 1370/23
& 850/16
\\
4 & TO & 14830/4067 & 13480/1004 & 10630/181 & 7000/163 & 2170/34
& 1130/28
\\
5 & - & TO & TO & 85370/452 & 12470/99 & 31480/73 & 2050/36
\\
6 & - & - & - & 180890/416 & 15780/38 & 31950/73 & 9850/147
\\
\hline Logistics 1   & 560/169 & 530/102 & 680/46 & 970/58 &
730/21 & 650/9 & 560/9\\
2 & TO & TO & TO & 2520/32 & 6420/105 & 2310/20 & 910/15
\\
3 & - & - & - & 27820/927 & 4050/83 & 2000/15 & 1130/14
\\
4 & - & - & - & 5740/27 & 29180/211 & 53470/382 & 3180/46
\\
5 & - & - & - & 42980/59 & 51380/152 & 471850/988 & 6010/42
\\
\hline  BT 2 & 450/3 & 460/2 & 460/3 & 450/2 & 450/2 & 500/2 &
460/2
\\
10 & 760/1023 & 590/428 & 1560/1023 & 6200/428 & 820/10 & 880/10 &
520/10
\\
20 & TO & TO & TO & TO & 6740/20 & 6870/20 & 1230/20
\\
30 & - & - & - & - & 41320/30 & 44260/30 & 4080/30
\\
40 & - & - & - & - & 179930/40 & 183930/40 & 11680/40
\\
50 & - & - & - & - & 726930/50 & 758140/50 & 28420/50
\\
60 & - & - & - & - & TO & TO & 59420/60
\\
70 & - & - & - & - & - & - & 113110/70
\\
80 & - & - & - & - & - & - & 202550/80
\\
\hline BTC 2 &  460/5 & 460/4 & 450/5 & 460/4 & 460/3 & 470/3 &
460/3
\\
10 & 1090/2045 & 970/1806 & 3160/2045 & 18250/1806 & 980/19 &
990/19 & 540/19
\\
20 & TO & TO & TO & TO & TO & 9180/39 & 1460/39
\\
30 & - & - & - & - & - & 54140/59 & 4830/59
\\
40 & - & - & - & - & - & 251140/79 & 14250/79
\\
50 & - & - & - & - & - & 1075250/99 & 34220/99
\\
60 & - & - & - & - & - & TO & 71650/119
\\
70 & - & - & - & - & - & - & 134880/139
\\
\hline CubeCenter 3 & 10/184 & 30/14 & 90/34 & 1050/61 & 370/9 &
0430/11 & 70/11
\\
5 & 180/3198 & 20/58 & 3510/1342 & 60460/382 & 11060/55 & 14780/82
& 1780/205
\\
7 & 1940/21703 & 40/203 & 46620/10316 & TO & 852630/359 &
1183220/444 & 27900/1774
\\
9 & TO & 70/363 & 333330/46881 & - & TO & TO & 177790/7226
\\
11 & - & 230/1010 & TO & - & - & - & 609540/17027
\\
13 & - & 700/2594 & - & - & - & - & TO
\\
\hline Ring 2 & 20/15 & 20/7 & 30/15 & 80/8 & 80/7 & 80/8 & 30/8
\\
3 & 20/59 & 20/11 & 70/59 & 1500/41 & 500/8 & 920/19 & 70/10
\\
4 & 30/232 & 20/15 & 350/232 & 51310/77 & 6370/11 & 19300/40 &
250/24
\\
5 & 160/973 & 20/19 & 2270/973 & TO & 283780/16 & TO & 970/44
\\
6 & 880/4057 & 30/23 & 14250/4057 & - & TO & - & 4080/98
\\
7 & 5940/16299 & 40/27 & 83360/16299 & - & - & - & 75020/574
\\
8 & 39120/64657 & 40/31 & 510850/64657 & - & - & - & 388300/902
\\
9 & 251370/261394 & 50/35 & TO & - & - & - & TO
\\
10 & TO & 70/39 & - & - & - & - & -
\\
 \hline
\end{tabular}}
\caption{\label{tab3}Results for $POND$ for conformant Rovers,
Logistics, BT, BTC, Cube Center, and Ring.  The data is Total Time
(ms)/\# Expanded Nodes, ``TO'' indicates a time out and ``-''
indicates no attempt.}
\end{table}

\begin{table}[t]
\center{\scalebox{.85}{
\begin{tabular}{|c||c|c||c||c|c|c||c|} \hline
Problem & $h_0$ & $h_{card}$ &  $h^{SG}_{RP}$ & $h^{MG}_{m-RP}$& $h^{MG}_{s-RP}$& $h^{MG}_{RPU}$  & $h^{LUG}_{RP}$ \\
\hline Rovers 1   & 550/36 & 480/21 & 580/6 & 570/6 & 570/6 &
580/6 &580/6
\\
2 & 1030/262 & 550/36 & 780/15 & 760/14 & 710/12 & 730/12 & 730/13
\\
3 & 1700/467 & 590/48 & 3930/248 & 830/15 & 830/15 & 910/17 &
810/16
\\
4 & 5230/1321 & 620/58 & 6760/387 & 1020/20 & 1040/21 & 1070/21 &
910/21
\\
5 & TO & TO & TO & 16360/175 & 11100/232 & 12810/209 & 7100/174
\\
6 & - & - & - & 31870/173 & 24840/159 & 30250/198 & 13560/174
\\
\hline Logistics 1 & 530/118 & TO & 740/46 & 580/10 & 570/10 &
600/10 & 570/10
\\
2 & TO & - & TO & 1630/30 & 1300/36 & 1360/36 & 1250/36
\\
3 & - & - & - & 1360/20 & 1250/19 & 1290/19 & 1210/19
\\
4 & - & - & - & 4230/59 & 3820/57 & 3940/57 & 4160/57
\\
5 & - & - & - & 27370/183 & 19620/178 & 20040/178 & 20170/178
\\
\hline  BT 2   & 460/5 & 460/3 & 450/3 & 460/3 & 450/3 & 470/3 &
460/3
\\
10 & TO & 470/19 & 111260/7197 & 970/19 & 970/19 & 1020/19 &
550/19
\\
20 & - & 510/39 & TO & 9070/39 & 9060/39 & 9380/39 & 1610/39
\\
30 & - & 620/59 & - & 52410/59 & 52210/59 & 55750/59 & 5970/59
\\
40 & - & 850/79 & - & 207890/79 & 206830/79 & 233720/79 & 17620/79
\\
50 & - & 1310/99 & - & 726490/99 & 719000/99 & TO & 43020/99
\\
60 & - & 2240/119 & - & TO & TO & - & 91990/119
\\
70 & - & 24230/139  & - & - & - & - & 170510/139
\\
80 & - &  45270/159 & - & - & - & - & 309940/159
\\
\hline BTC 2 & 450/6 & 460/3 & 470/5 & 470/3 & 460/3 & 470/3 &
470/3
\\
10 & TO & 480/19 & 271410/10842 & 1150/19 & 1140/19 & 1200/19 &
590/19
\\
20 & - & 510/39 & TO & 11520/39 & TO & 11610/39 & 1960/39
\\
30 & - & 660/59 & - & 62060/59 & - & 64290/59 & 6910/59
\\
40 & - & 970/79 & - & 251850/79 & - & 274610/79 & 19830/79
\\
50 & - & 1860/99 & - & 941220/99 & - & TO & 49080/99
\\
60 & - & 4010/119 & - & TO & - & - & 103480/119
\\
70 & - & 7580/139 & - & - & - & - & 202040/139
\\
\hline
\end{tabular}}}
\caption{\label{tab5} Results for $POND$ for conditional Rovers,
Logistics, BTS, BTCS. The data is Total Time (ms)/\# Expanded
Nodes, ``TO'' indicates a time out (20 minutes) and ``-''
indicates no attempt.}
\end{table}

In the conformant domains, $POND$ generally does better than
\caltalt.  This may be attributed in part to implementation-level
details.  $POND$ makes use of an existing (highly optimized) BDD
package for belief state generation in progression, but as
previously mentioned, \caltalt\ relies on a less optimized implementation
for belief state generation in regression.  As we will see in the
next section, regression planners that employ a more sophisticated
implementation perform much better, but could still benefit from our
heuristics.  Aside from a few differences that we will mention, we
see similar trends in the performance of the various heuristics in
both \caltalt\ and $POND$. Namely, the $NG$ and $SG$ heuristics have
limited ability to help the planner scale, the $MG$ heuristics help
the planner scale better but are costly, and the $LUG$ provides the
best scalability. The difference between the $MG$ and the $LUG$ are
especially pronounced in Cube Center and Ring, where the size of the
initial belief state is quite large as the instances scale.
Interestingly in Ring, breadth first search and the single graph
relaxed plan are able to scale due to reduced heuristic computation
time and the low branching factor in search. The $LUG$ is able to provide good
search guidance, but tends to take a long time computing heuristics in Ring.

We are also now able to compare the choices for aggregating the
distance measures from relaxed plans for multiple graphs.  We see
that taking the maximum of the relaxed plans, $h_{m-RP}^{MG}$, in
assuming positive interaction among worlds is useful in Logistics
and Rovers, but loses the independence of worlds in the BT and BTC
domains.  However, taking the summation of the relaxed plan values
for different worlds, $h_{s-RP}^{MG}$ is able to capture the
independence in the BT domain.  We notice that the summation does
not help $POND$ in the BTC domain; this is because we overestimate
the heuristic value for some nodes by counting the Flush action once
for each world when it in fact only needs to be done once (i.e. we
miss positive interaction). Finally, using the $h_{RPU}^{MG}$
heuristic we do well in every domain, aside from the cost of
computing multiple graph heuristics, because we account for both
positive interaction and independence by taking the overlap of
relaxed plans.  Again, with the $LUG$ relaxed plan, analogous to the
multiple graph unioned relaxed plan, $POND$ scales well because we
measure overlap and lower the cost of computing the heuristic
significantly.

The main change we see in using $POND$ versus \caltalt\ is that the
direction of search is different, so the $h_{card}$ heuristic
performs unlike before.  In the BT and BTC domains cardinality does
not work well in progression because the size of belief states does
not change as we get closer to the goal (it is impossible to ever
know which package contains the bomb). However, in regression we
start with a belief state containing all states consistent with the
goal and regressing actions limits our belief state to only those
states that can reach the goal through those actions.  Thus in
regression the size of belief states decreases, but in progression
remains constant.

The performance of $POND$ in the conditional domains exhibits
similar trends to the conformant domains, with a few exceptions.
Like the conformant domains, the $MG$ relaxed plans tend to
outperform the $SG$ relaxed plan, but the $LUG$ relaxed plan does
best overall.  Unlike the conformant domains, The $h_{m-RP}^{MG}$
performs much better in BTS and BTCS over BT and BTC partly because
the conditional plans have a lower average cost.  The $h_{card}$
heuristic does better in BTS and BTCS over BT and BTC because the
belief states actually decrease in size when they are partitioned by
sensory actions.

\subsection{Sampling Worlds} Our evaluations to this point have
considered the effectiveness of different heuristics, each
computed with respect to all possible worlds of a belief state.
While we would like to use as many of the possible worlds as we
can, we can reduce computation cost and hopefully still get
reasonable heuristics by considering a subset of the worlds.  Our
scheme for considering subsets of worlds in the heuristics is to
sample a single world ($SG^1$), or sample a given percentage of
the worlds and build multiple graphs, or the $LUG$.

\begin{figure}[t]
\center\hspace*{-.5cm}\scalebox{.54}{
\includegraphics{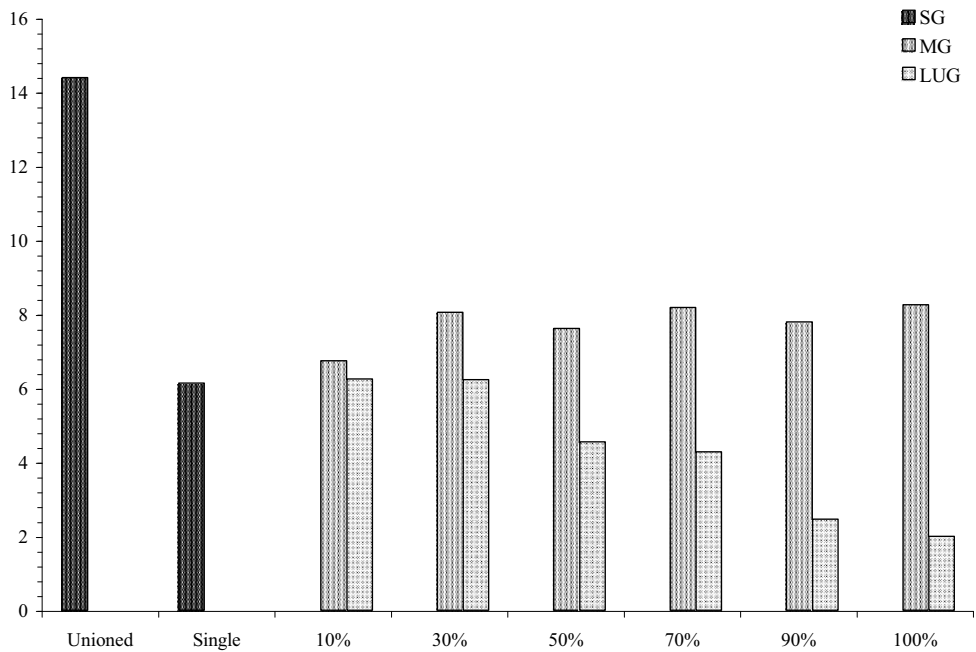}}
\caption{\label{approxTimes} Total Time (hours) for $POND$ to solve
all conformant and conditional problems when sampling worlds to use
in heuristic computation.}
\end{figure}

With these sampling approaches, we use the $h^{SG}_{RP}$,
$h^{MG}_{RPU}$, and $h^{LUG}_{RP}$ relaxed plans.  We build the $MG$
and $LUG$ for 10\%, 30\%, 50\%, 70\%, and 90\% of the worlds in each
belief state, sampled randomly.  In Figure \ref{approxTimes}, we
show the total time taken (ms) to solve every problem in our test
set (79 problems over 10 domains). Each unsolved problem contributed
20 minutes to the total time. For comparison we show the previously
mentioned heuristics: $h^{SG}_{RP}$ computed on a unioned single
graph $SG^U$, denoted as ``Unioned'' compared to the sampled single
graph $SG^1$ denoted as ``Single'', and $h^{MG}_{RPU}$ and
$h^{LUG}_{RP}$ computed for all worlds denoted as ``100\%''. The
total time for any heuristic that samples worlds is averaged over
ten runs.

There are two major points to see in Figure \ref{approxTimes}.
First, the $h^{SG}_{RP}$ heuristic is much more effective when
computed on $SG^1$ versus $SG^U$.  This is because the $SG^1$ is
less optimistic.  It builds a planning graph for a real world
state, as opposed to the union of literals in all possible world
states, as in $SG^U$. Respecting state boundaries and considering
only a single state is better than ignoring state boundaries to
naively consider all possible states.  However, as we have seen
with the $MG$ and $LUG$ heuristics, respecting state boundaries
and considering several states can be much better, bringing us to
the second point.

We see very different performance when using
more possible worlds to build multiple graphs compared to the
$LUG$.  We are better off using fewer worlds if we have to build
multiple graphs because they can become very costly as the number
of worlds increases.  In contrast, performance improves with more
possible worlds when we use the $LUG$.  Using more possible worlds
to compute heuristics is a good idea, but it takes a more
efficient substrate to exploit them.

\subsection{Accuracy}

\begin{figure}[t]
\center\hspace*{-.5cm}\scalebox{1.1}{\input{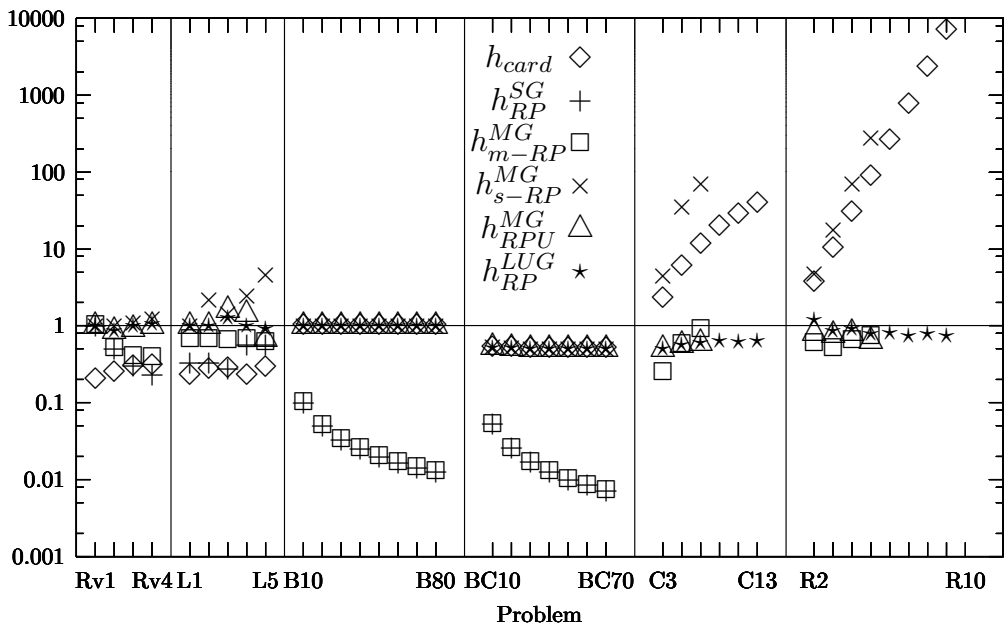} }
\caption{\label{hest} Ratio of heuristic estimates for distance
between $BS_I$ and $BS_G$ to optimal plan length. Rv = Rovers, L =
Logistics, B = BT, BC = BTC, C = Cube Center, R = Ring.}
\end{figure}

The heuristics that account for overlap in the possible worlds
should be more accurate than the heuristics that make an assumption
of full positive interaction or full independence. To check our
intuitions, we compare the heuristic estimates for the distance
between the initial belief state and the goal belief state for all
the heuristics used in conformant problems solved by $POND$. Figure
\ref{hest} shows the ratio of the heuristic estimate for $h(BS_I)$
to the optimal serial plan length $h^*(BS_I)$ in several problems.
The points below the line (where the ratio is one) are
under-estimates, and those above are over-estimates. Some of the
problem instances are not shown because no optimal plan length is
known.

We note that in all the domains the $h^{LUG}_{RP}$ and
$h^{MG}_{RPU}$ heuristics are very close to $h^*$, confirming our
intuitions. Interestingly, $h^{MG}_{s-RP}$ and $h^{MG}_{m-RP}$ are
both close to $h^*$ in Rovers and Logistics; whereas the former is
close in the BT and BTC problems, and the latter is close in
CubeCenter and Ring. As expected, assuming independence (using
summation) tends to over-estimate, and assuming positive interaction
(using maximization) tends to under-estimate. The $h^{SG}_{RP}$
heuristic tends to under-estimate, and in some cases (CubeCenter and
Ring) gives a value of zero (because there is an initial state that
satisfies the goal). The $h_{card}$ heuristic is only accurate in BT
and BTC, under-estimates in Rovers and Logistics, and over-estimates
in Cube Center and Ring.

The accuracy of heuristics is in some cases disconnected from
their run time performance.  For instance $h_{card}$ highly
overestimates in Ring and Cube Center, but does well because the
domains exhibit special structure and the heuristic is fast to
compute. On the other hand, $h^{LUG}_{RP}$ and $h^{MG}_{RPU}$ are
very accurate in many domains, but suffer in Ring and Cube Center
because they can be costly to compute.

\subsection{Inter-Heuristic Conclusions} Our findings fall into two
main categories: one, what are effective estimates for belief
state distances in terms of state to state distances, and two, how
we can exploit planning graphs to support the computation of these
distance measures.

In comparing ways to aggregate state distance measures to compute
belief state distances, we found that measuring no interaction as
in single graph heuristics tends to poorly guide planners,
measuring independence and positive interaction of worlds works
well in specific domains, and measuring overlap (i.e. a
combination of positive interaction and independence) tends to
work well in a large variety of instances.  By studying the
accuracy of our heuristics we found that in some cases the most
accurate were not the most effective.  We did however find that
the most accurate did best over the most cases.

Comparing graph structures that provide the basis for belief state
distance measures, we found that the heuristics extracted from the
single graph fail to systematically account for the independence
or positive interaction among different possible worlds.  Despite
this lack in the distance measure, single graphs can still
identify some structure in domains like Rovers and Logistics. To
more accurately reflect belief state distances, multiple graphs
reason about reachability for each world independently.  This
accuracy comes at the cost of computing a lot of redundant $MG$
structure and is limiting in instances with large belief states.
We can reduce the cost of the $MG$ structure by sampling worlds
used in its construction.  However planners are able to exhibit
better scalability by considering more worlds through optimizing
the representation of the redundant structure as in the $LUG$. The
improvement in scalability is attributed to lowering the cost of
heuristic computation, but retaining measures of multiple state
distances. The $LUG$ makes a trade-off of using an exponential
time algorithm for evaluation of labels instead of building an
exponential number of planning graphs.  This trade-off is
justified by our experiments.

\section{Empirical Evaluation: Inter-Planner Comparison}

We first compare \caltalt\ and $POND$ with several planners on our
conformant domains, then compare $POND$ with the conditional
planners on the conditional domains.  Our purpose in this section is to identify the advantages
of our techniques over the state of the art planners.
We end the section with a discussion of general conclusions drawn from the evaluation.

\begin{table}\footnotesize
\center \begin{tabular}{l|l|l|c|l|l} Planner & Search Space  &
Search Direction  & Conditional & Heuristic &
Implementation\\
\hline \caltalt\ & Belief Space & Backward && Planning Graph & C \\
$POND$& Belief Space & Forward &$\surd$ & Planning Graph & C \\
MBP & Belief Space & Forward/Backward  & $\surd$ & Cardinality & C \\
KACMBP & Belief Space & Forward  &  & Cardinality& C \\
CGP & Planning Graph & Backward  & & Planning Graph & Lisp \\
SGP & Planning Graph & Backward  &$\surd$  &Planning Graph & Lisp \\
GPT & Belief Space & Forward   & $\surd$ & State Space Plans & C \\
YKA & Belief Space & Backward  & $\surd$ & Cardinality & C \\
CFF & Belief Space & Forward   & & Planning Graph & C \\
\end{tabular}
\caption{\label{planner-comps} Comparison of planner features.}
\end{table}

\subsection{Conformant Planning}

Although this work is aimed at giving a general comparison of
heuristics for belief space planning, we also present a comparison
of the best heuristics within  \caltalt\ and $POND$ to some of the
other leading approaches to conformant planning.  Table
\ref{planner-comps} lists several features of the evaluated
planners, such as their search space, their search direction,
whether they are conditional, the type of heuristics, and the
implementation language.  Note, since each approach uses a different
planning representation (BDDs, GraphPlan, etc.), not all of which
even use heuristics, it is hard to get a standardized comparison of
heuristic effectiveness. Furthermore, not all of the planners use
PDDL-like input syntax; MBP, and KACMBP use AR encodings which may
give them an advantage in reducing the number of literals and
actions. We gave the MBP planners the same grounded and filtered
action descriptions that we used in \caltalt\ and $POND$. We also
tried, but do not report results, giving the MBP planners the full
set of ground actions without filtering irrelevant actions. It
appears that the MBP planners do not use any sort of action
pre-processing because performance was much worse with the full
grounded set of actions. Nevertheless, Table \ref{tab4} compares
MBP, KACMBP, GPT, CGP, YKA, and CFF with $h^{LUG (DyX-SX)}_{RP}$ in
\caltalt\ and $h^{LUG}_{RP}$ in $POND$ with respect to run time and
plan length.

\begin{table}[t]
\center\scalebox{.75}{
\begin{tabular}{|c||c|c||c|c||c||c||c|c|} \hline
Problem &  \caltalt\ & POND & MBP & KACMBP &   GPT & CGP &   YKA & CFF\\
& $h^{LUG (DyX-SX)}_{RPU} $ &  $h^{LUG}_{RP}$  & & & & & &\\
\hline  Rovers 1 & 16070/5 & 590/5 & 66/5& 9293/5 &  3139/5 & 70/5 &
1220/7 & 70/5
\\
2 & 10457/8& 680/9 & 141/8& 9289/15 &   4365/8 & 180/8 &
   2050/10 & 30/8
\\
3 & 10828/10& 850/11 &484/10 & 9293/16 &   5842/10 & 460/10 &
1740/12 &10/10
\\
4 & 15279/13& 1130/16 &3252/13 & 9371/18 &   7393/13 & 1860/13&
2010/16 &10/13
\\
5 & 64870/29& 2050/25 & OoM & 39773/40 &   399525/20 & OoM & 7490/27
& 18/22
\\
6 & 221051/25& 8370/25 & 727/32& TO &   TO &
  - & 24370/26 & 21/23
\\
\hline  Logistics 1 & 907/9& 560/9 &37/9 & 127/12 &   916/9 & 60/6 &
250/13 & 10/9
\\
2 & 2862/15& 910/15 &486/24 & 451/19 &   1297/15& 290/6 &   670/19 &
12/15
\\
3 & 10810/15& 1130/14&408/14 & 1578/18 &   1711/11& 400/8 & 20280/21
& 14/12
\\
4 & 24862/19& 3180/22& 2881/27 & 8865/22 &   9828/18& 1170/8 &
17530/27 & 12/18
\\
5 & 54726/34& 6010/29 & OoM & 226986/42 &   543865/28& TO &
141910/40 & 25/28
\\
\hline  BT 2 & 16/2& 460/2 &6/2 & 10/2 &   487/2& 20/1 &   0/2 & 0/2
\\
10 & 71/10& 520/10 & 119/10 & 16/10 &   627/10 & 520/1 &    0/10 &
30/10
\\
20 & 552/20& 1230/20 & 80/20 & 84/20 &   472174/20 & 3200/1 & 20/20
& 4400/20
\\
30 & 2415/30& 4080/30& 170/30 & 244/30  &   TO & 10330/1 &   80/30 &
4500/30
\\
40 & 7543/40& 11680/40 & 160/40 & 533/40 &   - & 24630/1 &   160/40
& 26120/40
\\
50 & 17573/50& 28420/50 & 300/50 & 1090/50  & - & 49329/1 &   250/50
& 84730/50
\\
60 & 35983/60& 59420/60 &  480/60 & 2123/60 &   - & 87970/1 & 420/60
& 233410/60
\\
70 & 67690/70& 113110/70  & 730/70 & 3529/70 &   -& 145270/1 &
620/70 & 522120/70
\\
80 & 157655/80& 202550/80 &  1080/80 & 1090/80 &   - & TO & 3310/80
& 979400/80
\\
\hline  BTC 2 & 16/3& 460/3 & 8/3 & 18/3 &   465/3 & 0/3 &   10/3 &
10/3
\\
10 & 89/19& 540/19 &504/19 & 45/19 &   715/19 & 39370/19 &   30/19
&57/19
\\
20 & 651/39& 1460/39 &98/39  & 211/39 &   - & - & 240/39 & 2039/39
\\
30 & 2721/59& 4820/59 & 268/59& 635/59 &   -&- & 1210/59 & 23629/59
\\
40 & 8009/79& 14250/79 &615/79 & 1498/79 & -&- & 3410/79 & 116156/79
\\
50 & 19074/99& 34220/99& 1287/99 & 10821/99 &    -& - & 8060/50 &
334879/99
\\
60 & 38393/119& 71650/119 & 2223/119& 5506/119 &   -& - & 15370/119
& TO
\\
70 & 65448/139 & 134880/139 & 3625/139& 2640/139& -& - & 27400/139 &
-
\\
\hline CubeCenter 3 &  TO & 70/9 & 10/9  & 20/9  & 40/9 & 28990/3 & 0/9 & 20/15 \\
5 & - &   1780/18  & 16/18  & 20/18  & 363/18 & TO & 0/19  & 28540/45 \\
7&  -& 27900/29  &  35/27 & 70/27 & 4782/27 & - &  20/34 & TO \\
9 & - &  177790/36  & 64/36  & 120/36 & 42258/36 & -  & 80/69  &  -\\
11 & - & 609540/47  &  130/45 & 230/45 & 26549/45 & -   & 190/68 &  -\\
\hline Ring 2 & TO & 30/6  & 0/5   & 0/5 & 31/5 &  TO & 0/5 & 360/12 \\
3 & - &   70/8  & 0/8  & 40/8  & 35/8 & - & 0/8  &  TO \\
4 & - &  250/13  & 10/11  & 30/11 & 60/11 &  -   & 20/11 & - \\
5 & - &  970/17  & 20/14  & 50/14 & 635/14 &  -  & 80/14 & - \\
6 & - &  4080/22  &  30/17 & 120/18 & 51678/17 & -    & 110/17 & - \\
7 & - &  75020/30  &  80/20 & 230/21 & TO &   -  & 300/20 & - \\
8 & - &  388300/29  &  160/23 & 600/24 & - &   -  & 480/23 &-  \\
\hline
\end{tabular}}
\caption{\label{tab4}Results for \caltalt\ using $h^{LUG
(DyX-SX)}_{RP}$, $POND$ using $h^{LUG}_{RP}$, MBP, KACMBP, GPT, CGP,
YKA, and CFF for conformant Rovers, Logistics, BT, BTC, Cube Center,
and Ring. The data is Total Time / \# Plan Steps, ``TO'' indicates a
time out (20 minutes), ``OoM'' indicates out of memory (1GB), and
``-'' indicates no attempt.}
\end{table}

\und{MBP} The MBP planner uses a cardinality heuristic that in many
cases overestimates plan distances (as per our implementation with
$h_{card}$).  MBP uses regression search for conformant plans, but
progression search for conditional plans. It is interesting to note
that in the more difficult problem instances in the Rovers and
Logistics domains MBP and KACMBP tend to generate much longer plans
than the other planners. MBP does outperform $POND$ in some cases
but does not find solutions in certain instances (like Rovers 5),
most likely because of its heuristic.  We note that KACMBP and MBP
are quite fast on the Cube Center and Ring domains, but have more
trouble on domains like Rovers and Logistics. This illustrates how a
heuristic modeling knowledge as opposed to reachability can do well
in domains where the challenge is uncertainty not reachability.

\und{Optimal Planners} The optimal approaches (CGP and GPT) tend not
to scale as well, despite their good solutions.  CGP has trouble
constructing its planning graphs as the parallel conformant plan
depth increases.  CGP spends quite a bit of time computing mutexes,
which increases the planning cost as plan lengths increase.  CGP
does much better on shallow and parallel domains like BT, where it
can find one step plans that dunk every package in parallel.

GPT performs progression search that is guided by a heuristic that
measures the cost of fully observable plans in state space. GPT
finds optimal serial plans but is not as effective when the size of
the search space increases. GPT fails to scale with the search space
because it becomes difficult to even compute its heuristic (due to a
larger state space as well).

\und{YKA}  YKA, like \caltalt\, is a regression planner, but the
search engine is very different and YKA uses a cardinality
heuristic. YKA performs well on all the domains because of its
search engine based on BDDs.  We notice a difference in progression
and regression by comparing $POND$ to YKA, similar to trends found
in the comparison between $POND$ and \caltalt. Additionally, it
seems YKA has a stronger regression search engine than \caltalt.
$POND$ is able to do better than YKA in the Rovers and Logistics
domains, but it is unclear whether that it is because of the search
direction or heuristics.

\und{CFF} Conformant FF, a progression planner using a relaxed plan
similar to the $LUG$ relaxed plan, does very well in the Rovers and
Logistics domains because it uses the highly optimized FF search
engine as well as a cheap to compute relaxed plan heuristic.
However, CFF does not do as well in the BT, BTC, Cube Center, and
Ring problems because there are not as many literals that will be
entailed by a belief state.  CFF relies on implicitly representing belief states in terms of
the literals that are entailed by the belief state, the initial belief state, and the action history.
When there are very few literals that can be entailed by the belief state,
reasoning about the belief state requires inference about the action history.
Another possible reason CFF suffers is our
encodings.  The Cube Center and Ring domains are naturally expressed
with multi-valued state features, and in our transformation to
binary state features we describe the values that must hold but also
the values that must not hold.  This is difficult for CFF because
the conditional effect antecedents contain several literals and
its heuristic is restricted to considering only one such literal.
It may be that CFF is choosing the wrong literal or simply not
enough literals to get effective heuristics.  However in BT and BTC
where we used only one literal in effect antecedents CFF still
performs poorly.

\begin{table}[t]
\center{\scalebox{.8}{
\begin{tabular}{|c||c||c||c||c||c|} \hline
Problem & POND & MBP & GPT & SGP & YKA \\
&   $h^{LUG}_{RP}$ & & & &\\
\hline Rovers 1 & 580/5 &  3312/11 & 3148/5 & 70/5 & 3210/5
\\
2 & 730/8 & 4713/75 & 5334/7 & 760/7& 6400/7
\\
3 & 810/8 & 5500/119  & 7434/8 & TO& 7490/8
\\
4 & 910/10 & 5674/146  & 11430/10& -& 11210/10
\\
5 &  7100/19 & 16301/76  & TO & -& TO
\\
6 &  13560/22 &  OoM  & -& -& -
\\
\hline Logistics 1 & 570/7 & 41/16 & 1023/7 & 5490/6& 1390/8
\\
2 &  1250/12 & 22660/177 & 5348/12 & TO & TO
\\
3 & 1210/9  &  2120/45 & 2010/8 & - & TO
\\
4 &  4160/15 & OoM  & TO &-& -
\\
5 &  20170/22 &  - & -&-& -
\\
\hline BT 2 & 460/2  & 0/2  & 510/2& 0/1& 0/2
\\
10 & 550/10  &  240/10 & 155314/10 & 70/1& 20/10
\\
20 &  1610/20 &  OoM & OoM & 950/1& 60/20
\\
30 & 5970/30  &  - & - & 4470/1& 200/30
\\
40 & 17620/40  & - & -& 13420/1& 400/40
\\
50 &  43020/50 & - & -& 32160/1& 810/50
\\
60 & 91990/60  &  -&- & 90407/1& 1350/60
\\
70 &  170510/70 & -  & -& 120010/1& 2210/70
\\
80 & 309940/80  & - & -& TO & 3290/80
\\
\hline BTC 2 & 470/2  & 20/2  & 529/2 & 10/2& 0/4
\\
10 & 590/10  &  280/10 & 213277/10 & TO & 210/12
\\
20 & 1960/20  & OoM & TO & -& 2540/22
\\
30 & 6910/30  & - & - & -& 13880/32
\\
40 & 19830/40  & - & - & -& 46160/42
\\
50 &  49080/50 & - & - & -& 109620/52
\\
60 & 103480/60  & - & - & -& 221460/62
\\
70 & 202040/70  & - & - & -& 41374/72
\\
\hline
\end{tabular}}}
\caption{\label{tab6}Results for $POND$ using $h^{LUG}_{RP}$,
MBP, GPT, SGP, and YKA for conditional Rovers, Logistics, BT, and
BTC. The data is Total Time / \# Maximum possible steps in a
execution, ``TO'' indicates a time out (20 minutes), ``OoM''
indicates out of memory (1GB), and ``-'' indicates no attempt.}
\end{table}

\subsection{Conditional Planning}

Table \ref{tab6} shows the results for testing the conditional
versions of the domains on $POND$, MBP, GPT, SGP, and YKA.

\und{MBP}  The $POND$ planner is very similar to MBP in that it uses
progression search.  $POND$ uses an AO* search, whereas the MBP
binary we used uses a depth first And-Or search. The depth first
search used by MBP contributes to highly sub-optimal maximum length
branches (as much as an order of magnitude longer than $POND$). For
instance, the plans generated by MBP for the Rovers domain have the
rover navigating back and forth between locations several times
before doing anything useful; this is not a situation beneficial for
actual mission use. MBP tends to not scale as well as $POND$ in all
of the domains we tested.  A possible reason for the performance of
MBP is that the Logistics and Rovers domains have sensory actions
with execution preconditions, which prevent branching early and
finding deterministic plan segments for each branch.  We
experimented with MBP using sensory actions without execution
preconditions and it was able to scale somewhat better, but plan
quality was much longer.

\und{Optimal Planners} GPT and SGP generate better solutions but
very slowly.  GPT does better on the Rovers and Logistics problems
because they exhibit some positive interaction in the plans, but
SGP does well on BT because its planning graph search is well
suited for shallow, yet broad (highly parallel) problems.

\und{YKA}  We see that YKA fares similar to GPT in Rovers and
Logistics, but has trouble scaling for other reasons.  We think
that YKA may be having trouble in regression because of sensory
actions since it was able to scale reasonably well in the
conformant version of the domains.  Despite this, YKA proves to do
very well in the BT and BTC problems.

\subsection{Empirical Evaluation Conclusions}

In our internal comparisons of heuristics within \caltalt\ and
$POND$, as well as external comparisons with several state of the
art conformant and conditional planners we have learned many
interesting lessons about heuristics for planning in belief space.

\begin{itemize}
    \item Distance based heuristics for belief space search help control conformant and conditional plan length
    because, as opposed to cardinality, the heuristics model
    desirable plan quality metrics.
    \item Planning graph heuristics for belief space search scale
    better than planning graph search and admissible heuristic
    search techniques.
    \item Of the planning graph heuristics presented, relaxed
    plans that take into account the overlap of individual plans
    between states of the source and destination belief states are the most accurate and tend to perform well across many domains.
    \item The LUG is an effective planning graph
    for both regression and progression search heuristics.
    \item In regression search, planning graphs that maintain
    only same-world mutexes provide the best trade-off between graph
    construction cost and heuristic informedness.
    \item Sampling possible worlds to construct planning graphs
    does reduce computational cost, but considering more worlds by exploiting planning graph
    structure common to possible worlds (as in the $LUG$), can
    be more efficient and informed.
    \item The LUG heuristics help our conditional planner, $POND$, to scale up in conditional domains, despite the fact that the
heuristic computation does not model observation actions.
\end{itemize}

\section{Related Work \& Discussion}

We discuss connections with several related works that involve
heuristics and/or conditional planning in the first half of this
section.  In the second part of the section we discuss how we can
extend our work to directly handle non-deterministic outcomes of
actions in heuristic computation.

\subsection{Related Work}
Much interest in conformant and conditional planning can be traced
to CGP \citep{AAAI98_IAAI98*889}, a conformant version of GraphPlan
\citep{blum95fast}, and SGP \citep{SGP}, the analogous conditional
version of GraphPlan.  Here the graph search is conducted on several
planning graphs, each constructed from one of the possible initial
states. More recent work on C-plan \citep{castellini01satconformant}
and Frag-Plan \citep{kurien02fragplan} generalize the CGP approach
by ordering the searches in the different worlds such that the plan
for the hardest to satisfy world is found first, and is then
extended to the other worlds. Although $\caltalt$\ and $POND$
utilize planning graphs similar to CGP and Frag-plan it only uses
them to compute reachability estimates. The search itself is
conducted in the space of belief states.

Another strand of work models conformant and conditional planning as
a search in the space of belief states. This started with
\citet{aaai93*724}, who concentrated on formulating a set of
admissible pruning conditions for controlling search. There were no
heuristics for choosing among unpruned nodes. GPT
\citep{bonet00planning} extended this idea to consider a simple form
of reachability heuristic. Specifically, in computing the estimated
cost of a belief state, GPT assumes that the initial state is fully
observable. The cost estimate itself is done in terms of
reachability (with dynamic programming rather than planning graphs).
GPT's reachability heuristic is similar to our $h_{m-RP}^{MG}$
heuristic because they both estimate the cost of the farthest
(maximum distance) state by looking at a deterministic relaxation of
the problem.  In comparison to GPT, $\caltalt$\ and $POND$ can be
seen as using heuristics that do a better job of considering the
cost of the belief state across the various possible worlds.

Another family of planners that search in belief states is the
MBP-family of planners---MBP \citep{bertoli01planning}, and KACMBP
\citep{bertoli02kacmbp}.  In contrast to $\caltalt$ but similar to
$POND$, the MBP-family of planners all represent belief states in
terms of binary decision diagrams.  Action application is modeled as
modifications to the BDDs. MBP supports both progression and
regression in the space of belief states, while KACMBP is a pure
progression planner.  Before computing heuristic estimates, KACMBP
pro-actively reduces the uncertainty in the belief state by
preferring uncertainty reducing actions. The motivation for this
approach is that applying cardinality heuristics to belief states
containing multiple states may not give accurate enough direction to
the search. While reducing the uncertainty seems to be an effective
idea, we note that (a) not all domains may contain actions that
reduce belief state uncertainty and (b) the need for uncertainty
reduction may be reduced when we have heuristics that effectively
reason about the multiple worlds (viz., our multiple planning graph
heuristics). Nevertheless, it could be very fruitful to integrate
knowledge goal ideas of KACMBP and the reachability heuristics of
$\caltalt$ and $POND$ to handle domains that contain both high
uncertainty and costly goals.

In contrast to these domain-independent approaches that only require
models of the domain physics, PKSPlan \citep{petrick02kbincomplete}
is a forward-chaining {\em knowledge-based planner} that requires
richer domain knowledge. The planner makes use of several knowledge
bases, as opposed to a single knowledge base taking the form of a
belief state.  The knowledge bases separate binary and multi-valued
variables, and planning and execution time knowledge.

YKA \citep{Rintanen03} is a regression conditional planner using
BDDs that uses a cardinality heuristic. Recently Rintanen has also
developed related reachability heuristics that consider distances
for groups of states, which do not rely on planning graphs
\citep{rintanen04}.

More recently, there has been closely related work on heuristics for
constructing conformant plans within the CFF planner \citep{cff}.
The planner represents belief states implicitly through a set of
known facts, the action history (leading to the belief state), and
the initial belief state.  CFF builds a planning graph forward from
the set of known literals to the goal literals and backwards to the
initial belief state.  In the planning graph, conditional effects
are restricted to single literals in their antecedent to enable
tractable 2-cnf reasoning.  From this planning graph, CFF extracts a
relaxed plan that represents supporting the goal belief state from
all states in the initial belief state. The biggest differences
between the $LUG$ and the CFF technique are that the $LUG$ reasons
only forward from the source belief state (assuming an explicit,
albeit symbolic, belief state), and the $LUG$ does not restrict the
number of literals in antecedents. As a result, the $LUG$ does not
lose the causal information nor perform backward reasoning to the
initial belief state.

Our handling of uncertainty through labels and label propagation is
reminiscent of and related to de Kleer's assumption based truth
maintenance system (ATMS) \citep{Kleer86}.  Where an ATMS uses
labels to identify the assumptions (contexts) where a particular
statement holds, a traditional truth maintenance system requires
extensive backtracking and consistency enforcement to identify other
contexts. Similarly, where we can reason about multiple possible
worlds (contexts) with the LUG simultaneously, the MG approach
requires, not backtracking, but reproduction of planning graphs for
other possible worlds.

Finally, $\caltalt$\ and $POND$ are also related to, and an
adaptation of the work on reachability heuristics for classical
planning, including $AltAlt$ \citep{nguyen02planning}, FF
\citep{hoffmann:nebel:jair-01} and HSP-r \citep{bonet99planning}.
$\caltalt$\ is the conformant extension to $AltAlt$ that uses
regression search (similar to HSP-r) guided by planning graph
heuristics.  $POND$ is similar to FF in that it uses progression
search with planning graph heuristics.

\subsection{Extension to Non-Deterministic Actions}

While the scope of our presentation and evaluation is restricted to
planning with initial state uncertainty and deterministic actions,
some of the planning graph techniques can be extended to include
non-deterministic actions of the type described by
\citet{rintanen-act}. Non-deterministic actions have effects that
are described in terms of a set of outcomes. For simplicity, we
consider Rintanen's conditionality normal form, where actions have a
set of conditional effects (as before) and each consequent is a
mutually-exclusive set of conjunctions (outcomes) -- one outcome of
the effect will result randomly.  We outline the generalization of
our single, multiple, and labelled planning graphs to reason with
non-deterministic actions.

\und{Single Planning Graphs}  Single planning graphs, that are built
from approximate belief states or a sampled state, do not lend
themselves to a straight-forward extension.  A single graph ignores
uncertainty in a belief state by unioning its literals or sampling a
state to form the initial planning graph layer. Continuing with the
single graph assumptions about uncertainty, it makes sense to treat
non-deterministic actions as deterministic. Similar to how we
approximate a belief state as a set of literals to form the initial
literal layer or sample a state, we can assume that a
non-deterministic effect adds all literals appearing in the effect
or samples an outcome as if the action were deterministic (i.e.
gives a set of literals).  Single graph relaxed plan heuristics thus
remain unchanged.

\und{Multiple Planning Graphs} Multiple planning graphs are very
much like Conformant GraphPlan \citep{AAAI98_IAAI98*889}.  We can
generalize splitting the non-determinism in the current belief state
into multiple initial literal layers to splitting the outcomes of
non-deterministic effects into multiple literal layers.  The idea is
to root a set of new planning graphs at each level, where each has
an initial literal layer containing literals supported by an
interpretation of the previous effect layer.  By interpretations of
the effect layer we mean every possible set of joint effect
outcomes. A set of effect outcomes is possible if no two outcomes
are outcomes of the same effect.  Relaxed plan extraction still
involves finding a relaxed plan in each planning graph. However,
since each planning graph is split many times (in a tree-like
structure) a relaxed plan is extracted from each ``path of the
tree''.

We note that this technique is not likely to scale because of the
exponential growth in redundant planning graph structure over
time.  Further, in our experiments CGP has enough trouble with
initial state uncertainty.  We expect that we should be able to do
much better with the $LUG$.

\und{Labelled Uncertainty Graph}  With multiple planning graphs we
are forced to capture non--determinism through splitting the
planning graphs not only in the initial literal layer, but also each
literal layer that follows at least one non-deterministic effect. We
saw in the $LUG$ that labels can capture the non-determinism that
drove us to split the initial literal layer in multiple graphs. As
such, these labels took on a syntactic form that describes subsets
of the states in our source belief state. In order to generalize
labels to capture non-determinism resulting from uncertain effects,
we need to extend their syntactic form. Our objective is to have a
label represent which sources of uncertainty (arising from the
source belief state or effects) causally support the labelled item.
We also introduce a graph layer ${\cal O}_k$ to represent outcomes
and how they connect effects and literals.

It might seem natural to describe the labels for outcomes in terms
of their affected literals, but this can lead to trouble. The
problem is that the literals in effect outcomes are describing
states at a different time than the literals in the projected belief
state. Further, an outcome that appears in two levels of the graph
is describing a random event at different times. Using state
literals to describe all labels will lead to confusion as to which
random events (state uncertainty and effect outcomes at distinct
steps) causally support a labelled item. A pathological example is
when we have an effect whose set of outcomes matches one-to-one with
the states in the source belief state. In such a case, by using
labels defined in terms of state literals we cannot distinguish
which random event (the state uncertainty or the effect uncertainty)
is described by the label.

We have two choices for describing effect outcomes in labels. In
both choices we introduce a new set of label variables to describe
how a literal layer is split. These new variables will be used to
describe effect outcomes in labels and will not be confused with
variables describing initial state uncertainty.  In the first case,
these variables will have a one-to-one matching with our original
set of literals, but can be thought of as time-stamped literals. The
number of variables we add to the label function is on the order of
2$F$ per level (the number of fluent literals -- assuming boolean
fluents). The second option is to describe outcomes in labels with a
new set of fluents, where each interpretation over the fluents is
matched to a particular outcome. In this case, we add on the order
of log $|{\cal O}_k|$ variables, where ${\cal O}_k$ is the $k^{th}$
outcome layer. It would actually be lower if many of the outcomes
were from deterministic effects because there is no need to describe
them in labels.  The former approach is likely to introduce fewer
variables when there are a lot of non-deterministic effects and they
affect quite a few of the same literals.  The latter will introduce
fewer variables when there are relatively few non-deterministic
effects whose outcomes are fairly independent.

With the generalized labelling, we can still say that an item is
reachable from the source belief state when its label is entailed by
the source belief state.  This is because even though we are adding
variables to labels, we are implicitly adding the fluents to the
source belief state.  For example, say we add a fluent $v$ to
describe two outcomes of an effect.  One outcome is labelled $v$,
the other $\neg v$.  We can express the source belief state $BS_P$
that is projected by the $LUG$ with the new fluent as $BS_P \wedge
(v \vee \neg v) = BS_P$.  An item labelled as $BS_P \wedge v$ will
not be entailed by the projected belief state (i.e. is unreachable)
because only one outcome causally supports it.  If both outcomes
support the item, then it will be reachable.

Given our notion of reachability, we can determine the level from
which to extract a relaxed plan.  The relaxed plan procedure does
not change much in terms of its semantics other than having the
extra graph layer for outcomes.  We still have to ensure that
literals are causally supported in all worlds they are labelled with
in a relaxed plan, whether or not the worlds are from the initial
state uncertainty or supporting non-deterministic effects.

\section{Conclusion}

With the intent of establishing a basis for belief state distance
estimates, we have:

\begin{itemize}
\item Discussed how heuristic measures can aggregate state
distance measures to capture positive interaction, negative
interaction, independence, and overlap.

\item Shown how to compute such heuristic measures on planning
graphs and provided empirical comparisons of these measures.

\item Found that exploiting planning graph structure to reduce the
cost of considering more possible states of a belief state is
preferable to sampling a subset of the states for the heuristics.

\item Shown that a labelled uncertainty graph can capture the same
support information as multiple graphs, and reduces the cost of
heuristic computation.

\item Shown that the labelled uncertainty graph is very useful for
conformant planning and, without considering observational actions
and knowledge, can perform well in conditional planning.
\end{itemize}

Our intent in this work was to provide a formal basis for measuring
the distance between belief states in terms of underlying state
distances.  We investigated several ways to aggregate the state
distances to reflect various assumptions about the interaction of
state to state trajectories.  The best of these measures turned out
to measure both positive interaction and independence, what we call
overlap.  We saw that planners using this notion of overlap tend to
do well across a large variety of domains and tend to have more
accurate heuristics.

We've also shown that planning with a Labelled Uncertainty planning
Graph $LUG$, a condensed version of the multiple graphs is useful
for encoding conformant reachability information.  Our main
innovation is the idea of ``labels" -- labels are attached to all
literals, actions, effect relations, and mutexes to indicate the set
of worlds in which those respective elements hold.  Our experimental
results show that the $LUG$ can outperform the multiple graph
approach.  In comparison to other approaches, we've also been able
to demonstrate the utility of structured reachability heuristics in
controlling plan length and boosting scalability for both conformant
and conditional planning.

We intend to investigate three additions to this work.  The first,
is to incorporate sensing and knowledge into the heuristics. We
already have some promising results without using these features in
the planning graphs, but hope that they will help the approaches
scale even better on conditional problems.  The second addition will
be to consider heuristics for stochastic planning problems.  The
major challenges here are to associate probabilities with labels to
indicate the likelihood of each possible world and integrate
reasoning about probabilistic action effects.

Lastly, we have recently extended the $LUG$ within the framework of
state agnostic planning graphs \citep{aaai05}, and hope to improve
the technique.  A state agnostic planning graph is essentially a
multiple source planning graph, where by analogy a conventional
planning graph has a single source.  Planning graphs are already
multiple destination, so in our generalization the state agnostic
planning graph allows us to compute the distance measure between any
pair of states or belief states.  The $LUG$ seeks to avoid
redundancy across the multiple planning graphs built for states in
the same belief state.  We extended this notion to avoid redundancy
in planning graphs built for every belief state. We have shown that
the state agnostic $LUG$ ($SLUG$) which is built once per search
episode (as opposed to a $LUG$ at each node) can reduce heuristic
computation cost without sacrificing informedness.

\paragraph{Acknowledgments} We would like to thank
Minh B. Do, Romeo Sanchez, Terry Zimmermam, Satish Kumar
Thittamaranahalli, and Will Cushing for helpful discussions and
feedback, Jussi Rintanen for help with the YKA planner, and
Piergiorgio Bertoli for help with the MBP planner.  This work was
supported in part by NASA grants NCC2-1225 and NAG2-1461, the NSF
grant IIS-0308139, the 2003 NASA RIACS SSRP, the ARCS Foundation,
and an IBM faculty award.

\appendix
\newpage
\renewcommand{\theequation}{A-\arabic{equation}}
\setcounter{equation}{0}

\section{Additional Heuristics}

For completeness, we present some additional heuristics adapted from
classical planning to reason about belief state distances in each
type of planning graph.  Many of these heuristics appeared in our
previous work \citep{ICAPS04-BrKa}.  We show how to compute the max,
sum, and level heuristics on the single graph $SG$, multiple graphs
$MG$, and the labelled uncertainty graph $LUG$. While these
heuristics tend to be less effective than the relaxed plan
heuristics, we provide them as reference.  As with Section 4, we
describe the heuristics in terms of regression search.

\subsection{Single Planning Graph Heuristics ($SG$)}

Like, the relaxed plan for the single unmodified planning graph, we
cannot aggregate state distances because all notion of separate
states is lost in forming the initial literal layer, thus we only
compute heuristics that do not aggregate state distances.

\und{No State Aggregation}
\begin{itemize}
\item {\bf Max}  In classical planning, the maximum cost literal
is used to get a max heuristic, but we use formulas to describe our
belief states, so we take the maximum cost clause as the cost of the
belief state to find the max heuristic $h^{SG}_{max}$.  The maximum
cost clause of the belief state, with respect to a single planning
graph, is:
\begin{equation}\label{sgmax}\notag
h^{SG}_{max}(BS_i) = \max\limits_{C\in{\kappa(BS_i)}} cost(C)
\end{equation}
where the cost of a clause is:
\begin{equation}\label{cost}\notag
    cost(C) = \min\limits_{l\in{C}} \min\limits_{k: l \in {\cal
    L}_k}k
\end{equation}
Here we find the cheapest literal as the cost of each clause to find
the maximum cost clause. This is an underestimate of the closest
state to our current belief state.

\item {\bf Sum}  Like the classical planning sum heuristic, we can
take the sum $h^{SG}_{sum}$ of the costs of the clauses in our
belief state to estimate our belief state distance
\begin{equation}\label{sgsum}\notag
    h^{SG}_{sum}(BS_i) = \sum\limits_{C\in{\kappa(BS_i)}} cost(C)
\end{equation}

This heuristic takes the summation of costs of literals in the
closest estimated state in the belief state, and is inadmissible
because there may be a single action that will support every clause,
and we could count it once for each clause.

\item {\bf Level}  When we have mutexes on the planning graph, we
can compute a level heuristic $h^{SG}_{level}$   (without mutexes
the level heuristic is equivalent to the max heuristic). The level
heuristic maintains the admissibility of the max heuristic but
improves the lower bound by considering what level of the planning
graph all literals in a constituent are non-pairwise mutex. The
level heuristic is computed by taking the minimum among the $\hat{S}
\in \hat{\xi}(BS_i)$, of the first level ($lev({S})$) in the
planning graph where literals of $\hat{S}$ are present with none of
them marked pairwise mutex. Formally:
\begin{equation}\label{sglevel}\notag
h^{SG}_{level}(BS_i) = \min\limits_{\hat{S}\in
\hat{\xi}(BS_i)}lev(\hat{S})
\end{equation}
\end{itemize}

\subsection{Multiple Planning Graph Heuristics ($MG$)}

Similar to the various relaxed plan heuristics for the multiple
graphs, we can compute a max, sum, or level heuristic on each of the
multiple planning graphs and aggregate them with a maximum or
summation to respectively measure positive interaction or
independence. The reason we cannot aggregate the individual graph
heuristics to measure overlap is that they are numbers, not sets of
actions. Measuring overlap involves taking the union of heuristics
from each graph and the union of numbers is not meaningful like the
union of action sets from relaxed plans.  Like before, there is no
reason to use multiple graphs if there is no state distance
aggregation.

\und{Positive Interaction Aggregation}
\begin{itemize}

\item {\bf Max} The max heuristic $h^{MG}_{m-max}$ is computed
with multiple planning graphs to measure positive interaction in the
$h^{MG}_{m-max}$ heuristic. This heuristic computes the maximum cost
clause in $\kappa(BS_i)$ for each graph $\gamma \in \Gamma$, similar
to how $h^{SG}_{m-max}(BS_i)$ is computed, and takes the maximum.
Formally:
\begin{equation}\label{mgmax}\notag
 h^{MG}_{m-max}(BS_i) =
\max\limits_{\gamma\in{\Gamma}}\left(h^{\gamma}_{max}(BS_i)\right)
\end{equation}
The $h^{MG}_{m-max}$ heuristic considers the minimum cost, relevant
literals of a belief state (those that are reachable given a
possible world for each graph $\gamma$) to get state measures. The
maximum is taken because the estimate accounts for the worst (i.e.,
the plan needed in the most difficult world to achieve the
subgoals).

\item {\bf Sum} The sum heuristic that measures positive
interaction for multiple planning graphs is $h^{MG}_{m-sum}$. It
computes the summation of the cost of the clauses in $\kappa(BS_i)$
for each graph $\gamma \in \Gamma$ and takes the maximum.  Formally:
\begin{equation}\label{mgsum}\notag
 h^{MG}_{m-sum}(BS_i) =
\max\limits_{\gamma\in{\Gamma}}\left(h^{\gamma}_{sum}(BS_i)\right)
\end{equation}
The heuristic considers the minimum cost, relevant literals of a
belief state (those that are reachable given the possible worlds
represented for each graph $\gamma$) to get state measures. As with
$h^{MG}_{m-max}$, the maximum is taken to estimate for the most
costly world.

\item {\bf Level} Similar to $h^{MG}_{m-max}$ and
$h^{MG}_{m-sum}$, the $h^{MG}_{m-level}$ heuristic is found by first
finding $h^{\gamma}_{level}$ for each graph $\gamma \in \Gamma$ to
get a state distance measure, and then taking the maximum across the
graphs. $h^{\gamma}_{level}(BS_i)$ is computed by taking the minimum
among the $\hat{S} \in \hat{\xi}(BS_i)$, of the first level
$lev^{\gamma}(\hat{S})$ in the planning graph $\gamma$ where
literals of $\hat{S}$ are present with none of them marked mutex.
Formally:
\begin{equation}\label{glevel}\notag
h^{\gamma}_{level}(BS_i) = \min\limits_{\hat{S}\in
\hat{\xi}(BS_i)}lev^{\gamma}(\hat{S})
\end{equation}
and
\begin{equation}\label{mglevel}\notag
 h^{MG}_{m-level}(BS_i) =
\max\limits_{\gamma\in{\Gamma}}(h^{\gamma}_{level}(BS_i))
\end{equation}

Note that this heuristic is admissible.  By the same reasoning as in
classical planning, the first level where all the subgoals are
present and non-mutex is an underestimate of the true cost of a
state. This holds for each of the graphs. Taking the maximum
accounts for the most difficult world in which to achieve a
constituent of $BS_i$ and is thus a provable underestimate of
$h^{*}$. GPT's max heuristic \citep{bonet00planning} is similar to
$h^{MG}_{m-level}$, but is computed with dynamic programming in
state space rather than planning graphs.
\end{itemize}

\und{Independence Aggregation} All heuristics mentioned for Positive
Interaction Aggregation can be augmented to take the summation of
costs found on the individual planning graphs rather than the
maximum.  We denote them as: $h^{MG}_{s-max}$, $h^{MG}_{s-sum}$, and
$h^{MG}_{s-level}$.  None of these heuristics are admissible because
the same action may be used in all worlds, but we count its cost for
every world by using summation.

\subsection{Labelled Uncertainty Graph ($LUG$)}

The max, sum, and level heuristics for the $LUG$ are similar to the
analogous multiple graph heuristics.  The main difference with these
heuristics for the $LUG$ is that it is much easier to compute
positive interaction measures than independence measures. The reason
positive interaction is easier to compute is that we find the cost
of a clause for all states in our belief state at once, rather than
on each of multiple planning graphs.  Like before, we do not
consider heuristics that do not aggregate state distances.

\und{Positive Interaction Aggregation}
\begin{itemize}
    \item {\bf Max}  The max heuristic $h^{LUG}_{m-max}$ for the $LUG$ finds
the maximum clause cost across clauses of the current belief state
$BS_i$.  The cost of a clause is the first level it becomes
reachable.  Formally:

\begin{equation}\label{lugmax}\notag
h^{LUG}_{m-max}(BS_i) = \max\limits_{C \in
\kappa(BS_i)}\left(\min\limits_{k: BS_P \models \ell_k^*(C)}
k\right)
\end{equation}

\item {\bf Sum}  The sum heuristic $h^{LUG}_{m-sum}$ for the $LUG$
sums the individual levels where each clause in $\kappa(BS_i)$ is
first reachable. Formally:
\begin{equation}\label{lugsum}\notag
    h^{LUG}_{m-sum}(BS_i) = \sum\limits_{C \in \kappa(BS_i)}\left(\min\limits_{
    k: BS_P \models \ell_k^*(C)} k\right)
\end{equation}

\item {\bf Level}  The level heuristic $h^{LUG}_{m-level}$ is the
index of the first level where $BS_i$ is reachable. Formally:

\begin{equation}\label{luglevel}\notag
    h^{LUG}_{m-level}(BS_i) = \min\limits_{
   k: BS_P \models \ell_k^*(BS_i) } i
\end{equation}
\end{itemize}

\und{Independence Aggregation} All heuristics mentioned for positive
interaction aggregation can be augmented to take the summation of
costs for each state in our belief state. This may be inefficient
due to the fact that we lose the benefit of having a $LUG$ by
evaluating a heuristic for each state of our $BS_P$, rather than all
states at once as in the positive interaction aggregation.  In such
a case we are doing work similar to the multiple graph heuristic
extraction, aside from the improved graph construction time. The
positive interaction aggregation is able to implicitly calculate the
maximum over all worlds for most of the heuristics, whereas for the
sum heuristic we need to explicitly find a cost for each world. We
denote the sum heuristics as: $h^{LUG}_{s-max}$, $h^{LUG}_{s-sum}$,
and $h^{LUG}_{s-level}$.

\section{Cross-World Mutexes}

Mutexes can develop not only in the same possible world but also
between two possible worlds, as described by
\citet{AAAI98_IAAI98*889}.  Cross-world mutexes are useful to
capture negative interactions in belief state distance measures
(mentioned in Section 3).  The representation of cross-world mutexes
requires another generalization for the labelling of mutexes.  Same
world mutexes require keeping only one label for the mutex to
signify all same possible worlds for which the mutex holds.  The
extended representation keeps a pair of labels, one for each element
in the mutex; if $x$ in possible world $S$ is mutex with $x'$ in
possible world $S'$, we denote the mutex as the pair
($\hat{\ell}_k(x) = S, \hat{\ell}_k(x') = S'$).

We can compute cross-world mutexes between several worlds of
elements $x$ and $x'$.  For example, if $\ell_k(x) = S_1 \vee S_2
\vee S_3$ and $\ell_k(x') = S_2 \vee S_3$, then to check for all
cross-world mutexes we need to consider mutexes for the world pairs
$(S_1, S_2), (S_1, S_3), (S_2, S_2), (S_2, S_3), (S_3, S_2)$, and
$(S_3, S_3)$.  We can also check for mutexes in the intersection of
the element labels $\ell_k(x) \wedge \ell_k(x') = S_2 \vee S_3$,
meaning the only cross world pairs we check for mutexes are $(S_2,
S_2), (S_2, S_3), (S_3, S_2)$, and $(S_3, S_3)$.

We can say that a formula $f$ is reachable from our projected belief
state $BS_P$, when considering cross-world mutexes, if for every
pair of states in $BS_P$, $f$ is reachable.  For a pair of states
$S$ and $S'$, $f$ is reachable if $S \wedge S' \models \ell_k^*(f)$
and for every pair of constituents $\hat{S''}, \hat{S'''} \in
\hat{f}$ such that $S \models \ell_k^*(\hat{S''})$ and $S' \models
\ell_k^*(\hat{S'''})$, there are no two literals in either
$\hat{S''}$ or $\hat{S'''}$ that are same-world mutex when $S = S'$,
and there is not a mutex between literals in $\hat{S''}$ and
$\hat{S'''}$, across the respective worlds $S$ and $S'$ when $S \neq
S'$. There is a mutex between a pair literals $l$ and $l'$,
respectively from $\hat{S''}$ and $\hat{S'''}$ if there is a mutex
$(\hat{\ell}_k(l), \hat{\ell}_k(l'))$ such that $S \models
\hat{\ell}_k(l)$ and $S' \models \hat{\ell}_k(l')$.

The computation of cross-world mutexes requires changes to some of
the mutex formulas, as outlined next. The major change is to check,
instead of all the single possible worlds $S$, all pairs of possible
worlds $S$ and $S'$ for mutexes.

\und{Action Mutexes} The action mutexes can now hold for actions
that are executable in different possible worlds.

\begin{itemize}
    \item {\bf Interference} Interference mutexes do not change
    for cross-world mutexes, except that there is a pair of labels where ($\hat{\ell}_k(a) = BS_P,
\hat{\ell}_k(a')  = BS_P$), instead of a single label.

    \item {\bf Competing Needs} Competing needs change mutexes for cross-world mutexes
    because two actions $a$ and $a'$, in worlds $S$ and $S'$
    respectively, could be competing.  Formally, a cross-world
    competing needs mutex ($(\hat{\ell}_k(a) = S,
\hat{\ell}_k(a') = S'$) exists between $a$ and $a'$ in worlds $S$
and $S'$ if:

    \begin{equation}\label{cractcompneedsmux}\notag
    \exists_{l \in \rho^e(a), l' \in \rho^e(a')} (\hat{\ell}_k(l)=S, \hat{\ell}_k(l')=S')
     \end{equation}

\end{itemize}

\und{Effect Mutexes} The effect mutexes can now hold for effects
that occur in different possible worlds.

\begin{itemize}
    \item {\bf Interference} Effect interference mutexes do not change
    for cross-world mutexes, except that there is a pair of labels where ($\hat{\ell}_k(
    \varphi^i(a)) = BS_P,
\hat{\ell}_k(\varphi^j(a'))  = BS_P$), instead of a single label.
    \item {\bf Competing Needs} Effect competing needs mutexes change for cross-world mutexes
    because two effects $\varphi^i(a)$ and $\varphi^j(a')$, in worlds $S$ and $S'$
    respectively, could be competing.  Formally, a cross-world
    competing needs mutex ($\hat{\ell}_k(\varphi^i(a)) = S,
\hat{\ell}_k(\varphi^j(a')) = S'$) exists between $\varphi^i(a)$ and
$\varphi^j(a')$ in worlds $S$ and $S'$ if:
\begin{equation}\label{creffcompneedsmux}\notag
    \exists_{l \in \rho^i(a), l' \in \rho^j(a')} (\hat{\ell}_k(l)=S, \hat{\ell}_k(l')=S')
\end{equation}

\item {\bf Induced} Induced mutexes change slightly for
cross-world mutexes.  The worlds where one effect induces another,
remains the same, but the mutex changes slightly.

If $\varphi^j(a)$ in $\hat{\ell}_k(\varphi^j(a))$ is mutex with
$\varphi^{h}(a')$ in $\hat{\ell}_k(\varphi^h(a'))$, and
$\varphi^i(a)$ induces effect $\varphi^j(a)$ in the possible worlds
described by $\ell_k(\varphi^i(a)) \wedge \ell_k(\varphi^j(a))$,
then there is an induced mutex between $\varphi^i(a)$ in
$\hat{\ell_k}(\varphi^j(a)) \wedge \ell_k(\varphi^i(a))$ and
$\varphi^h(a')$ in $\hat{\ell_k}(\varphi^{h}(a'))$ (see Figure
\ref{crinduced}).

\begin{figure}[t]
\center{\scalebox{.43}{\includegraphics{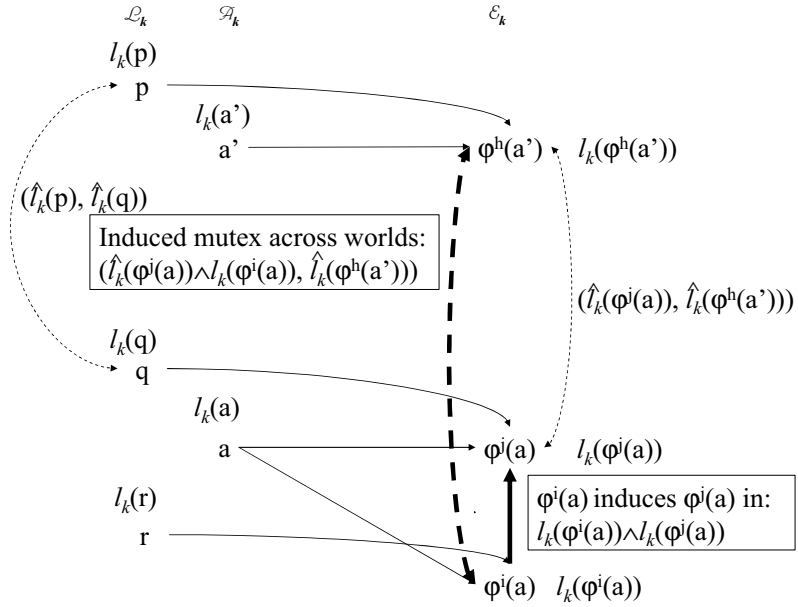}}
 \caption{\label{crinduced} Example of a cross-world induced effect mutex.}}
\end{figure}
\end{itemize}

\und{Literal Mutexes}  The literal mutexes can now hold for
literals that are supported in different possible worlds.

\begin{itemize}
    \item {\bf Inconsistent Support} changes for cross-world
    mutexes.  A mutex ($\hat{\ell}_k(l) = S, \hat{\ell}_k(l') = S'$) holds for $l$ in $S$ and $l'$ in $S'$
    if $\forall \varphi^i(a), \varphi^j(a') \in {\cal E}_{k-1}$ where $l \in \varepsilon^i(a), l' \in
                    \varepsilon^j(a')$, there is a mutex $\hat{\ell}_{k-1}(\varphi^i(a)) =
                    S, \hat{\ell}_{k-1}(\varphi^j(a')) =
                    S')$.
\end{itemize}

\newpage

%%%%%%%PS/PDF%%%%%%%%%%%%%%%%%%%%%%%%%%%
\bibliography{bryce05a}
\bibliographystyle{theapa}
%%%%%%%%%%%%%%%%%%%%%%%%%%%%%%%%%%%%%%%%

%%%%%%%%%%%%HTML%%%%%%%%%%%%%%%%%%%%%%%%
%\bibliography{bryce05a}
%\bibliographystyle{plain}
%%%%%%%%%%%%%%%%%%%%%%%%%%%%%%%%%%%%%%%%

\end{document}